\pdfoutput=1

\documentclass[10pt,twocolumn,letterpaper]{article}

\usepackage[pagenumbers]{cvpr}

\usepackage[english]{babel} %
\usepackage{dirtytalk}      %
\usepackage{placeins}       %
\usepackage{adjustbox}

\usepackage{cuted} %

\usepackage{currfile} %

\usepackage{caption} %

\newcommand{\TODO}[1]{\textbf{\color{red}[TODO: #1]}}
\renewcommand{\TODO}[1]{}

\renewcommand{\topfraction}{0.95}
\renewcommand{\dbltopfraction}{0.95}

\renewcommand{\textfraction}{0.05}
\renewcommand{\floatpagefraction}{0.85}
\renewcommand{\dblfloatpagefraction}{0.85}
\definecolor{bestgreen}{HTML}{CDE8C4}
\newcommand{\tc}[3]{\begin{tabular}[c]{@{}c@{}}#1\\#2\\#3\end{tabular}}
\newcommand{\tb}[1]{{\setlength{\fboxsep}{0.6pt}\colorbox{bestgreen}{\textbf{#1}}}}

\newif\ifdraft
\draftfalse

\ifdraft
  \newcommand{\PF}[1]{{\color{red}{\bf PF: #1}}}
  
  \definecolor{fsgreen}{rgb}{0.3, 0.7, 0.3}
  \newcommand{\FS}[1]{{\color{fsgreen}{\bf FS: #1}}}
  
  \newcommand{\NL}[1]{{\color{blue}{\bf NL: #1}}} 
  
  \newcommand{\AF}[1]{{\color{violet}{\bf AF: #1}}} 
  
  \newcommand{\LD}[1]{{\color{orange}{\bf LD: #1}}}
  
  \newcommand{\SJ}[1]{{\color{cyan}{\bf SJ: #1}}}

\else
  \newcommand{\PF}[1]{}
  
  \newcommand{\FS}[1]{}
  
  \newcommand{\NL}[1]{}
  
  \newcommand{\AF}[1]{}
  
  \newcommand{\LD}[1]{}
  
  \newcommand{\SJ}[1]{}
  
\fi

\definecolor{cvprblue}{rgb}{0.21,0.49,0.74}
\usepackage[pagebackref,breaklinks,colorlinks,allcolors=cvprblue]{hyperref}

\def\confName{CVPR}
\def\confYear{2027}

\title{AnyPrune: Optimization-Free Pruning for Any Feedforward 3D Gaussian Splatting Pipeline}

\author{
Nicola Leonetti$^{1}$ \enspace
Federico Stella$^{2}$ \enspace
Aoxiang Fan$^{2}$ \enspace
Saqib Javed$^{2}$ \enspace
Pascal Fua$^{2}$ \enspace
Luigi Di Stefano$^{1}$\\[4pt]
$^{1}$University of Bologna, Italy \qquad
$^{2}$EPFL, Switzerland\\
{\tt\small nicola.leonetti2@studio.unibo.it, luigi.distefano@unibo.it}\\
{\tt\small \{federico.stella, aoxiang.fan, saqib.javed, pascal.fua\}@epfl.ch}
}

\begin{document}
\maketitle
\begin{strip}
\centering
\includegraphics[width=\linewidth]{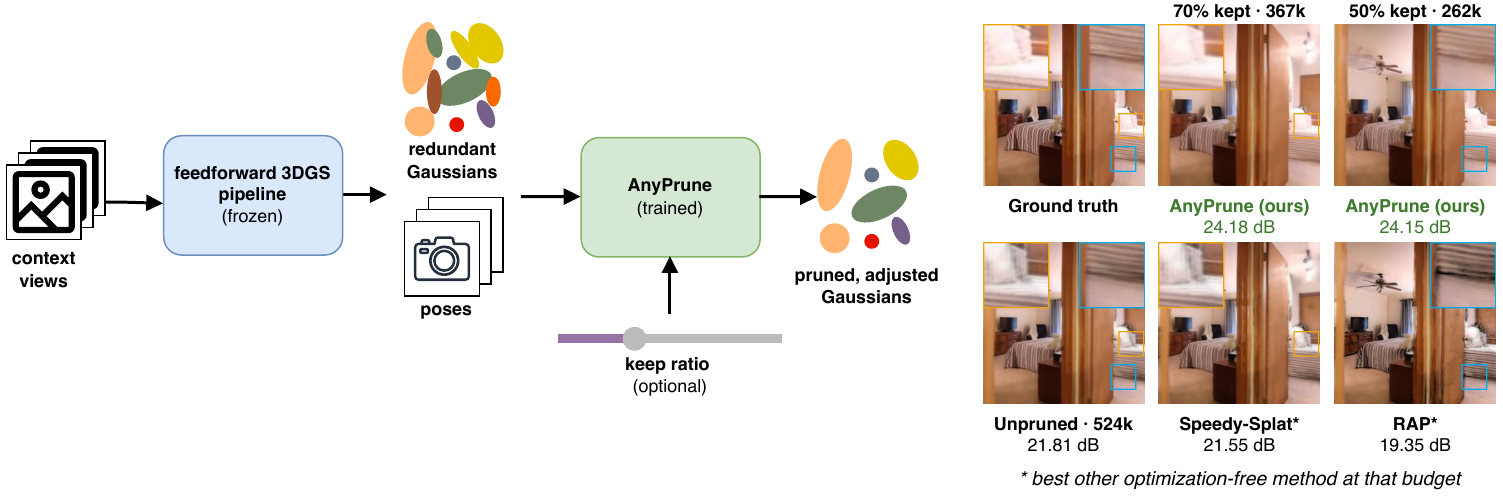}
\captionsetup{hypcap=false}
\captionof{figure}{\textbf{AnyPrune prunes the output of a frozen feedforward 3DGS pipeline.}
\textit{Left:} the pipeline uses the input images to predict a redundant set of 3D Gaussians.
AnyPrune, the only trained component, takes as input these Gaussians, the input images and their camera poses, scores the importance of every Gaussian and adjusts their parameters. The user can either retain the top-$k$ most important Gaussians by importance score to reach a specific pruning ratio or let the network pick a ratio.
\textit{Right:} a RealEstate10K test view reconstructed by F4Splat~\cite{f4splat} from 16 input images with its own internal pruning dial set to 50\% (524k Gaussians), pruned to 70\% (367k) and 50\% (262k) of its Gaussians.
The top row shows the ground truth and AnyPrune; the bottom row shows the unpruned render and, at each budget, the strongest competing optimization-free pruning method, \ie the one with the highest mean test PSNR at that budget: Speedy-Splat~\cite{speedy-splat} at 70\% and RAP~\cite{rap} at 50\%. Numbers are the PSNR of this view. AnyPrune is able to further prune and improve image quality starting from a pipeline's own pruning, showing that the pipeline's own pruning does not address all of its redundancy patterns.}
\captionsetup{hypcap=true}
\label{fig:\currfilebase}%
\end{strip}

\begin{abstract}
Feedforward 3D Gaussian Splatting (3DGS) pipelines reconstruct a scene from images in a single forward pass, representing it as real-time-renderable 3D Gaussians. However, the resulting representations contain far more Gaussians than are needed to render the scene faithfully, especially when there are many input views. This drastically increases memory and storage requirements. Existing pruning methods either require costly re-optimization, which defeats the purpose of a single forward pass, or rely on simple strategies that do not exploit all the available information. We present AnyPrune, a transformer-based, optimization-free pruning module designed to work on top of any feedforward 3DGS pipeline. To remove unnecessary Gaussians, AnyPrune learns and exploits pipeline-specific redundancy patterns through a short single-GPU training on the frozen target feedforward pipeline. AnyPrune achieves better rendering quality than existing optimization-free pruning methods in most settings and generalizes to unseen datasets, as shown through extensive experiments across multiple feedforward pipelines, datasets, and input view counts. It prunes a 1.6M-Gaussian field in about 5\,s on a consumer GPU, faster than every method that merges Gaussians or re-optimizes the field. It can also be combined with the pruning built into feedforward pipelines, pruning further at moderate ratios while preserving or improving rendering quality. Finally, it can even suggest a scene-specific pruning ratio that retains most of the quality. 

\end{abstract}

\section{Introduction}

3D Gaussian Splatting (3DGS)~\cite{kerbl-3DGS} achieves high-quality, real-time novel view synthesis by representing a scene as a set of 3D Gaussians. While the original formulation optimizes the parameters of the Gaussians for each scene by minimizing a rendering loss, feedforward pipelines~\cite{pixelsplat,anysplat,yonosplat,zipsplat} leverage 3D priors learned from large-scale training to construct a set of splats in a few seconds, avoiding costly iterative optimization. In both cases, the resulting representation is redundant~\cite{compact-3dgs,lightgaussian,pup-3dgs,splatt3rfusion,salon3r}, with far more Gaussians than are needed to render a scene faithfully. This increases memory and storage requirements and slows rendering. It also hinders deployment on resource-constrained hardware and makes scenes costlier to transmit.

Redundancy can be reduced by removing unnecessary Gaussians from the field predicted by the 3DGS reconstruction while preserving rendering quality~\cite{lightgaussian,radsplat,minigs,nanogs}. However, most pruning methods need to perform re-optimization steps on the Gaussian parameters to recover the rendering quality lost through pruning, which prevents their use in a feedforward scenario without defeating its very purpose of reconstructing a scene in a single forward pass. While it is possible to use these methods without the optimization steps, we show that their results have room for improvement  . Optimization-free methods~\cite{rap,pointsplat,nanogs} have therefore been proposed to avoid per-scene re-optimization, but they do not exploit all the available information, such as the input images. Another approach is to predict fewer Gaussians directly~\cite{resplat,ecosplat} or to move away from pixel alignment, in which one or more Gaussians are predicted for every pixel of every input view, to voxels, anchors, or tokens~\cite{learning-compact-3d-repr,volsplat,anchorsplat,tokengs}. However, these solutions are often designed to target the redundancies produced by a specific model, such as low-frequency regions~\cite{compactsplat,adaptivesplat} or overlapping views~\cite{ggn,freesplat}.

In short, there are no generic pruning methods competitive when used in conjunction with arbitrary feedforward 3DGS pipelines without requiring an expensive post-processing stage. To bridge this gap, we introduce AnyPrune: a trainable, optimization-free pruning module designed to work on top of any feedforward 3DGS pipeline. Our key insight is that a learned module running on top of a frozen 3DGS pipeline learns to fix the pipeline's specific redundancy patterns without being explicitly programmed to target them. To this end, given the Gaussians predicted by the pipeline and the input images they were produced from, a transformer-based network predicts (i) per-Gaussian importance scores that make it possible to prune the scene at any desired ratio, (ii) a scene-specific pruning ratio that aims to remove a large fraction of Gaussians with little to no loss in rendering quality, and (iii) parameter adjustments that let the kept Gaussians compensate for the removed ones. \cref{fig:teaser} illustrates our proposed pipeline. While being particularly well suited to the feedforward setting, AnyPrune makes no assumption about the architecture of the pipeline producing 3D Gaussians, because the feedforward pipeline is treated as a black box that takes images as input and outputs Gaussians that can be rendered from the input images' viewpoints, without making any further assumptions. Supporting a new pipeline does not require designing a new pruning method, but only a short training run of AnyPrune on the frozen pipeline's output.

We train AnyPrune separately on each one of several state-of-the-art feedforward 3DGS pipelines on DL3DV~\cite{dl3dv} and evaluate the trained modules zero-shot on RealEstate10K~\cite{realestate10k} and ACID~\cite{acid}. AnyPrune consistently outperforms existing pruning methods across pruning ratios and datasets. Keeping 50\% of the Gaussians, AnyPrune improves novel-view PSNR by 1.8\,dB over the best competing method, averaged over all pipelines, datasets and 2 to 32 input views.
AnyPrune prunes a 1.6M-Gaussian field in about 5\,s on a single consumer-grade GPU, faster than every competing method that merges Gaussians or re-optimizes the field.
AnyPrune outperforms existing general optimization-free pruning methods, giving the best novel-view PSNR in all 112 DL3DV settings from 2 to 16 views with 20--70\% of the Gaussians kept.  %

\noindent In summary, we make the following contributions:
\begin{itemize}
    \item We introduce AnyPrune, a transformer-based network that scores the importance of each Gaussian predicted by a feedforward 3DGS pipeline, suggests a scene-specific pruning ratio, and adjusts the surviving Gaussians to preserve rendering quality, all without per-scene optimization.
    
    \item By training it on the outputs of several feedforward 3DGS pipelines, we show that AnyPrune learns the redundancy patterns of a pipeline while generalizing zero-shot to unseen datasets.
    
    \item Our extensive experiments demonstrate that AnyPrune outperforms existing optimization-free pruning methods in most settings, including those built into feedforward 3DGS pipelines. It trains in a short run on a single GPU and is much faster than optimization-based baselines.
\end{itemize}

\section{Related Work}

\paragraph{Feedforward 3DGS and its redundancy.}
Pixel-aligned feedforward pipelines~\cite{pixelsplat,mvsplat,depthsplat,noposplat} predict one or more Gaussians for every pixel of every input view, so the number of Gaussians grows with the number of views and their resolution, and overlapping views produce duplicate primitives~\cite{long-lrm,splatt3rfusion}.
Several architectures are designed to tackle this problem.
Some pipelines decouple Gaussian prediction from pixels, predicting them from learnable tokens~\cite{tokengs,learning-compact-3d-repr,globalsplat}, from image-patch tokens merged across views~\cite{zipsplat}, from anchors or voxels in 3D space~\cite{anchorsplat,volsplat}, or exploiting detected keypoints~\cite{off-the-grid}.
Some allocate primitives according to image complexity~\cite{compactsplat,pixelgaussian,sparsesplat,f4splat} or predict them in a subsampled space and refine them recurrently~\cite{resplat}.
Others try to reduce the field after it has been predicted: by fusing Gaussians across views~\cite{ggn,freesplat,splatt3rfusion}, within voxels~\cite{anysplat} or into anchors~\cite{salon3r}; by pooling them along a Z-order curve~\cite{z-order-transformer}; by regularizing opacity so that transparent Gaussians can be discarded~\cite{yonosplat,long-lrm}; by learning to rank them for a target count~\cite{ecosplat}.
These mechanisms are designed for a specific architecture and trained with it, so there is often no obvious way to transfer them to other pipelines. They also target, in their design, only specific types of redundancy.
AnyPrune's core design idea is instead to treat the 3DGS pipeline as a black box to make it compatible with every pipeline, possibly stacking AnyPrune's pruning on top of the existing pruning solutions that a pipeline deploys.

\paragraph{Pruning per-scene optimized 3DGS.}
In the classic setting of 3DGS with per-scene optimization~\cite{kerbl-3DGS}, most pruning methods compute an importance score for each Gaussian, such as its contribution to the rendered training views~\cite{lightgaussian,radsplat,mesongs}, a weighted combination of its projected position, distance to other Gaussians and opacity~\cite{minigs}, or the sensitivity of the reconstruction error to its parameters~\cite{pup-3dgs}.
They then remove the Gaussians below a threshold or sample them in proportion to their score~\cite{minisplatting}, and fine-tune the survivors to recover rendering quality.
The threshold or ratio is set by hand, and a poor choice can collapse rendering quality~\cite{safeguardgs}.
Learned masks avoid this by letting the ratio emerge from a sparsity loss during optimization, and the resulting ratios may differ across scenes and datasets~\cite{compact-3dgs,lp3dgs,maskgaussian}.
Few methods compensate for the removed Gaussians with a learned module rather than by fine-tuning the survivors alone: SUNDAE~\cite{sundae} adds a neural rendering head, and FlexGS~\cite{flexgs} pairs a learned selector, which keeps Gaussians for any given ratio, with a transform field that adjusts the selected ones.
All methods above are optimized per scene, which defeats the purpose of feedforward reconstruction.

\paragraph{Optimization-free and general feedforward pruning.}
Some recent methods are designed to prune per-scene optimized 3DGS without requiring per-scene optimization.
Most derive importance analytically, from an approximate Hessian~\cite{refine}, from the effect of removing each Gaussian on the rendered training views, computed exactly~\cite{potr} or in a single forward pass~\cite{gaussianpop}, or by sensitivity sampling with provable guarantees~\cite{coreset-pruning}, while NanoGS~\cite{nanogs} progressively merges neighboring pairs of Gaussians.
RAP~\cite{rap} instead learns to predict importance from Gaussian attributes, and PointSplat~\cite{pointsplat} ranks Gaussians with a hand-crafted score and refines the survivors with a learned transformer.
None of these methods exploits information from the input images to the pipeline, and all of them leave the ratio to the user.
Some pruning strategies are designed in a way that makes them applicable, in principle, to any feedforward 3DGS pipeline.
AdaptiveSplat~\cite{adaptivesplat} prunes low-texture regions under a user-set budget and re-predicts the attributes of the remaining Gaussians, while Saliency-Guided Merging~\cite{saliency-merging} merges the Gaussians of a frozen backbone within superpixels seeded by a corner detector.
Both apply to several backbones, but rely on hand-crafted cues to decide where Gaussians are removed.
PhGS~\cite{phgs} uses an edge prior to prune the outputs of a frozen single-view pipeline and refines the survivors with a learned recurrent module, but does not cover redundancy patterns caused by multi-view overlap.
Learned modules such as Fuse-and-Refine~\cite{fuse-and-refine}, which fuses pixel-aligned Gaussians in a sparse voxel hierarchy, and ZPressor~\cite{zpressor}, which compresses the features of redundant views before decoding, need joint retraining with the pipeline.
None of these methods predicts a pruning ratio for each scene and not all of them offer a way to control how many primitives to keep.
In contrast, AnyPrune works on a frozen backbone and combines a learned criterion that uses the input views with a predicted scene-specific ratio.

\section{Method}
\label{sec:method}

\begin{figure*}[t]
\centering
\includegraphics[width=\linewidth]{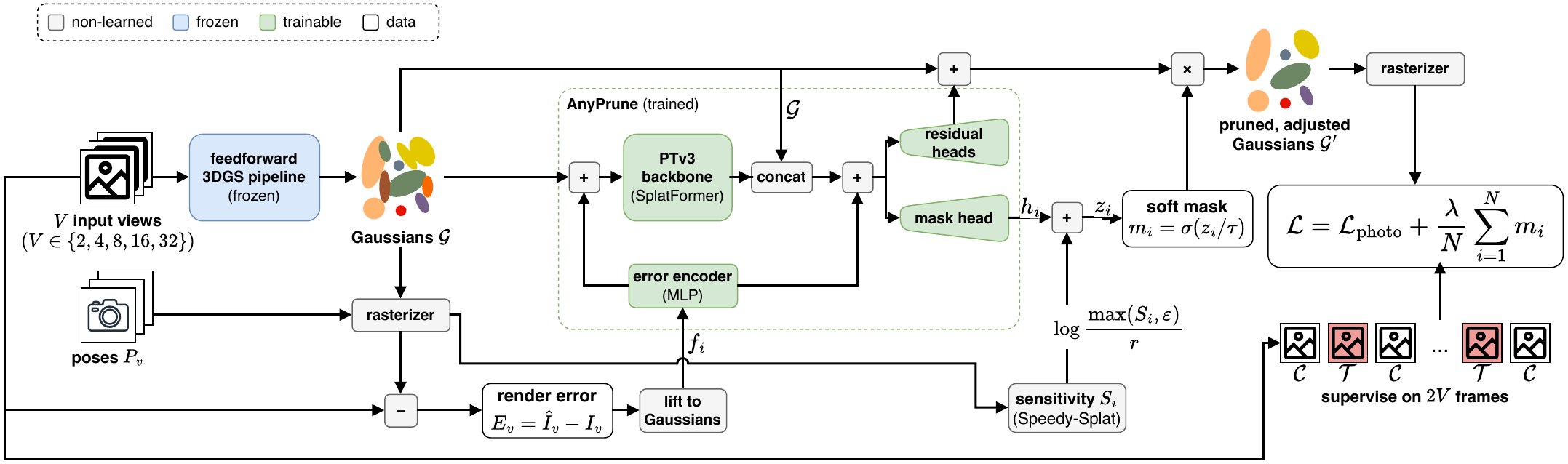}
\caption{\textbf{Overview of AnyPrune.}
A frozen feedforward 3DGS pipeline predicts the Gaussians $\mathcal{G}$ from the input views.
$\mathcal{G}$ is rendered from the input poses and the render error $E_v$ is used to compute a per-Gaussian feature vector $f_i$ which is encoded by an MLP and processed by the transformer together with the other features.
As in SplatFormer, every head reads the per-Gaussian features computed by the backbone and, via a residual connection, the attributes of the input Gaussians.
The residual heads predict adjustments to refine $\mathcal{G}$ into $\mathcal{G}'$; the mask head predicts a log-space adjustment $h_i$ to be added to the Speedy-Splat sensitivity score $S_i$ to obtain each Gaussian's keep logit $z_i$ (\cref{eq:logit}).
$z_i$ is used at training time to compute a factor $m_i=\sigma(z_i/\tau)$ by which to multiply 
the refined opacities of the Gaussians before rendering them.
The masked field is rendered from the input and held-out views and supervised with \cref{eq:loss}, whose photometric term is denoted $\mathcal{L}_{\mathrm{photo}}$. A sparsity term in the loss pushes the network to prune the redundant Gaussians.
At inference, instead of using the soft mask, the user can either keep the $k$ Gaussians with the highest $z_i$ to meet a budget, or those with $z_i>0$ for the model's own budget.}
\label{fig:\currfilebase}
\end{figure*}

AnyPrune operates on the output of a frozen feedforward 3DGS pipeline.
Given the predicted Gaussian field and the input images it was reconstructed from, 
a single forward pass of a point transformer~\cite{ptv3} refines the parameters of every Gaussian and assigns each of them a logit indicating its importance for rendering the scene.
The user can either keep the $k$ refined Gaussians with the highest logit to meet a budget or keep only the Gaussians with logit $> 0$, which corresponds to the ratio suggested by the model.
All of this happens without the need of per-scene optimization.

\subsection{Setting}
\label{sec:method-overview}
We assume that an existing feedforward 3DGS pipeline takes as input a set of images $\mathcal{I}=\{I_v\}_{v=1}^{V}$, optionally with their corresponding camera parameters $\mathcal{P}=\{P_v\}_{v=1}^{V}$, and produces a set of $N$ Gaussians $\mathcal{G}=\{(\mu_i,s_i,q_i,o_i,\kappa_i)\}_{i=1}^{N}$, where $\mu_i$, $s_i$, $q_i$, $o_i$ and $\kappa_i$ are the mean, per-axis scales, rotation quaternion, opacity and spherical-harmonic (SH) coefficients of the $i$-th Gaussian, following the original 3D Gaussian Splatting formulation~\cite{kerbl-3DGS}. If the pipeline uses pose-free input images, we assume that it will predict the corresponding camera parameters $\mathcal{P}$ as well. 
We treat the pipeline as a black box and only assume that $\mathcal{G}$ can be rendered from the input poses with a differentiable rasterizer.

Given $\mathcal{G}$, $\mathcal{I}$ and $\mathcal{P}$, AnyPrune uses a neural network $f_\theta$ to predict a refined field $\mathcal{G}'$ and, for every Gaussian $\mathcal{G}_i$, one \say{keep logit} $z_i$ that indicates how much that Gaussian contributes to the renderings and, therefore, how important it is to keep it. Formally:
\begin{equation}
  (\mathcal{G}',z)=f_\theta\big(\mathcal{G},\{(I_v,P_v)\}_{v=1}^{V}\big),\qquad z\in\mathbb{R}^{N},
  \label{eq:anyprune}
\end{equation}
where $I_v$ and $P_v$ are the RGB image and pose of view $v$.

To prune the field to a budget of $k$ Gaussians, we keep the $k$ Gaussians of $\mathcal{G}'$ with the highest keep logit.

To make our network learn the redundancy patterns of a pipeline, we train $f_\theta$ on the outputs of a frozen pipeline, supervising the rendered results with a photometric loss.
\Cref{fig:pipeline} shows the full training pipeline.

\subsection{Architecture}
\label{sec:method-network}

The neural network $f_\theta$ used to score and refine $\mathcal{G}$ is SplatFormer~\cite{splatformer}, a Point Transformer V3 (PTv3)~\cite{ptv3} with one MLP head per Gaussian attribute, initialized from its released checkpoint trained on Objaverse~\cite{objaverse}.
On top of the original SplatFormer architecture, we make two main modifications: 
\begin{itemize}
    \item A way to condition the backbone on the original field's render error. %
    \item A head predicting each Gaussian's importance logit $z_i$.
\end{itemize}

\paragraph{Input attributes and residual heads.}

We follow the original SplatFormer~\cite{splatformer} to feed the Gaussians into the network.
Each input Gaussian field $\mathcal{G}$ is first normalized to the unit cube  with an isotropic rescaling, so that fields from every pipeline enter at a common size, and then passed to the Point Transformer.  For each Gaussian, the network takes as input its mean, log-scales, opacity logit, rotation, and SH coefficients up to degree~1.
Fields with SH of degree 0 gain degree-1 coefficients, initialized to zero and predicted by the network, and coefficients above degree~1 bypass the network unchanged.
The backbone outputs a per-Gaussian feature vector, which each prediction head combines with the Gaussian's input attributes to predict a residual on a single attribute in the normalized space. Since field size and attribute statistics vary across pipelines and view counts, batch normalization uses the statistics of the field being processed at both training and inference time.

\paragraph{Render-error conditioning.}
To limit the loss in rendering quality caused by pruning, our method exploits the rendering error of the initial field with respect to the input views.
This error indicates where the reconstruction disagrees with the views it was built from, which gives our method an advantage over methods that only read Gaussian attributes~\cite{rap,pointsplat}.

Before running the network, we render $\mathcal{G}$ from the input poses, which are either required as input to the pipeline or are predicted by it during the reconstruction process.

We compute the signed error $E_v=\hat I_v-I_v$ between the rendering $\hat I_v$ and the input image $I_v$.
Rather than feeding the error maps directly to the network, we lift them onto the Gaussians: each Gaussian receives an extra four-dimensional feature, consisting of the weighted mean color error of the pixels it contributes to and its log-scale total visibility.
Formally:
\begin{equation}
  f_i=\bigg[\,\frac{\sum_{v,p}w_{vi}(p)\,E_v(p)}{\max\!\big(\sum_{v,p}w_{vi}(p),\,10^{-6}\big)}\,,\;\log\Big(1+\sum_{v,p}w_{vi}(p)\Big)\bigg]\in\mathbb{R}^{4},
  \label{eq:lift}
\end{equation}
where $w_{vi}(p)$ denotes the alpha-blending weight of Gaussian $i$ at pixel $p$ of view $v$.

A small MLP encodes the standardized $f_i$, and its code is added to the input embedding and to every head.

\paragraph{Bounded moves.}
Since the network occasionally predicted displacements of several hundred times the scale of a Gaussian, severely degrading rendering quality, we smoothly bound the move of each center to three times the Gaussian's largest scale (supplementary material).

\paragraph{Keep logits and sensitivity prior.}
Predicting the keep logits from scratch performed poorly, as we show in the ablation studies of the supplementary material. Instead, the network predicts an adjustment over a prior importance score.
As prior, we use the sensitivity score $S_i$ of Speedy-Splat~\cite{speedy-splat}, an efficient version of the Hessian-based sensitivity score of PUP~3D-GS~\cite{pup-3dgs}.
Computed on the initial field $\mathcal{G}$ from all input views, $S_i$ measures how much the renderings of the input views change when the footprint of Gaussian $i$ is perturbed, and is zero for Gaussians not visible in any view.
We choose this score because it is mathematically principled: it derives from a second-order approximation of the reconstruction error, rather than from heuristics on the Gaussian attributes.
A three-layer MLP head reads the same per-Gaussian input as the other adjustment heads and outputs an offset $h_i$, which corrects the Speedy-Splat prior score in log space:
\begin{equation}
  z_i=\log\max(S_i,\varepsilon)-\log r+h_i.
  \label{eq:logit}
\end{equation}
We use $r=0.0125$ as a reference score where $z_i$ is $0$ and $\varepsilon=10^{-6}$ as a floor value for $S_i$.
We initialize the last layer of the pruning head to zero, so that before training the model ranks the Gaussians by the Speedy-Splat score and AnyPrune's own cut keeps those with $S_i>r$.  %
Training thus starts from a strong optimization-free criterion and learns adjustments for a given pipeline.

\subsection{Training}
\label{sec:method-training}

\paragraph{Sampling.}
At each training step, we sample a scene and a number of input views $V$ within the range supported by the pipeline. We then take $2V$ consecutive frames starting from a random point in the capture.
The set $\mathcal{C}$ of the $V$ frames at even positions in the sequence will be given as input to the pipeline, while the set $\mathcal{T}$ of frames at odd positions will be held-out views.
The held-out cameras are aligned to the frame of the reconstruction using the input poses.

\paragraph{Objective.}
For each Gaussian $i$, the network predicts a keep logit $z_i$.
To ensure differentiability at training time, we approximate thresholding the keep logit by computing a \say{soft pruning mask} $m_i=\sigma(z_i/\tau)$ with temperature $\tau=0.5$.
We apply the mask by scaling the opacities of the refined field $\mathcal{G}'$, so that Gaussian $i$ is rendered with opacity $o_i'm_i$.

We render every view in $\mathcal{V} = \mathcal{C}\cup\mathcal{T}$, giving the $v$-th render $\hat I_v'$, and minimize
\begin{equation}
  \mathcal{L} = \frac{1}{|\mathcal{V}|} \sum_{v \in \mathcal{V}} \frac{\|\hat{I}_v' - I_v\|_1}{3HW} + \frac{\lambda}{N} \sum_{i=1}^{N} m_i .
  \label{eq:loss}
\end{equation}
The photometric term trains the network to refine the parameters of the Gaussians in the field, while the sparsity term encourages pruning.
AnyPrune's own cut, $z_i>0$ (equivalently $m_i>0.5$), emerges from the balance between the two terms, so the resulting pruning ratio adapts to each scene without a target ratio being specified. %
Since the refined field is rendered with the mask applied, the refinement learns to compensate for the Gaussians the mask removes.
Supervising held-out views in addition to the input views penalizes removing Gaussians that are needed only away from the input viewpoints.

\paragraph{Schedule.}
Training consists of two stages of 500 steps.
The first stage uses $\lambda=0.033$ and a constant learning rate.
The second stage starts from the weights of the first with a new optimizer state, lowers the sparsity weight to $\lambda=0.0083$ and decays the learning rate linearly to zero.
The higher sparsity weight of the first stage teaches the model to find good solutions under aggressive pruning, while the second stage targets a more balanced regime.
Both stages render only AnyPrune's own cut, so the adjustments never learn to compensate for cuts much heavier than the one the model suggests.
Where heavy cuts matter, we therefore add a third stage of 500 steps.
It freezes the pruning head and drops the sparsity term: at each step, it cuts the field by the ranking of the keep logits at three kept ratios drawn uniformly in $[0.05,1]$, and renders all three.
The third stage improves heavy cuts, but it can cost quality when most of the Gaussians are kept.
We use the final weights of the last stage, without checkpoint selection.

\subsection{Inference}
\label{sec:method-inference}

At test time, we compute $S$ on all input views, lift the rendering error from the same renderings, run $f_\theta$ once and cut the refined field according to the pruning logits $z$.
The soft mask is replaced by a hard selection, and the kept Gaussians retain their refined opacities $o_i'$. %

\paragraph{Residual scaling.}
During training, the full residuals are applied to the Gaussians.
At inference, we scale every residual in the normalized space by
\begin{equation}
  \begin{aligned}
    \gamma(N)&=\min\!\Big(1,\big(N_0/N\big)^{1/3}\Big),\\
    x_i^{\mathrm{out}}&=x_i+\gamma(N)\,(x_i'-x_i),
  \end{aligned}
  \label{eq:residual-scale}
\end{equation}
before the normalization is undone.
This is done so that the residuals shrink with the mean spacing between neighboring Gaussians, which scales as $N^{-1/3}$.
We set $N_0=100{,}352$, the size of a Gaussian field predicted by a pixel-aligned pipeline, \eg YoNoSplat~\cite{yonosplat}. The logits are not scaled.
Applying the same scaling during training led the network to predict larger raw residuals to compensate, which degraded performance at high view counts, while applying it at inference improves performance.
We therefore apply it at inference only (see the ablation studies in the supplementary material).

\section{Experiments}
\label{sec:experiments}

\begin{table*}[t]
\centering
\caption{\textbf{Pruning quality on DL3DV at 4 and 16 input views.}
Each cell gives novel-view PSNR\,$\uparrow$\,/\,SSIM\,$\uparrow$\,/\,LPIPS\,$\downarrow$ at a given share of Gaussians kept, for AnyPrune and the best other optimization-free method.
For pipelines with internal pruning, the share kept by it is in parentheses, kept ratios are relative to this reduced field, and ``+ own pruning'' is the internal pruning alone at the same number of Gaussians.
Superscripts name the best other method (once per cell if the same for all metrics): R~RAP~\cite{rap} retrained, R$'$~RAP released weights, M~Mini-Splatting~\cite{minisplatting}, S~Speedy-Splat~\cite{speedy-splat}, L~LightGaussian~\cite{lightgaussian}, G~GaussianPOP~\cite{gaussianpop}, Rd~RadSplat~\cite{radsplat}, C~provable coresets~\cite{coreset-pruning}, F~REFINE~\cite{refine}.
$^\ast$:~three-stage schedule. $^\dagger$:~12 views in the 16-view column. ReSplat at 16 views uses a second AnyPrune model.}
\label{tab:pruning}
\scriptsize
\setlength{\tabcolsep}{1.5pt}
\adjustbox{max width=\textwidth}{\begin{tabular}{@{}l@{\hskip 5pt}ccc@{\hskip 5pt}ccc@{}}
\toprule
 & \multicolumn{3}{c}{4 views} & \multicolumn{3}{c}{16 views} \\
\cmidrule(lr){2-4} \cmidrule(lr){5-7}
Method & 20\% & 50\% & 70\% & 20\% & 50\% & 70\% \\
\midrule
YoNoSplat~\cite{yonosplat} + AnyPrune & \tb{24.96}\,/\,\tb{0.794}\,/\,\tb{0.294} & \tb{27.25}\,/\,\tb{0.849}\,/\,\tb{0.192} & \tb{27.22}\,/\,\tb{0.851}\,/\,\tb{0.180} & \tb{24.40}\,/\,\tb{0.768}\,/\,\tb{0.272} & \tb{25.39}\,/\,\tb{0.801}\,/\,\tb{0.205} & \tb{25.38}\,/\,\tb{0.802}\,/\,\tb{0.201} \\
\quad + best other & 22.17\,/\,0.770\,/\,0.297\textsuperscript{\,R} & 25.72\,/\,0.838\,/\,0.198\textsuperscript{\,S} & 25.95\textsuperscript{\,S}\,/\,0.841\textsuperscript{\,S}\,/\,0.186\textsuperscript{\,G} & 23.13\,/\,0.753\,/\,0.276\textsuperscript{\,R} & 24.91\,/\,0.794\,/\,0.209\textsuperscript{\,S} & 24.98\textsuperscript{\,G}\,/\,0.795\textsuperscript{\,S}\,/\,0.205\textsuperscript{\,G} \\
\cmidrule(lr){1-7}
ReSplat$^{\ast}$~\cite{resplat} + AnyPrune & \tb{21.55}\,/\,\tb{0.733}\,/\,\tb{0.375} & \tb{26.97}\,/\,\tb{0.857}\,/\,\tb{0.234} & \tb{27.84}\,/\,\tb{0.878}\,/\,\tb{0.196} & \tb{23.28}\,/\,\tb{0.780}\,/\,0.316 & \tb{27.17}\,/\,\tb{0.867}\,/\,\tb{0.206} & \tb{27.84}\,/\,\tb{0.880}\,/\,\tb{0.181} \\
\quad + best other & 21.51\,/\,0.714\,/\,0.385\textsuperscript{\,R} & 25.01\,/\,0.834\,/\,0.244\textsuperscript{\,R} & 26.22\,/\,0.862\,/\,0.206\textsuperscript{\,M} & 23.20\,/\,0.774\,/\,\tb{0.301}\textsuperscript{\,R} & 25.49\textsuperscript{\,M}\,/\,0.851\textsuperscript{\,M}\,/\,0.212\textsuperscript{\,R} & 26.87\,/\,0.868\,/\,0.193\textsuperscript{\,M} \\
\cmidrule(lr){1-7}
TokenGS~\cite{tokengs} + AnyPrune & \tb{28.51}\,/\,\tb{0.877}\,/\,\tb{0.207} & \tb{28.52}\,/\,\tb{0.878}\,/\,\tb{0.205} & \tb{28.52}\,/\,\tb{0.878}\,/\,\tb{0.205} & \tb{25.39}\,/\,\tb{0.800}\,/\,\tb{0.294} & \tb{25.40}\,/\,\tb{0.800}\,/\,\tb{0.293} & \tb{25.40}\,/\,\tb{0.800}\,/\,\tb{0.293} \\
\quad + best other & 26.47\textsuperscript{\,R}\,/\,0.866\textsuperscript{\,G}\,/\,0.218\textsuperscript{\,G} & 26.36\textsuperscript{\,R$'$}\,/\,0.866\textsuperscript{\,F}\,/\,0.217\textsuperscript{\,G} & 26.36\textsuperscript{\,G}\,/\,0.866\textsuperscript{\,F}\,/\,0.217\textsuperscript{\,G} & 23.88\textsuperscript{\,R}\,/\,0.788\textsuperscript{\,G}\,/\,0.303\textsuperscript{\,G} & 23.85\textsuperscript{\,R}\,/\,0.788\textsuperscript{\,Rd}\,/\,0.303\textsuperscript{\,S} & 23.85\textsuperscript{\,R$'$}\,/\,0.788\textsuperscript{\,Rd}\,/\,0.303\textsuperscript{\,S} \\
\cmidrule(lr){1-7}
Off The Grid$^{\ast\dagger}$~\cite{off-the-grid} + AnyPrune & \tb{22.84}\,/\,\tb{0.755}\,/\,\tb{0.316} & \tb{25.35}\,/\,0.805\,/\,0.251 & \tb{25.40}\,/\,\tb{0.806}\,/\,0.245 & \tb{22.14}\,/\,\tb{0.728}\,/\,\tb{0.295} & \tb{23.44}\,/\,\tb{0.756}\,/\,0.253 & \tb{23.45}\,/\,\tb{0.755}\,/\,0.253 \\
\quad + best other & 19.08\textsuperscript{\,R}\,/\,0.679\textsuperscript{\,M}\,/\,0.348\textsuperscript{\,M} & 24.70\textsuperscript{\,R}\,/\,\tb{0.808\textsuperscript{\,R}}\,/\,\tb{0.217\textsuperscript{\,M}} & 24.63\,/\,\tb{0.806}\,/\,\tb{0.210}\textsuperscript{\,M} & 21.03\,/\,0.704\,/\,0.304\textsuperscript{\,M} & 23.14\,/\,0.755\,/\,\tb{0.240}\textsuperscript{\,M} & 23.09\,/\,0.753\,/\,\tb{0.238}\textsuperscript{\,M} \\
\cmidrule(lr){1-7}
EcoSplat$^{\ast}$~\cite{ecosplat} (50\%) + AnyPrune & \tb{20.52}\,/\,\tb{0.633}\,/\,\tb{0.431} & \tb{21.68}\,/\,\tb{0.664}\,/\,\tb{0.379} & \tb{21.76}\,/\,\tb{0.667}\,/\,\tb{0.370} & \tb{17.78}\,/\,\tb{0.541}\,/\,\tb{0.507} & \tb{18.16}\,/\,\tb{0.552}\,/\,\tb{0.487} & \tb{18.17}\,/\,\tb{0.553}\,/\,\tb{0.485} \\
\quad + best other & 14.93\,/\,0.449\,/\,0.519\textsuperscript{\,R} & 18.47\textsuperscript{\,M}\,/\,0.546\textsuperscript{\,R}\,/\,0.455\textsuperscript{\,L} & 19.40\,/\,0.580\,/\,0.426\textsuperscript{\,S} & 16.77\,/\,0.514\,/\,0.524\textsuperscript{\,R} & 17.59\textsuperscript{\,S}\,/\,0.533\textsuperscript{\,S}\,/\,0.493\textsuperscript{\,G} & 17.61\textsuperscript{\,M}\,/\,0.535\textsuperscript{\,S}\,/\,0.490\textsuperscript{\,G} \\
\quad + own pruning & 17.15\,/\,0.514\,/\,0.479 & 18.50\,/\,0.555\,/\,0.446 & 18.96\,/\,0.570\,/\,0.431 & 17.39\,/\,0.533\,/\,\tb{0.507} & 17.57\,/\,0.536\,/\,0.496 & 17.59\,/\,0.536\,/\,0.493 \\
\cmidrule(lr){1-7}
F4Splat~\cite{f4splat} (50\%) + AnyPrune & 19.04\,/\,0.614\,/\,0.404 & \tb{23.98}\,/\,0.750\,/\,0.287 & \tb{24.02}\,/\,0.754\,/\,0.277 & 16.82\,/\,0.494\,/\,0.467 & \tb{19.65}\,/\,\tb{0.591}\,/\,\tb{0.362} & \tb{19.82}\,/\,\tb{0.598}\,/\,\tb{0.343} \\
\quad + best other & 15.11\,/\,0.473\,/\,0.451\textsuperscript{\,C} & 21.59\textsuperscript{\,M}\,/\,0.689\textsuperscript{\,C}\,/\,0.307\textsuperscript{\,C} & 23.65\textsuperscript{\,M}\,/\,0.743\textsuperscript{\,Rd}\,/\,0.249\textsuperscript{\,C} & 13.37\textsuperscript{\,R}\,/\,0.384\textsuperscript{\,L}\,/\,0.516\textsuperscript{\,R} & 16.60\textsuperscript{\,R}\,/\,0.499\textsuperscript{\,L}\,/\,0.429\textsuperscript{\,R} & 17.81\textsuperscript{\,S}\,/\,0.539\textsuperscript{\,L}\,/\,0.396\textsuperscript{\,R} \\
\quad + own pruning & \tb{22.61}\,/\,\tb{0.731}\,/\,\tb{0.294} & 23.38\,/\,\tb{0.759}\,/\,\tb{0.240} & 23.58\,/\,\tb{0.763}\,/\,\tb{0.227} & \tb{18.84}\,/\,\tb{0.561}\,/\,\tb{0.377} & 18.65\,/\,0.562\,/\,0.368 & 18.52\,/\,0.562\,/\,0.368 \\
\cmidrule(lr){1-7}
AdaptiveSplat$^{\ast}$~\cite{adaptivesplat} (60\%) + AnyPrune & \tb{15.46}\,/\,\tb{0.501}\,/\,\tb{0.518} & \tb{21.65}\,/\,\tb{0.663}\,/\,\tb{0.388} & \tb{22.61}\,/\,\tb{0.689}\,/\,\tb{0.355} & \tb{16.50}\,/\,\tb{0.472}\,/\,\tb{0.504} & \tb{19.30}\,/\,\tb{0.557}\,/\,\tb{0.417} & \tb{19.46}\,/\,\tb{0.565}\,/\,\tb{0.406} \\
\quad + best other & 12.94\textsuperscript{\,R}\,/\,0.392\textsuperscript{\,M}\,/\,0.570\textsuperscript{\,M} & 18.80\,/\,0.588\,/\,0.441\textsuperscript{\,R} & 20.35\,/\,0.646\,/\,0.380\textsuperscript{\,R} & 14.17\,/\,0.405\,/\,0.546\textsuperscript{\,R} & 17.66\,/\,0.515\,/\,0.445\textsuperscript{\,R} & 18.30\textsuperscript{\,S}\,/\,0.535\textsuperscript{\,Rd}\,/\,0.423\textsuperscript{\,R} \\
\quad + own pruning & -- & 20.68\,/\,0.631\,/\,0.432 & 20.84\,/\,0.647\,/\,0.399 & -- & 18.44\,/\,0.539\,/\,0.463 & 18.43\,/\,0.542\,/\,0.442 \\
\bottomrule
\end{tabular}
}
\end{table*}

\subsection{Experimental setup}
\label{sec:exp-setup}

\paragraph{Feedforward 3DGS pipelines.}
We run AnyPrune on the output of seven recent feedforward 3DGS pipelines, each of them producing Gaussians in different ways: the pixel-aligned YoNoSplat~\cite{yonosplat}, the recurrent ReSplat~\cite{resplat}, the token-based TokenGS~\cite{tokengs}, the keypoint-based Off The Grid~\cite{off-the-grid}, the ranking-based EcoSplat~\cite{ecosplat}, the densification-based F4Splat~\cite{f4splat} and the texture-aware AdaptiveSplat~\cite{adaptivesplat}.
EcoSplat, F4Splat and AdaptiveSplat expose a way to set how many Gaussians they predict, and we run them with this setting roughly halfway through its range, where they keep 50\% of the Gaussians of their densest field (60\% for AdaptiveSplat ).
This is the setting in which these pipelines would be deployed, since it removes many Gaussians at little or no cost on DL3DV .
On those pipelines we therefore train and test AnyPrune both with the models' own pruning turned on and off, and we compare with it as well.
With the integrated pruning off, AnyPrune's results are weaker, with the best PSNR in 43 of 48 settings and the best LPIPS in 34 of them with 20--70\% kept (supplementary material).
We evaluate each pipeline from 2 views up to 12 to 32 views, depending on the pipeline; view counts not reported here and further details of the setup are given in the supplementary material.

\paragraph{Datasets.}
We train AnyPrune on Gaussian fields generated from scenes of DL3DV-10K~\cite{dl3dv} and test it on the 140 scenes of its test set. Without retraining, we test on the first 50 test clips of RealEstate10K~\cite{realestate10k} and ACID~\cite{acid}.
As in training (\cref{sec:method-training}), each test scene is made of consecutive frames that alternate between input and held-out views.

\paragraph{Pruning protocol and metrics.}
Every method prunes the same predicted field and uses all the input views that were given to the pipeline. Methods that prune according to an importance score are run without iteratively optimizing the Gaussian parameters after the pruning.
We keep from 10\% to 90\% of the Gaussians in steps of 10\% and report PSNR, SSIM~\cite{ssim} and LPIPS (VGG)~\cite{lpips} on the held-out views at a resolution of $448\times448$, averaged over the views of each scene and then over the scenes.
We compare against the optimization-free pruning scores RadSplat~\cite{radsplat}, Mini-Splatting~\cite{minisplatting}, LightGaussian~\cite{lightgaussian}, Speedy-Splat~\cite{speedy-splat}, GaussianPOP~\cite{gaussianpop}, REFINE~\cite{refine}, provable coresets~\cite{coreset-pruning} and RAP~\cite{rap}, the latter both with its released weights and retrained on each pipeline, and report the best of them for each setting and metric.
Methods that merge Gaussians or optimize the field are too slow to run on every pipeline and setting, especially on large fields, and we include them only in the runtime comparison (\cref{sec:exp-cost}). The exception is NanoGS~\cite{nanogs}, which we run on YoNoSplat: with 20--90\% kept, it trails AnyPrune in PSNR in 143 of 144 settings, by $0.7$\,dB on average, and leads only at 10\%. %

\paragraph{Training.}
We train one AnyPrune model for each feedforward 3DGS pipeline using the two-stage schedule of \cref{sec:method-training}.
We choose the schedule for each pipeline by looking at its DL3DV test results: on the pipelines where the two-stage model performs poorly when few Gaussians are kept (ReSplat, Off The Grid, EcoSplat and AdaptiveSplat), we add the third stage of $500$ steps of \cref{sec:method-training}.
On ReSplat, which releases separate weights for 8, 16 and 32 views, we train a second model for its 16-view weights (supplementary material).
Training one model takes 20\,min to 5.9\,h on one NVIDIA A100 GPU, depending on the pipeline.  %

\subsection{Pruning to a target budget}
\label{sec:exp-pruning}

\begin{figure}[t]
\centering
\includegraphics[width=\linewidth]{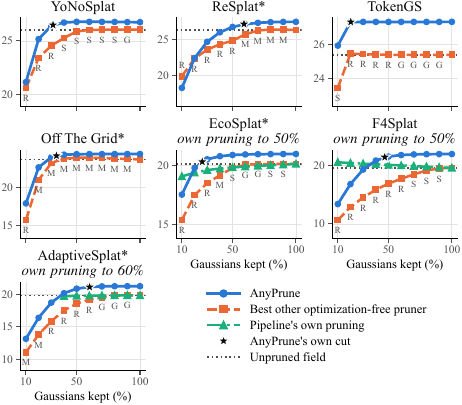}
\caption{\textbf{PSNR across kept ratios at 8 views on DL3DV.}
AnyPrune against the best other optimization-free method at each ratio (letters as in \cref{tab:pruning}) and, where available, the pipeline's own pruning.}
\label{fig:curves8}
\end{figure}

\paragraph{Comparison with other pruning methods.}
\Cref{tab:pruning} compares AnyPrune with the best other pruning method on DL3DV at 4 and 16 views, and \cref{fig:curves8} shows the full curves at 8 views. Full results of every method are in the supplementary.
AnyPrune gives better PSNR than the competing methods in all settings with 20--70\% kept, and better SSIM and LPIPS in most of them.
No single competitor is consistently second: RAP retrained on each pipeline is the strongest in about half of the settings, followed by Mini-Splatting and Speedy-Splat, depending on the pipeline, the view count and the kept ratio.

In the supplementary material we also show that AnyPrune often improves the field even when nothing is removed, since it adjusts the Gaussians, and when keeping half of the Gaussians it renders better than the unpruned field in 25 of 28 settings.
On TokenGS, both AnyPrune and the other methods maintain almost constant performance down to 20\% kept (\cref{fig:curves8}), since most of the Gaussians it predicts from its tokens are nearly transparent; there, AnyPrune's lead of about 2\,dB comes from its adjustments.

On DL3DV at 24 and 32 views, AnyPrune stays ahead of the other methods at every kept ratio on most pipelines; these results are in the supplementary material. %

\paragraph{Pruning on top of built-in pruning.}
\label{sec:exp-dial}
EcoSplat, F4Splat and AdaptiveSplat each have their own built-in way of reducing the number of Gaussians they produce, so training AnyPrune on those methods with the pruning turned on tests whether AnyPrune is able to learn additional pruning on top of these patterns.
For these pipelines, \cref{tab:pruning} starts from the reduced field that the pipeline gives and compares AnyPrune with turning the pruning dial (the setting that gives the method a budget of Gaussians) down to produce the same number of Gaussians.
When keeping 50--70\% of the reduced field, AnyPrune renders better in PSNR than the pipeline's own pruning in almost all settings across datasets, showing that AnyPrune's gains can be stacked on top of the ones coming from the model's internal pruning strategy.
AnyPrune's own cut keeps 48\% of the reduced field on average and still has a higher PSNR than it in 11 of 12 settings from 2 to 16 views (\cref{tab:own-cut} and supplementary material).
F4Splat's own pruning is ahead at low kept ratios: its dial allocates the Gaussians while predicting the field instead of removing them from a fixed one, and at 16 views its fields are already of low quality, with 18.3\,dB for the reduced field. Retrained on a more aggressive F4Splat setting, AnyPrune beats it at the same total Gaussian count (supplementary material).
The dials of EcoSplat and AdaptiveSplat are ahead of AnyPrune only in a few settings at the lowest kept ratio each can reach (\cref{fig:curves8} and supplementary material).
\label{sec:exp-datasets}
AnyPrune is trained only on DL3DV, and we test it on RealEstate10K and ACID without retraining.
Over the same pipelines, view counts and kept ratios, it still gives the best PSNR in nearly all settings, with a lead close to that on DL3DV (full tables in the supplementary material).

\subsection{Suggested pruning ratio}
\label{sec:exp-own-cut}

\begin{table}[t]
\centering
\caption{\textbf{AnyPrune's suggested ratio at 8 views.}
Share of Gaussians kept at $z_i>0$, averaged over DL3DV scenes, and change in novel-view metrics against the unpruned field. Notation as in \cref{tab:pruning}; other view counts are in the supplementary.}
\label{tab:own-cut}
\small
\setlength{\tabcolsep}{4pt}
\begin{tabular}{@{}l@{\hskip 6pt}rrrr@{}}
\toprule
 & Kept & PSNR & SSIM & LPIPS \\
\midrule
YoNoSplat & 31\% & +0.43 & -0.004 & +0.043 \\
ReSplat$^{\ast}$ & 59\% & +0.84 & +0.005 & +0.013 \\
TokenGS & 20\% & +1.96 & +0.011 & -0.009 \\
Off The Grid$^{\ast}$ & 34\% & +0.49 & +0.007 & +0.028 \\
EcoSplat$^{\ast}$ (50\%) & 26\% & +0.19 & +0.010 & +0.020 \\
F4Splat (50\%) & 47\% & +1.83 & +0.032 & +0.006 \\
AdaptiveSplat$^{\ast}$ (60\%) & 60\% & +1.27 & +0.018 & +0.001 \\
\bottomrule
\end{tabular}

\end{table}

\Cref{tab:own-cut} reports the cut AnyPrune suggests at 8 views, \ie keeping only the Gaussians with $z_i>0$; the other view counts, which behave similarly, are in the supplementary material.
The suggested ratio performs well across the board: from 2 to 16 views, AnyPrune keeps 43\% of the Gaussians on average, from 17\% for EcoSplat at 16 views to 85\% for AdaptiveSplat at 2 views, and it generally keeps fewer of them as the number of views, and with it the overlap between views, grows.
At AnyPrune's own cut, the pruned field renders better than the unpruned field in PSNR in 27 of 28 settings, by 1.1\,dB on average, and in SSIM in 22 of them.
The exception in PSNR is F4Splat at 4 views.
LPIPS is slightly worse than on the unpruned field in 22 of the 28 settings, by 0.04 or less in 16 of them, which is expected since AnyPrune is trained with an $L_1$ loss only. The largest, 0.08 and 0.09, are on F4Splat at 2 and 4 views.

\subsection{Runtime and peak memory usage}
\label{sec:exp-cost}

\begin{figure}[t]
\centering
\includegraphics[width=\linewidth]{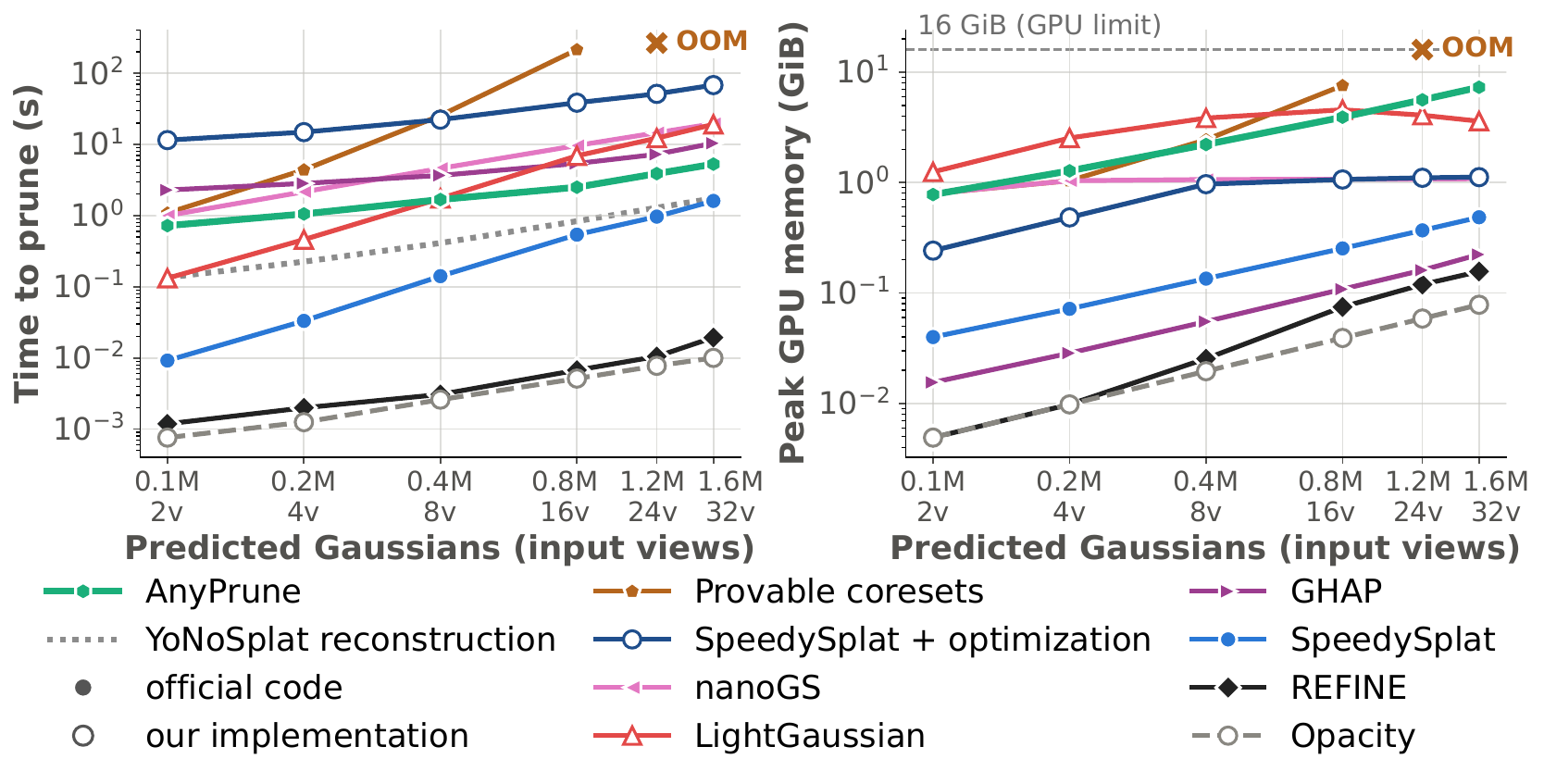}
\caption{\textbf{Pruning cost.}
Wall-clock time (\textit{left}) and peak GPU memory (\textit{right}) to prune YoNoSplat~\cite{yonosplat} fields of growing size on a consumer GPU.}
\label{fig:\currfilebase}
\end{figure}

\Cref{fig:timing} reports the time and peak GPU memory to prune YoNoSplat fields of 100k to 1.6M Gaussians on a consumer GPU (NVIDIA RTX 4060 Ti, 16\,GiB).
AnyPrune is not the fastest pruning method: single-score rules such as opacity, REFINE and Speedy-Splat take from milliseconds to 1.6\,s, and AnyPrune runs its refiner on top of the Speedy-Splat score.
It is, however, fast enough to follow a feedforward pipeline: it prunes the 1.6M-Gaussian field predicted from 32 views in 5.3\,s with 7.3\,GiB of peak memory, and it remains faster than the methods that merge Gaussians or optimize the field. %
The protocol, exact numbers and cost breakdown are in the supplementary material.

\FloatBarrier

\section{Conclusion}
\label{sec:conclusion}

We presented AnyPrune, a learned, optimization-free pruning module that works on the outputs of feedforward 3DGS pipelines.
Given the Gaussians predicted by the pipeline, the input views and their poses, a single forward pass ranks the Gaussians, suggests a pruning ratio and adjusts the Gaussians' parameters to compensate for the rendering quality lost through pruning, all without the need for per-scene optimization.
Trained separately on the outputs of seven pipelines that work in very different ways, AnyPrune renders better than the best other optimization-free method in most settings, by 1.8\,dB of PSNR on average when keeping half of the Gaussians. Notably, our method still finds and fixes redundancy in fields that a pipeline has already reduced with its own built-in pruning.
Our experiments show good zero-shot generalization on RealEstate10K and ACID.
AnyPrune runs in a few seconds on a consumer GPU.

\paragraph{Limitations and future directions.}
Even though the training time is relatively short, AnyPrune has to be re-trained for each pipeline, as it does not currently generalize well across pipelines, see supplementary material. Moreover, when keeping a low percentage of the initial Gaussians, AnyPrune still does not reach the same quality as some of the pipelines that offer a runtime dial to reduce the number of produced Gaussians, like F4Splat.
Finally, to be competitive at kept percentages far from the suggested percentage of AnyPrune, a third training stage of 500 steps at random cuts is necessary.
As a natural next step, a single larger model could be trained on the outputs of multiple pipelines to generalize across them. Also, a backbone other than SplatFormer, more specialized for the pruning task, may be explored. Finally, the performance of AnyPrune on per-scene optimized 3DGS is yet to be explored.

\clearpage
{
    \small
    \bibliographystyle{ieeenat_fullname}
    \bibliography{biblio}
}

\clearpage
\appendix
\renewcommand\thefigure{A.\arabic{figure}}
\renewcommand\thetable{A.\arabic{table}}
\renewcommand\thesection{\Alph{section}}
\setcounter{figure}{0}
\setcounter{table}{0}
\setcounter{section}{0}
\maketitlesupplementary
\renewcommand{\topfraction}{0.9}
\renewcommand{\dbltopfraction}{0.9}
\renewcommand{\textfraction}{0.07}
\renewcommand{\floatpagefraction}{0.5}
\renewcommand{\dblfloatpagefraction}{0.5}
\setcounter{topnumber}{3}
\setcounter{dbltopnumber}{3}
\makeatletter
\newcommand{\suppltop}[1]{\twocolumn[{%
  \renewenvironment{table*}[1][]{\def\@captype{table}\noindent\begin{minipage}{\textwidth}}{\end{minipage}\par\vspace{\floatsep}}%
  \renewenvironment{figure*}[1][]{\def\@captype{figure}\noindent\begin{minipage}{\textwidth}}{\end{minipage}\par\vspace{\floatsep}}%
  #1\vspace{\dimexpr\textfloatsep-\floatsep}}]}
\makeatother

\newcommand{\supplentry}[3][0em]{\noindent\hspace*{#1}\hyperref[#2]{\ref*{#2}\quad #3}\dotfill\pageref*{#2}\par}
\noindent\textbf{Contents}\par\smallskip
{\small
\supplentry{sec:suppl-impl}{Implementation details}
\supplentry[1em]{sec:suppl-impl-anyprune}{AnyPrune}
\supplentry[1em]{sec:suppl-setup}{Experimental setup}
\supplentry{sec:suppl-ablation}{Ablation studies}
\supplentry{sec:suppl-experiments}{Additional experiments}
\supplentry[1em]{sec:suppl-results}{Pruning results}
\supplentry[1em]{sec:suppl-cost}{Pruning cost}
\supplentry[1em]{sec:suppl-transfer}{Transfer across pipelines}
\supplentry[1em]{sec:suppl-train-length}{Training length}
\supplentry{sec:suppl-qualitative}{Qualitative results}
}
\vspace{1em}

\noindent 
This supplementary material provides additional details on our paper.
Please also see the accompanying file \texttt{experiments.pdf}, where we provide comprehensive charts and tables for the results on different pruning methods, pipelines, datasets, view counts and pruning ratios.
We will also release code and checkpoints.

\section{Implementation details}
\label{sec:suppl-impl}

\subsection{AnyPrune}
\label{sec:suppl-impl-anyprune}

\paragraph{Input encoding.}
As in SplatFormer \cite{splatformer}, before going into the backbone each Gaussian field is normalized on its own:
\begin{itemize}
  \item The centers of the Gaussians are translated so that the minimum corner of their axis-aligned bounding box is at the origin. They are also scaled in all directions by the same factor $a$ that brings the longest side of the box to 1, so that the $i$-th normalized center $\tilde\mu_i$ belongs to $[0,1]^3$.
  \item The $i$-th scale $s_i$ is written as $\log s_i+\log a$
  \item Opacities are written as logits 
  \item Rotations are written as unit quaternions.
\end{itemize}
The normalized means are voxelized on a $384^3$ grid for serialization.

\paragraph{Network.}
The PTv3~\cite{ptv3} backbone has five encoder and four decoder stages with serialized patch attention over z-order and Hilbert curves, and outputs a 96-dimensional feature per Gaussian.
Each residual head is a four-layer MLP that reads this feature concatenated with the 23 input channels.
The means are updated as $\tilde\mu_i'=\tilde\mu_i+\tanh(\Delta\mu_i)$, and the other attributes additively, $x_i'=x_i+\Delta x_i$, for log-scales, opacity logits, quaternions and SH; the refined quaternion is renormalized.
Covariances are rebuilt from the refined scales and rotations.
The render-error encoder maps $f_i$ to 64 channels through a hidden layer of width 128, and its code is added through bias-free linear layers to the input embedding of the backbone and to the first hidden layer of every head, including the mask head.
These layers are initialized to zero, so before training the conditioned network coincides with the pretrained one.
The mask head is a three-layer MLP of width 128.
The model has $48.0$M parameters, $47.6$M of which are in the backbone.

\paragraph{Bound on Gaussian center displacement.}
In our initial experiments, on some scenes AnyPrune predicted displacements for 1\% of the Gaussians that were several hundred times the scale of a single Gaussian.
After undoing the normalization, we therefore bound the displacement $\Delta_i=\mu_i'-\mu_i$ of each center by a multiple of the Gaussian's own size:
\begin{equation}
  \mu_i^{\mathrm{out}}=\mu_i+\Delta_i\,\frac{\ell_i\tanh(\|\Delta_i\|/\ell_i)}{\|\Delta_i\|},\qquad \ell_i=3\max_k s_{ik},
  \label{eq:bound}
\end{equation}
with $s_{ik}$ the input scales.
This mapping is close to the identity for small displacements, never moves a Gaussian farther than $\ell_i$, and is differentiable everywhere.

\paragraph{Render error conditioning.}
The error maps are computed on the renders downsampled to $112\times112$.
Before entering the network, each channel of $f_i$ is standardized to zero mean and unit variance, using the mean and standard deviation of that channel over all $N$ Gaussians of the field being processed. Gaussians that are not visible in any view receive $f_i=[0,0,0,0]$.

\paragraph{Held-out cameras.}
Each pipeline reconstructs in a frame of its own, related to that of the dataset by an unknown similarity transform, which we estimate from the input cameras, known in both frames.
The held-out cameras are mapped into the reconstruction frame by this transform and use the mean of the intrinsics of the input views, as the pipeline predicts them or is given them.
The Gaussians are rendered from the input views with the pipeline's own cameras.

\paragraph{Sampling.}
Each training step draws a training scene and a view count $V$ uniformly at random, the latter among the view counts at which the pipeline is tested (\cref{tab:pipelines}), and takes $2V$ consecutive frames from a uniformly random start, without skipping frames.
Predicted fields larger than $3\cdot10^{6}$ Gaussians are subsampled uniformly at random to that size, which only affects AdaptiveSplat~\cite{adaptivesplat} at 16 views or more, and the render error is lifted onto the subsampled field.

\paragraph{Optimization.}
We use Adam~\cite{adam} with $\beta=(0.9,0.999)$, $\epsilon=10^{-15}$ and no weight decay.
The learning rate is $3\cdot10^{-5}$ for the backbone and the residual heads, $1.5\cdot10^{-6}$ for the SH head, $3\cdot10^{-4}$ for the render-error conditioning and $10^{-3}$ for the mask head.
Gradients are clipped to a global norm of $2$, and the refiner runs in mixed precision with gradient checkpointing, on one scene per step.
Images are resized to a short side of 448 pixels and center-cropped to $448\times448$, the resolution of all renderings, losses and scores.
Renderings for the loss and the lifted error use gsplat~\cite{gsplat}.

\paragraph{Training schedules.}
In the two-stage schedule, the second stage starts from the weights of the first with a new optimizer state.
In the three-stage schedule, the first two stages use $\lambda=0.03$ and $\lambda=0.01$ instead of the values of \cref{sec:method-training}, and the three stages run one after the other with the same optimizer state.
Each stage starts from the learning rates above, kept constant in the first stage and decayed linearly to zero in the others.
The third stage sets the learning rate of the mask head to zero and drops the sparsity term.
Each of its steps draws three kept ratios, one uniformly in each third of $[0.05,1]$; for each ratio, it sets to zero the opacity of the refined Gaussians outside the top of the keep logits, renders all $2V$ views, and the loss is the mean of the three photometric terms.

\subsection{Experimental setup}
\label{sec:suppl-setup}

\begin{table}[t]
\centering
\caption{\textbf{Feedforward 3DGS pipelines.}
View counts they were trained at (n.r.: not reported); cameras required; Gaussians per view of the densest field; view counts we train and test AnyPrune on.}
\label{tab:pipelines}
\footnotesize
\setlength{\tabcolsep}{3pt}
\begin{tabular}{@{}llcr@{\hspace{8pt}}l@{}}
\toprule
Pipeline & \begin{tabular}[b]{@{}l@{}}Training\\views\end{tabular} & \begin{tabular}[b]{@{}c@{}}Cameras\\required\end{tabular} & \begin{tabular}[b]{@{}r@{}}Gaussians\\/ view\end{tabular} & Views \\
\midrule
YoNoSplat~\cite{yonosplat} & 2--32 & -- & 50{,}176 & 2--32 \\
ReSplat~\cite{resplat} & 8, 16, 32 & poses, intrinsics & 12{,}544 & 2--32 \\
TokenGS~\cite{tokengs} & 2, 4, 6 & poses, intrinsics & 131{,}072 & 2--16 \\
Off The Grid~\cite{off-the-grid} & ${\le}12$ & -- & 38{,}400 & 2--12 \\
EcoSplat~\cite{ecosplat} & 10 & intrinsics & 62{,}259 & 2--24 \\
F4Splat~\cite{f4splat} & 2, 24 & -- & 65{,}536 & 2--24 \\
AdaptiveSplat~\cite{adaptivesplat} & n.r. & -- & 200{,}704 & 2--32$^\ast$ \\
ZipSplat~\cite{zipsplat} & n.r. & -- & 20{,}736 & 2--32 \\
\bottomrule
\end{tabular}

\vspace{2pt}
\end{table}

\begin{figure}[t]
\centering
\includegraphics[width=\linewidth]{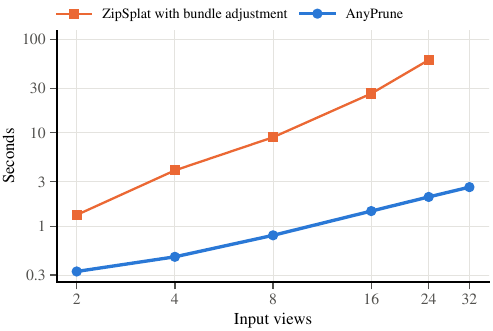}
\caption{\textbf{Cost of ZipSplat's bundle adjustment.}
Seconds per DL3DV scene for ZipSplat with its bundle adjustment and for AnyPrune, median of 5 scenes on an RTX 4060 Ti. AnyPrune's time covers only its pruning of the field ZipSplat outputs, not the forward pass and bundle adjustment that produce it.}
\label{fig:zipsplat-timing}
\end{figure}

\paragraph{Pipelines.}
\Cref{tab:pipelines} lists some additional information about the pipelines AnyPrune was trained on.
Every field is rendered at $448\times448$. We evaluate each pipeline at 2, 4, 8, 16, 24 and 32 views, up to the largest view count its paper trains or tests on, and train AnyPrune at the same view counts.
With its internal pruning dial on, AdaptiveSplat is costly in both time and memory. We therefore stop it at 16 views.
In addition to this, we report some peculiarities of some of the pipelines:
\begin{itemize}
  \item \label{sec:suppl-resplat} ReSplat~\cite{resplat} releases separate weights for 8, 16 and 32 views, and runs its 8-view weights up to 8 views. In the supplementary and in \texttt{experiments.pdf}, AnyPrune on ReSplat is one model trained, at each view count, on the fields of the weights that ReSplat runs there. \Cref{tab:pruning,tab:own-cut} instead use an AnyPrune model trained on the 8-view weights of ReSplat at every view count, and a second model trained on the 16-view weights for 16 views.
  \item \label{sec:suppl-f4splat} F4Splat's 24-view weights are trained at $256\times256$ only~\cite{f4splat}. Rendered at $448\times448$, their fields have holes, since the renderer~\cite{gsplat} dilates each projected Gaussian by a fixed number of pixels. This is why F4Splat's unpruned scores drop from 8 views on and stay below those of its paper. AnyPrune adjusts the Gaussians and partly fills these holes, so part of its lead on F4Splat from 8 views on comes from this mismatch.
  \item \label{sec:suppl-zipsplat} We leave ZipSplat~\cite{zipsplat} out of the main paper because its inference ends with test-time bundle adjustment. This puts it among the methods that optimize after pruning (\cref{sec:exp-setup}), and the bundle adjustment is costly: from 2 to 24 views, ZipSplat's forward pass and bundle adjustment together take 4 to 29 times longer than AnyPrune takes to prune their output (\cref{fig:zipsplat-timing}). We run its released 504-pixel checkpoint.
\end{itemize}

\paragraph{Internal pruning dials.}
EcoSplat~\cite{ecosplat}, F4Splat, AdaptiveSplat and ZipSplat have an internal pruning mechanism, whose dial sets how many Gaussians they predict.
EcoSplat keeps the $\rho hw$ most opaque Gaussians of each view, with $\rho$ from $1/16$ to $0.95$; F4Splat keeps $b$ Gaussians per input pixel, from $b=1/16$ to $1$; AdaptiveSplat prunes a share $p$ of the pixels of each view, from $0$ to $80\%$; ZipSplat keeps a share $c$ of its tokens.
With the internal pruning dial on, we set $\rho=0.475$, $b=0.5$, $p=40\%$ and $c=0.5$, halfway through each range.
AnyPrune is trained and tested on these reduced fields, and the pipeline's own pruning is obtained by turning this dial further down.

\paragraph{Data.}
AnyPrune is trained on 785 scenes of the first 1K subset of DL3DV-10K~\cite{dl3dv}, a random 80\% of it. We leave out the 19 scenes of DL3DV-Benchmark. We read it at $480\times270$.
At $V$ views, a DL3DV test scene is made of $2V$ consecutive frames from a starting point drawn randomly once per scene with a fixed seed: the pipeline receives the frames at even positions, and we evaluate on those at odd positions, at most 16 of them evenly spaced.
On RealEstate10K~\cite{realestate10k} and ACID~\cite{acid}, we take the first 50 test clips with at least 253 frames, in the order of pixelSplat's~\cite{pixelsplat} chunks, read at a stride of 13 frames up to 8 views and of 4 frames from 12 views on.
Metrics are averaged over the held-out views of each scene and then over the scenes.

\paragraph{Pruning baselines.}
When comparing to other pruning methods, we cut the predicted field to its top-scoring 
Gaussians, without any additional optimization.
We use the released code of each method where it exists.
Provable coresets~\cite{coreset-pruning} are too costly beyond 4 views, so we leave it out beyond that (\cref{sec:suppl-cost}).
RAP \cite{rap} is reported with its released weights and retrained on the fields of each pipeline following its paper.

\paragraph{Methods that merge or optimize.}
NanoGS~\cite{nanogs}, GHAP and Speedy-Splat~\cite{speedy-splat} followed by 3{,}000 iterations of 3DGS optimization~\cite{kerbl-3DGS} on the input views are slower than AnyPrune at every view count (\cref{tab:timing}).
We leave them out of the quality comparison also because, when used on Gaussian fields produced by a feedforward method, the merging-based methods do not make up for their cost in quality: we ran NanoGS and GHAP at their released defaults on YoNoSplat, on the first 20 test scenes of each dataset, and compared them with AnyPrune on the same scenes.
NanoGS trails AnyPrune whenever at least 20\% of the Gaussians are kept, and is ahead only at 10\%, where its field is itself degraded (\texttt{experiments.pdf}).
GHAP is ahead of AnyPrune in only 3 of the 162 settings, by at most 0.05\,dB.

\section{Ablation studies}
\label{sec:suppl-ablation}

\begin{table}[t]
\centering
\caption{\textbf{Ablation on YoNoSplat.}
PSNR at 8 views on held-out DL3DV scenes over different pruning ratios; rows give the difference with the paper's model.
Baselines cut the unadjusted field by the Speedy-Splat score or by opacity.
``w/ residual scaling'' also applies the residual scaling in training; it is trained on the full DL3DV training set at 2 to 32 views and compared with the paper's model on the full test set.}
\label{tab:ablation-yonosplat}
\footnotesize
\setlength{\tabcolsep}{1pt}
\begin{tabular}{@{}lccc@{\hskip 4pt}cc@{}}
\toprule
 & \multicolumn{3}{c}{$\Delta$PSNR, kept} & \multicolumn{2}{c}{AnyPrune's own cut} \\
\cmidrule(lr){2-4} \cmidrule(lr){5-6}
Model & 10--20\% & 30--50\% & 60--100\% & Kept & $\Delta$PSNR \\
\midrule
AnyPrune & 23.26 & 27.43 & 27.70 & 33\% & 27.21 \\
\midrule
\multicolumn{6}{@{}l}{\textit{Network}} \\
w/o Speedy-Splat prior & -1.94 & -2.09 & -2.21 & 99\% & -1.78 \\
w/o learned logit offset & -6.87 & -1.57 & -0.01 & 38\% & -2.07 \\
w/o adjustments & -0.24 & -0.95 & -1.16 & 41\% & -0.50 \\
w/o SH adjustments & -0.17 & -0.01 & -0.01 & 33\% & +0.00 \\
w/o render-error input & +0.21 & -0.30 & -0.29 & 36\% & -0.22 \\
random initialization & -0.51 & -0.24 & -0.15 & 35\% & -0.18 \\
\midrule
\multicolumn{6}{@{}l}{\textit{Training}} \\
w/o sparsity term & -8.52 & -2.50 & -0.16 & 100\% & +0.49 \\
w/o first stage & -0.24 & -0.12 & -0.04 & 37\% & +0.06 \\
first stage only & +1.40 & -0.13 & -0.32 & 15\% & -2.41 \\
one stage of 1000 steps & +1.63 & -0.19 & -0.50 & 16\% & -1.71 \\
w/ residual scaling & +0.84 & +0.02 & -0.08 & 34\% & +0.16 \\
\midrule
\multicolumn{6}{@{}l}{\textit{Switched off at test time}} \\
adjustments & -1.69 & -1.64 & -1.23 & -- & -- \\
learned logit offset & -5.79 & -0.89 & +0.00 & 38\% & -1.09 \\
render-error input & -0.32 & -0.55 & -0.56 & 33\% & -0.73 \\
residual scaling & +0.75 & +0.07 & -0.20 & 33\% & +0.33 \\
\midrule
\multicolumn{6}{@{}l}{\textit{Baselines}} \\
Speedy-Splat~\cite{speedy-splat} & -7.11 & -2.58 & -1.23 & -- & -- \\
opacity & -5.00 & -2.70 & -1.28 & -- & -- \\
\bottomrule
\end{tabular}

\end{table}

\begin{table}[!b]
\centering
\caption{\textbf{Effect of the third training stage.}
Change in PSNR, SSIM and LPIPS on DL3DV from adding the third stage, averaged over view counts, for YoNoSplat and Off The Grid. Bold: the third stage helps.}
\label{tab:ablation-recipe}
\footnotesize
\setlength{\tabcolsep}{3pt}
\adjustbox{max width=\columnwidth}{\begin{tabular}{@{}lcccccc@{}}
\toprule
Kept & 10\% & 20\% & 30\% & 50\% & 70\% & 100\% \\
\midrule
YoNoSplat & \tc{\textbf{+1.38}}{\textbf{+0.018}}{\textbf{-0.009}} & \tc{\textbf{+0.53}}{\textbf{+0.006}}{\textbf{-0.004}} & \tc{\textbf{+0.03}}{+0.000}{+0.001} & \tc{-0.11}{-0.003}{+0.006} & \tc{-0.08}{-0.003}{+0.008} & \tc{-0.04}{-0.002}{+0.007} \\ \addlinespace[3pt]
Off The Grid & \tc{\textbf{+4.21}}{\textbf{+0.189}}{\textbf{-0.077}} & \tc{\textbf{+3.00}}{\textbf{+0.081}}{\textbf{-0.043}} & \tc{\textbf{+0.67}}{\textbf{+0.008}}{+0.006} & \tc{-0.68}{-0.023}{+0.041} & \tc{-0.69}{-0.023}{+0.043} & \tc{-0.63}{-0.021}{+0.041} \\ \addlinespace[3pt]
\bottomrule
\end{tabular}
}
\end{table}

\begin{table*}[t]
\centering
\caption{\textbf{Pruning quality on RealEstate10K at 4 and 16 input views,} with AnyPrune trained on DL3DV only.
Layout and notation as in \cref{tab:pruning}.
In parentheses, the share kept by the pipeline's internal pruning dial.}
\label{tab:pruning-re10k}
\scriptsize
\setlength{\tabcolsep}{3pt}\renewcommand{\arraystretch}{0.9}
\adjustbox{max width=\textwidth}{\begin{tabular}{@{}l@{\hskip 5pt}ccc@{\hskip 5pt}ccc@{}}
\toprule
 & \multicolumn{3}{c}{4 views} & \multicolumn{3}{c}{16 views} \\
\cmidrule(lr){2-4} \cmidrule(lr){5-7}
Method & 20\% & 50\% & 70\% & 20\% & 50\% & 70\% \\
\midrule
YoNoSplat~\cite{yonosplat} + AnyPrune & \tb{23.02}\,/\,\tb{0.768}\,/\,\tb{0.281} & \tb{23.32}\,/\,0.770\,/\,\tb{0.235} & \tb{23.31}\,/\,0.770\,/\,\tb{0.234} & \tb{23.16}\,/\,\tb{0.766}\,/\,\tb{0.255} & \tb{23.27}\,/\,0.765\,/\,\tb{0.225} & \tb{23.25}\,/\,0.765\,/\,\tb{0.225} \\
\quad + best other & 21.85\,/\,0.759\,/\,0.284\textsuperscript{\,R} & 23.08\,/\,\tb{0.771}\,/\,0.236\textsuperscript{\,S} & 23.13\textsuperscript{\,S}\,/\,\tb{0.772\textsuperscript{\,R}}\,/\,\tb{0.234\textsuperscript{\,S}} & 22.40\,/\,0.757\,/\,0.259\textsuperscript{\,R} & 23.13\textsuperscript{\,S}\,/\,\tb{0.766\textsuperscript{\,R}}\,/\,0.226\textsuperscript{\,S} & 23.15\textsuperscript{\,S}\,/\,\tb{0.767\textsuperscript{\,R}}\,/\,\tb{0.225\textsuperscript{\,S}} \\
\cmidrule(lr){1-7}
ReSplat$^{\ast}$~\cite{resplat} + AnyPrune & \tb{23.13}\,/\,\tb{0.795}\,/\,\tb{0.345} & \tb{26.18}\,/\,\tb{0.860}\,/\,\tb{0.256} & \tb{26.62}\,/\,\tb{0.870}\,/\,\tb{0.236} & \tb{24.94}\,/\,\tb{0.832}\,/\,0.297 & \tb{27.94}\,/\,\tb{0.882}\,/\,\tb{0.227} & \tb{28.34}\,/\,\tb{0.889}\,/\,\tb{0.214} \\
\quad + best other & 22.21\,/\,0.779\,/\,0.355\textsuperscript{\,R} & 24.81\,/\,0.847\,/\,0.263\textsuperscript{\,R} & 25.82\textsuperscript{\,M}\,/\,0.861\textsuperscript{\,M}\,/\,0.245\textsuperscript{\,R} & 24.19\,/\,0.829\,/\,\tb{0.284}\textsuperscript{\,R} & 26.59\textsuperscript{\,M}\,/\,0.872\textsuperscript{\,M}\,/\,0.241\textsuperscript{\,R} & 27.57\,/\,0.882\,/\,0.226\textsuperscript{\,M} \\
\cmidrule(lr){1-7}
TokenGS~\cite{tokengs} + AnyPrune & \tb{26.99}\,/\,\tb{0.876}\,/\,\tb{0.220} & \tb{27.00}\,/\,\tb{0.876}\,/\,\tb{0.220} & \tb{27.00}\,/\,\tb{0.876}\,/\,\tb{0.220} & \tb{27.84}\,/\,\tb{0.883}\,/\,\tb{0.218} & \tb{27.84}\,/\,\tb{0.883}\,/\,\tb{0.217} & \tb{27.84}\,/\,\tb{0.883}\,/\,\tb{0.217} \\
\quad + best other & 25.91\textsuperscript{\,R}\,/\,0.870\textsuperscript{\,S}\,/\,0.227\textsuperscript{\,G} & 25.84\textsuperscript{\,R}\,/\,0.870\textsuperscript{\,F}\,/\,0.226\textsuperscript{\,F} & 25.85\textsuperscript{\,C}\,/\,0.870\textsuperscript{\,F}\,/\,0.226\textsuperscript{\,F} & 26.42\textsuperscript{\,R}\,/\,0.876\textsuperscript{\,S}\,/\,0.225\textsuperscript{\,G} & 26.32\textsuperscript{\,R}\,/\,0.876\textsuperscript{\,Rd}\,/\,0.224\textsuperscript{\,G} & 26.32\textsuperscript{\,R$'$}\,/\,0.876\textsuperscript{\,Rd}\,/\,0.224\textsuperscript{\,G} \\
\cmidrule(lr){1-7}
Off The Grid$^{\ast\dagger}$~\cite{off-the-grid} + AnyPrune & \tb{20.28}\,/\,\tb{0.709}\,/\,\tb{0.363} & \tb{21.66}\,/\,\tb{0.745}\,/\,\tb{0.300} & \tb{21.72}\,/\,\tb{0.747}\,/\,0.294 & \tb{22.17}\,/\,\tb{0.759}\,/\,\tb{0.281} & \tb{22.93}\,/\,\tb{0.773}\,/\,0.250 & \tb{22.94}\,/\,\tb{0.773}\,/\,0.249 \\
\quad + best other & 16.53\textsuperscript{\,R}\,/\,0.620\textsuperscript{\,M}\,/\,0.424\textsuperscript{\,M} & 20.97\,/\,0.731\,/\,0.305\textsuperscript{\,M} & 21.04\textsuperscript{\,M}\,/\,0.737\textsuperscript{\,M}\,/\,\tb{0.286\textsuperscript{\,C}} & 21.15\,/\,0.729\,/\,0.312\textsuperscript{\,M} & 22.74\textsuperscript{\,M}\,/\,0.771\textsuperscript{\,M}\,/\,\tb{0.246\textsuperscript{\,Rd}} & 22.70\,/\,0.770\,/\,\tb{0.242}\textsuperscript{\,M} \\
\cmidrule(lr){1-7}
EcoSplat$^{\ast}$~\cite{ecosplat} (50\%) + AnyPrune & \tb{22.13}\,/\,\tb{0.751}\,/\,\tb{0.375} & \tb{23.24}\,/\,\tb{0.777}\,/\,\tb{0.336} & \tb{23.36}\,/\,\tb{0.780}\,/\,\tb{0.329} & \tb{22.85}\,/\,\tb{0.766}\,/\,\tb{0.340} & \tb{23.55}\,/\,\tb{0.783}\,/\,\tb{0.307} & \tb{23.58}\,/\,\tb{0.784}\,/\,\tb{0.304} \\
\quad + best other & 16.75\,/\,0.614\,/\,0.451\textsuperscript{\,R} & 20.68\,/\,0.700\,/\,0.400\textsuperscript{\,S} & 21.24\,/\,0.722\,/\,0.375\textsuperscript{\,S} & 20.72\,/\,0.728\,/\,0.371\textsuperscript{\,R$'$} & 22.42\textsuperscript{\,M}\,/\,0.770\textsuperscript{\,G}\,/\,0.316\textsuperscript{\,G} & 22.43\textsuperscript{\,M}\,/\,0.771\textsuperscript{\,G}\,/\,0.313\textsuperscript{\,G} \\
\quad + own pruning & 18.81\,/\,0.666\,/\,0.412 & 20.09\,/\,0.695\,/\,0.391 & 20.65\,/\,0.708\,/\,0.381 & 21.56\,/\,0.753\,/\,0.342 & 22.05\,/\,0.765\,/\,0.324 & 22.19\,/\,0.768\,/\,0.319 \\
\cmidrule(lr){1-7}
F4Splat~\cite{f4splat} (50\%) + AnyPrune & 18.78\,/\,0.676\,/\,0.393 & \tb{23.33}\,/\,0.783\,/\,0.281 & \tb{23.40}\,/\,0.787\,/\,0.270 & 19.17\,/\,0.654\,/\,0.403 & \tb{23.40}\,/\,\tb{0.779}\,/\,0.246 & \tb{23.62}\,/\,\tb{0.787}\,/\,\tb{0.224} \\
\quad + best other & 15.25\textsuperscript{\,R}\,/\,0.551\textsuperscript{\,R}\,/\,0.456\textsuperscript{\,C} & 20.86\,/\,0.728\,/\,0.319\textsuperscript{\,C} & 22.97\textsuperscript{\,M}\,/\,0.774\textsuperscript{\,C}\,/\,0.261\textsuperscript{\,C} & 14.61\textsuperscript{\,R}\,/\,0.543\textsuperscript{\,L}\,/\,0.441\textsuperscript{\,L} & 19.11\textsuperscript{\,R}\,/\,0.688\textsuperscript{\,R}\,/\,0.326\textsuperscript{\,L} & 20.66\textsuperscript{\,S}\,/\,0.730\textsuperscript{\,L}\,/\,0.280\textsuperscript{\,L} \\
\quad + own pruning & \tb{22.78}\,/\,\tb{0.781}\,/\,\tb{0.264} & 23.15\,/\,\tb{0.791}\,/\,\tb{0.238} & 23.23\,/\,\tb{0.793}\,/\,\tb{0.232} & \tb{22.29}\,/\,\tb{0.754}\,/\,\tb{0.246} & 22.06\,/\,0.756\,/\,\tb{0.245} & 21.87\,/\,0.755\,/\,0.249 \\
\cmidrule(lr){1-7}
AdaptiveSplat$^{\ast}$~\cite{adaptivesplat} (60\%) + AnyPrune & \tb{14.42}\,/\,\tb{0.479}\,/\,\tb{0.566} & \tb{19.23}\,/\,\tb{0.628}\,/\,\tb{0.454} & \tb{20.15}\,/\,\tb{0.661}\,/\,\tb{0.417} & \tb{16.83}\,/\,\tb{0.561}\,/\,\tb{0.487} & \tb{20.45}\,/\,0.675\,/\,\tb{0.370} & \tb{20.67}\,/\,\tb{0.689}\,/\,\tb{0.349} \\
\quad + best other & 12.00\textsuperscript{\,R}\,/\,0.412\textsuperscript{\,L}\,/\,0.591\textsuperscript{\,L} & 16.50\textsuperscript{\,R}\,/\,0.546\textsuperscript{\,L}\,/\,0.494\textsuperscript{\,L} & 17.86\textsuperscript{\,R}\,/\,0.597\textsuperscript{\,R$'$}\,/\,0.456\textsuperscript{\,R$'$} & 14.14\textsuperscript{\,R}\,/\,0.516\textsuperscript{\,L}\,/\,0.505\textsuperscript{\,L} & 18.54\textsuperscript{\,R}\,/\,0.639\textsuperscript{\,R$'$}\,/\,0.398\textsuperscript{\,R$'$} & 19.46\textsuperscript{\,S}\,/\,0.666\textsuperscript{\,R$'$}\,/\,0.368\textsuperscript{\,R$'$} \\
\quad + own pruning & -- & 18.83\,/\,0.617\,/\,0.480 & 18.77\,/\,0.624\,/\,0.456 & -- & 19.89\,/\,\tb{0.677}\,/\,0.394 & 19.84\,/\,0.678\,/\,0.377 \\
\bottomrule
\end{tabular}
}
\end{table*}

\begin{table}[t]
\centering
\footnotesize
\setlength{\tabcolsep}{3pt}
\caption{Pruning cost on an RTX 4060 Ti (16\,GiB): seconds (peak GiB beyond the field and views) to prune YoNoSplat fields on DL3DV, median of 5 scenes. Baselines keep 30\%. OOM: does not fit in 16\,GiB.}
\label{tab:timing}
\adjustbox{max width=\columnwidth}{\begin{tabular}{lcccc}
\toprule
Views (Gaussians) & 2 (100k) & 8 (401k) & 16 (803k) & 32 (1.61M) \\
\midrule
YoNoSplat reconstruction & 0.14 & 0.41 & 0.84 & 1.75 \\
\midrule
Opacity                 & 0.001 (0.005) & 0.003 (0.02) & 0.005 (0.04) & 0.010 (0.08) \\
REFINE~\cite{refine}     & 0.001 (0.005) & 0.003 (0.03) & 0.007 (0.07) & 0.019 (0.16) \\
Speedy-Splat~\cite{speedy-splat} & 0.009 (0.04) & 0.14 (0.13) & 0.54 (0.25) & 1.61 (0.49) \\
LightGaussian~\cite{lightgaussian} & 0.13 (1.25) & 1.74 (3.84) & 6.92 (4.57) & 19.1 (3.60) \\
Provable coresets~\cite{coreset-pruning} & 1.09 (0.80) & 25.5 (2.45) & 215 (7.57) & OOM \\
NanoGS~\cite{nanogs}     & 1.00 (0.80) & 4.61 (1.07) & 9.66 (1.07) & 19.6 (1.08) \\
GHAP                    & 2.29 (0.02) & 3.67 (0.06) & 5.40 (0.11) & 10.4 (0.22) \\
Speedy-Splat + opt.     & 11.5 (0.24) & 22.3 (0.97) & 38.6 (1.07) & 68.2 (1.12) \\
\midrule
AnyPrune (ours)         & 0.72 (0.78) & 1.68 (2.21) & 2.50 (3.91) & 5.29 (7.30) \\
\quad sensitivity       & 0.010 & 0.14 & 0.54 & 1.60 \\
\quad render-error lift & 0.004 & 0.054 & 0.22 & 0.88 \\
\quad refiner + cut     & 0.67 & 1.43 & 1.79 & 2.60 \\
\bottomrule
\end{tabular}}
\end{table}

\Cref{tab:ablation-yonosplat} removes one design choice at a time and retrains AnyPrune on YoNoSplat~\cite{yonosplat} with the two-stage schedule using up to 8 views.
We train on 541 DL3DV scenes and test on 50 held-out scenes.
We average the PSNR over very low (10--20\%), low (30--50\%) and high (60--100\%) kept ratios separately, since some design choices help one range and hurt another.
The pruning score matters most: without the Speedy-Splat prior, the learned offset or the sparsity term, AnyPrune degrades sharply, most at low kept ratios.
The adjustments, the render error and the Objaverse~\cite{objaverse} initialization all help.
Switching the adjustments off at test time costs more than training without them, as in the second case the network is trained not to rely on them.
Even without the adjustments, the learned ranking beats cutting the same field by the Speedy-Splat score or by opacity at 10--50\% kept.
Training a single stage, or stopping after the first, helps at very low kept ratios but hurts elsewhere and makes the own cut far too aggressive, while skipping the first stage hurts at low kept ratios.
Also applying the residual scaling of \cref{eq:residual-scale} during training helps when few Gaussians are kept but hurts when most are kept, so we apply it at inference only.
Switching off the residual scaling of \cref{eq:residual-scale} helps at very low kept ratios but hurts at high ones.

\Cref{tab:ablation-recipe} compares the three-stage schedule with the two-stage one on YoNoSplat and Off The Grid.
Training on random cuts helps at heavy cuts but costs quality when most of the Gaussians are kept.
The other pipelines behave alike, except TokenGS, where the third stage gives a small gain at every kept ratio.
We therefore use it only where the two-stage model performs poorly at low kept ratios (\cref{sec:exp-setup}).

\section{Additional experiments}
\label{sec:suppl-experiments}

\subsection{Pruning results}
\label{sec:suppl-results}

\begin{figure*}[t]
\centering
\includegraphics[width=\textwidth]{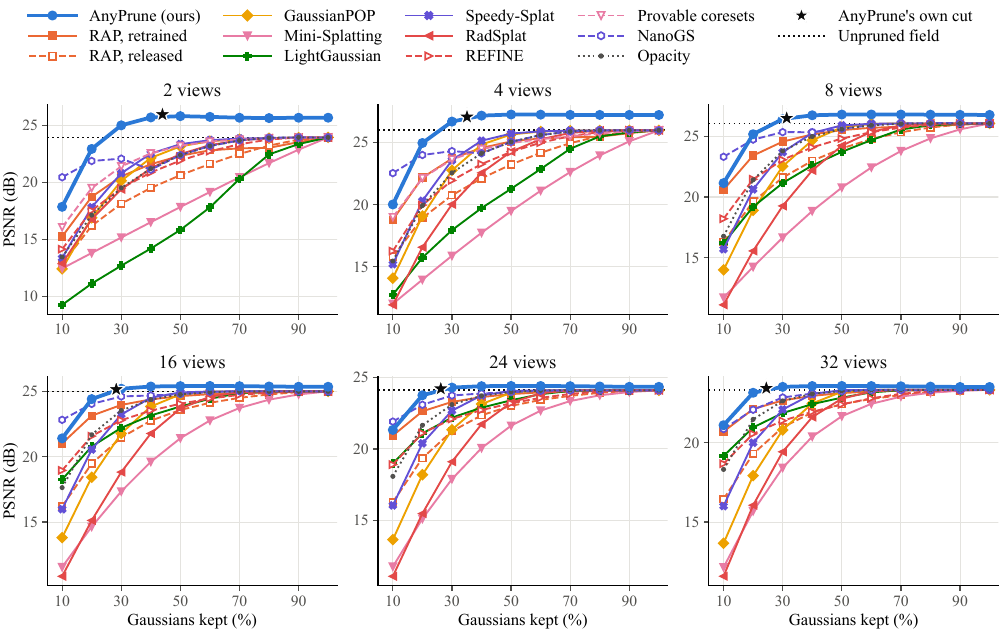}
\caption{\textbf{YoNoSplat on DL3DV:} novel-view PSNR against the share of Gaussians kept, for every pruning method we evaluate and every view count.}
\label{fig:suppl-curves-yonosplat}
\end{figure*}

\begin{figure*}[t]
\centering
\includegraphics[width=\textwidth]{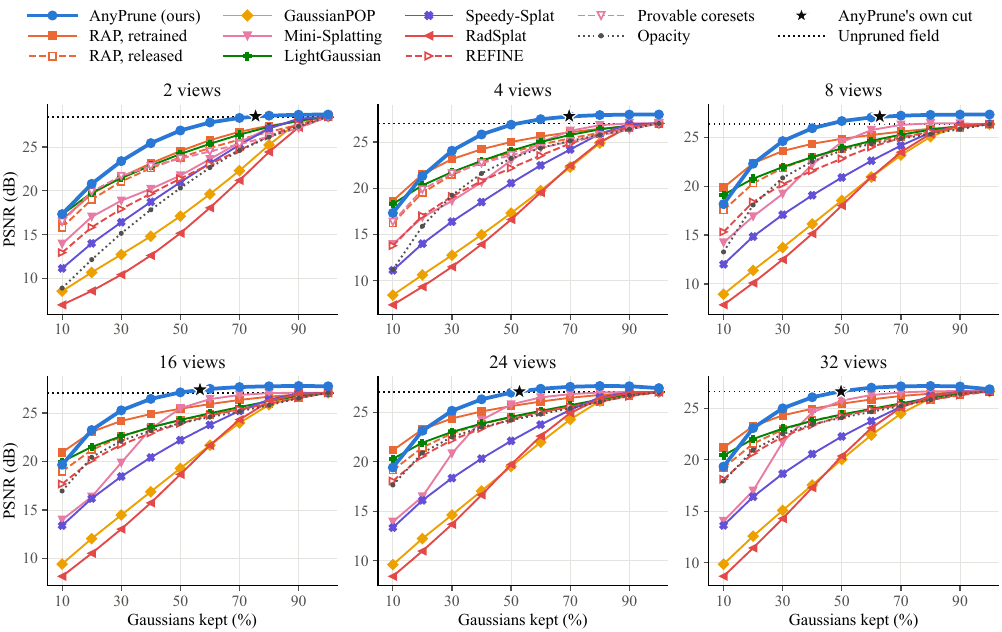}
\caption{\textbf{ReSplat on DL3DV:} novel-view PSNR against the share of Gaussians kept, for every pruning method we evaluate and every view count.}
\label{fig:suppl-curves-resplat}
\end{figure*}

\begin{figure*}[t]
\centering
\includegraphics[width=\textwidth]{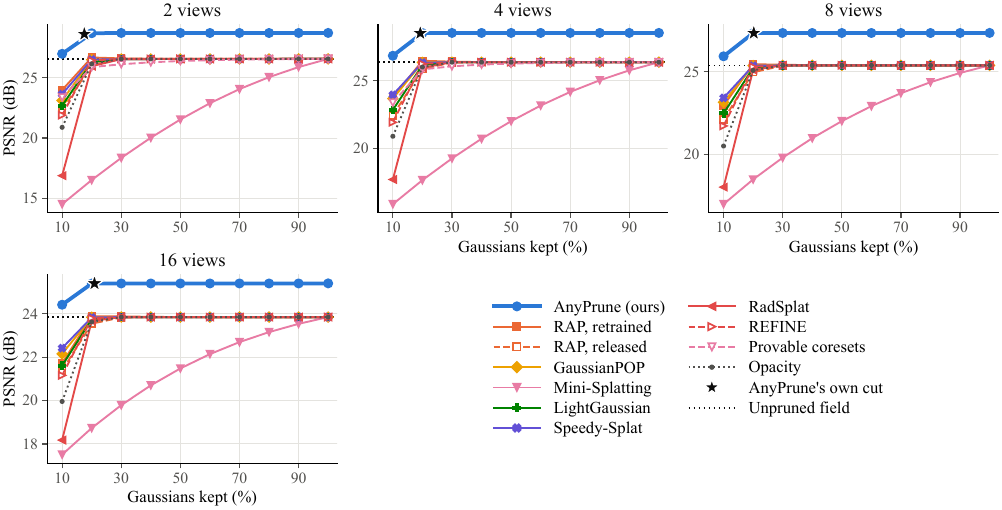}
\caption{\textbf{TokenGS on DL3DV:} novel-view PSNR against the share of Gaussians kept, for every pruning method we evaluate and every view count.}
\label{fig:suppl-curves-tokengs}
\end{figure*}

\begin{figure*}[t]
\centering
\includegraphics[width=\textwidth]{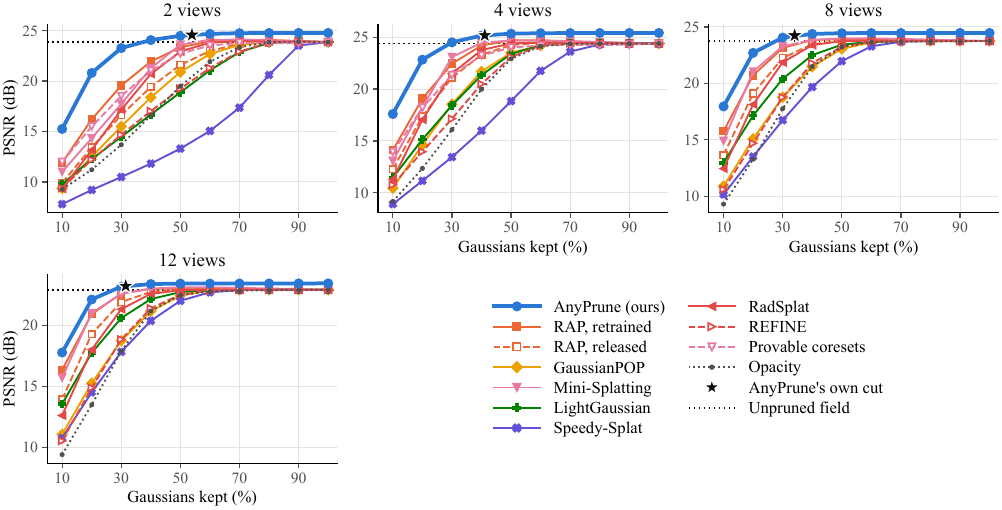}
\caption{\textbf{Off The Grid on DL3DV:} novel-view PSNR against the share of Gaussians kept, for every pruning method we evaluate and every view count.}
\label{fig:suppl-curves-otg}
\end{figure*}

\begin{figure*}[t]
\centering
\includegraphics[width=\textwidth]{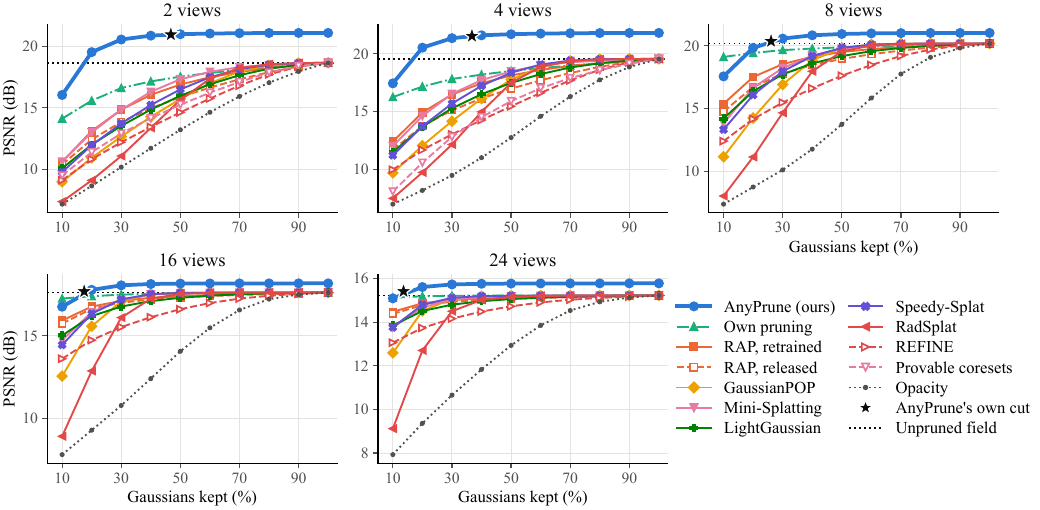}
\caption{\textbf{EcoSplat with its own pruning set to keep 50\% on DL3DV:} novel-view PSNR against the share of Gaussians kept, for every pruning method we evaluate and every view count. Kept ratios are relative to the reduced field.}
\label{fig:suppl-curves-eco-dial}
\end{figure*}

\begin{figure*}[t]
\centering
\includegraphics[width=\textwidth]{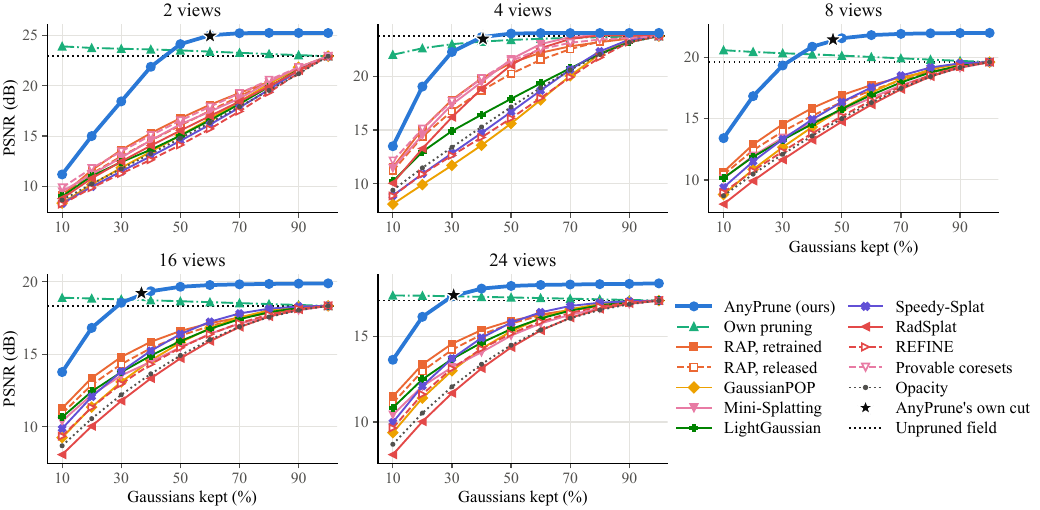}
\caption{\textbf{F4Splat with its own pruning set to keep 50\% on DL3DV:} novel-view PSNR against the share of Gaussians kept, for every pruning method we evaluate and every view count. Kept ratios are relative to the reduced field.}
\label{fig:suppl-curves-f4-dial}
\end{figure*}

\begin{figure*}[t]
\centering
\includegraphics[width=\textwidth]{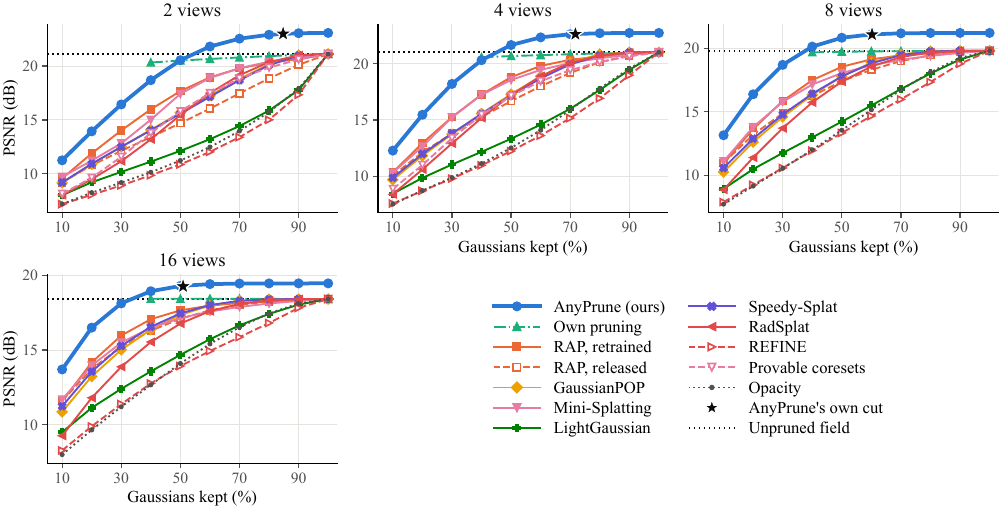}
\caption{\textbf{AdaptiveSplat with its own pruning set to keep 60\% on DL3DV:} novel-view PSNR against the share of Gaussians kept, for every pruning method we evaluate and every view count. Kept ratios are relative to the reduced field.}
\label{fig:suppl-curves-ada-dial}
\end{figure*}

\Cref{fig:suppl-curves-yonosplat,fig:suppl-curves-resplat,fig:suppl-curves-tokengs,fig:suppl-curves-otg,fig:suppl-curves-eco-dial,fig:suppl-curves-f4-dial,fig:suppl-curves-ada-dial} plot the PSNR of every pruning method against the share of Gaussians kept for each pipeline on DL3DV.
\Cref{tab:pruning-re10k,tab:pruning-acid} repeat the comparison of \cref{tab:pruning} on RealEstate10K and ACID without retraining AnyPrune, which is trained on DL3DV only; their other view counts are in \texttt{experiments.pdf}.
On all three datasets, AnyPrune gives the best PSNR in nearly every setting with 20--70\% kept, often improves the field when nothing is removed, and when keeping half of the Gaussians usually renders better than the unpruned field.

\Cref{tab:pruning-2432} extends \cref{tab:pruning} to 24 and 32 views, where AnyPrune keeps its lead on YoNoSplat, EcoSplat, F4Splat and AdaptiveSplat.
On ReSplat, it trails RAP at 10\% and 20\% kept and leads from 30\% on.

\begin{figure*}[t]
\centering
\includegraphics[width=\textwidth]{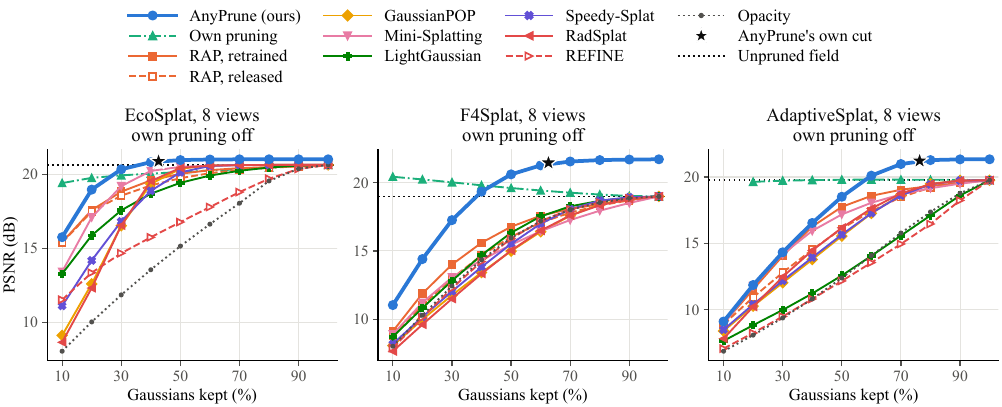}
\caption{\textbf{PSNR against the share of Gaussians kept at 8 views on DL3DV, with internal pruning disabled} (\cref{tab:pruning-dialoff}).
For the pipeline's own pruning, its internal pruning dial is turned back on to reach each kept ratio.}
\label{fig:curves8-dialoff}
\end{figure*}

In the main paper, EcoSplat, F4Splat and AdaptiveSplat run with their internal pruning dial halfway, the setting in which they would be deployed.
For completeness, \cref{tab:pruning-dialoff,fig:curves8-dialoff} train and test AnyPrune on the densest fields they predict with this dial off.
AnyPrune still leads in most settings, with smaller margins than on the reduced fields, and loses mainly on AdaptiveSplat at 2 and 4 views.

Since F4Splat's own pruning is ahead of AnyPrune at low kept ratios (\cref{sec:exp-dial}), \cref{tab:f4splat-dial20} trains a second AnyPrune model on the field F4Splat gives with its internal pruning dial at 20\%.
At the same number of Gaussians, it is ahead of F4Splat's own pruning, except at 2 views when keeping 10\%.
From 8 views on, part of this lead comes from the holes of \cref{sec:suppl-f4splat}.

\Cref{tab:own-cut-full} extends \cref{tab:own-cut} to every view count and to the pipelines with their internal pruning disabled.

\begin{table*}[t]
\centering
\caption{\textbf{Pruning quality on ACID at 4 and 16 input views,} with AnyPrune trained on DL3DV only.
Layout and notation as in \cref{tab:pruning}.
In parentheses, the share kept by the pipeline's internal pruning dial.}
\label{tab:pruning-acid}
\scriptsize
\setlength{\tabcolsep}{3pt}\renewcommand{\arraystretch}{0.9}
\adjustbox{max width=\textwidth}{\begin{tabular}{@{}l@{\hskip 5pt}ccc@{\hskip 5pt}ccc@{}}
\toprule
 & \multicolumn{3}{c}{4 views} & \multicolumn{3}{c}{16 views} \\
\cmidrule(lr){2-4} \cmidrule(lr){5-7}
Method & 20\% & 50\% & 70\% & 20\% & 50\% & 70\% \\
\midrule
YoNoSplat~\cite{yonosplat} + AnyPrune & \tb{24.24}\,/\,\tb{0.709}\,/\,\tb{0.341} & \tb{24.98}\,/\,\tb{0.723}\,/\,\tb{0.273} & \tb{24.89}\,/\,\tb{0.724}\,/\,\tb{0.271} & \tb{24.39}\,/\,\tb{0.708}\,/\,0.338 & \tb{24.59}\,/\,0.714\,/\,0.298 & \tb{24.58}\,/\,0.713\,/\,0.298 \\
\quad + best other & 22.20\textsuperscript{\,C}\,/\,0.689\textsuperscript{\,C}\,/\,0.347\textsuperscript{\,R} & 24.19\textsuperscript{\,C}\,/\,0.717\textsuperscript{\,C}\,/\,0.282\textsuperscript{\,S} & 24.42\textsuperscript{\,C}\,/\,0.719\textsuperscript{\,S}\,/\,0.275\textsuperscript{\,S} & 23.74\,/\,0.706\,/\,\tb{0.320}\textsuperscript{\,R} & 24.48\textsuperscript{\,S}\,/\,\tb{0.715\textsuperscript{\,R}}\,/\,\tb{0.291\textsuperscript{\,S}} & 24.51\textsuperscript{\,S}\,/\,\tb{0.714\textsuperscript{\,S}}\,/\,\tb{0.288\textsuperscript{\,M}} \\
\cmidrule(lr){1-7}
ReSplat$^{\ast}$~\cite{resplat} + AnyPrune & 22.58\,/\,0.729\,/\,\tb{0.386} & \tb{28.41}\,/\,\tb{0.850}\,/\,\tb{0.243} & \tb{29.47}\,/\,\tb{0.872}\,/\,\tb{0.208} & 23.65\,/\,0.750\,/\,0.340 & \tb{27.86}\,/\,\tb{0.831}\,/\,\tb{0.234} & \tb{28.47}\,/\,\tb{0.843}\,/\,\tb{0.216} \\
\quad + best other & \tb{23.66}\,/\,\tb{0.734}\,/\,\tb{0.386}\textsuperscript{\,R} & 27.55\,/\,0.840\,/\,0.250\textsuperscript{\,R} & 28.79\,/\,0.864\,/\,0.214\textsuperscript{\,R} & \tb{24.97}\,/\,\tb{0.757}\,/\,\tb{0.330}\textsuperscript{\,R} & 26.90\,/\,0.819\,/\,0.244\textsuperscript{\,R} & 28.01\,/\,0.838\,/\,0.217\textsuperscript{\,M} \\
\cmidrule(lr){1-7}
TokenGS~\cite{tokengs} + AnyPrune & \tb{29.14}\,/\,\tb{0.854}\,/\,\tb{0.264} & \tb{29.16}\,/\,\tb{0.854}\,/\,\tb{0.263} & \tb{29.16}\,/\,\tb{0.854}\,/\,\tb{0.263} & \tb{29.00}\,/\,\tb{0.848}\,/\,\tb{0.276} & \tb{29.02}\,/\,\tb{0.848}\,/\,\tb{0.275} & \tb{29.02}\,/\,\tb{0.848}\,/\,\tb{0.275} \\
\quad + best other & 27.50\textsuperscript{\,R}\,/\,0.844\textsuperscript{\,G}\,/\,0.273\textsuperscript{\,G} & 27.31\textsuperscript{\,R}\,/\,0.844\textsuperscript{\,S}\,/\,0.272\textsuperscript{\,S} & 27.31\textsuperscript{\,F}\,/\,0.844\textsuperscript{\,S}\,/\,0.272\textsuperscript{\,S} & 27.19\textsuperscript{\,R}\,/\,0.837\textsuperscript{\,G}\,/\,0.285\textsuperscript{\,G} & 27.00\textsuperscript{\,R}\,/\,0.837\textsuperscript{\,Rd}\,/\,0.285\textsuperscript{\,G} & 27.00\textsuperscript{\,G}\,/\,0.837\textsuperscript{\,Rd}\,/\,0.285\textsuperscript{\,G} \\
\cmidrule(lr){1-7}
Off The Grid$^{\ast\dagger}$~\cite{off-the-grid} + AnyPrune & \tb{23.31}\,/\,\tb{0.702}\,/\,\tb{0.355} & \tb{24.25}\,/\,\tb{0.719}\,/\,0.325 & \tb{24.27}\,/\,\tb{0.719}\,/\,0.323 & \tb{24.32}\,/\,\tb{0.726}\,/\,\tb{0.306} & \tb{24.85}\,/\,0.732\,/\,0.293 & \tb{24.84}\,/\,\tb{0.731}\,/\,0.294 \\
\quad + best other & 21.39\,/\,0.670\,/\,0.372\textsuperscript{\,M} & 24.15\textsuperscript{\,R}\,/\,0.717\textsuperscript{\,M}\,/\,\tb{0.295\textsuperscript{\,M}} & 24.07\,/\,0.716\,/\,\tb{0.292}\textsuperscript{\,M} & 23.90\,/\,0.714\,/\,0.312\textsuperscript{\,M} & 24.84\,/\,\tb{0.736}\,/\,\tb{0.270}\textsuperscript{\,M} & 24.70\,/\,\tb{0.731}\,/\,\tb{0.275}\textsuperscript{\,M} \\
\cmidrule(lr){1-7}
EcoSplat$^{\ast}$~\cite{ecosplat} (50\%) + AnyPrune & \tb{22.98}\,/\,\tb{0.686}\,/\,\tb{0.385} & \tb{24.97}\,/\,\tb{0.723}\,/\,\tb{0.330} & \tb{25.10}\,/\,\tb{0.727}\,/\,\tb{0.323} & \tb{23.25}\,/\,\tb{0.696}\,/\,\tb{0.365} & \tb{25.28}\,/\,\tb{0.727}\,/\,\tb{0.306} & \tb{25.30}\,/\,\tb{0.728}\,/\,\tb{0.304} \\
\quad + best other & 15.19\textsuperscript{\,R}\,/\,0.442\textsuperscript{\,L}\,/\,0.536\textsuperscript{\,L} & 19.69\textsuperscript{\,M}\,/\,0.561\textsuperscript{\,R}\,/\,0.458\textsuperscript{\,L} & 21.01\textsuperscript{\,Rd}\,/\,0.607\textsuperscript{\,G}\,/\,0.430\textsuperscript{\,L} & 20.84\,/\,0.638\,/\,0.425\textsuperscript{\,R} & 24.25\,/\,0.707\,/\,0.333\textsuperscript{\,G} & 24.29\,/\,0.709\,/\,0.326\textsuperscript{\,G} \\
\quad + own pruning & 17.25\,/\,0.496\,/\,0.501 & 19.44\,/\,0.562\,/\,0.452 & 20.28\,/\,0.589\,/\,0.433 & 22.87\,/\,0.670\,/\,0.387 & 23.93\,/\,0.697\,/\,0.349 & 24.13\,/\,0.703\,/\,0.337 \\
\cmidrule(lr){1-7}
F4Splat~\cite{f4splat} (50\%) + AnyPrune & 21.22\,/\,0.636\,/\,0.400 & \tb{25.32}\,/\,0.747\,/\,0.300 & \tb{25.44}\,/\,0.752\,/\,0.290 & 18.22\,/\,0.539\,/\,0.440 & \tb{24.38}\,/\,\tb{0.706}\,/\,\tb{0.290} & \tb{24.97}\,/\,\tb{0.723}\,/\,\tb{0.253} \\
\quad + best other & 16.96\,/\,0.513\,/\,0.432\textsuperscript{\,C} & 23.19\,/\,0.692\,/\,0.309\textsuperscript{\,C} & 25.25\textsuperscript{\,M}\,/\,0.738\textsuperscript{\,Rd}\,/\,0.262\textsuperscript{\,C} & 14.52\textsuperscript{\,R}\,/\,0.445\textsuperscript{\,L}\,/\,0.475\textsuperscript{\,L} & 19.12\textsuperscript{\,R}\,/\,0.596\textsuperscript{\,R}\,/\,0.371\textsuperscript{\,L} & 20.70\textsuperscript{\,R}\,/\,0.649\textsuperscript{\,L}\,/\,0.331\textsuperscript{\,L} \\
\quad + own pruning & \tb{23.99}\,/\,\tb{0.739}\,/\,\tb{0.300} & 24.72\,/\,\tb{0.757}\,/\,\tb{0.253} & 24.96\,/\,\tb{0.758}\,/\,\tb{0.242} & \tb{23.71}\,/\,\tb{0.686}\,/\,\tb{0.299} & 23.13\,/\,0.687\,/\,0.292 & 22.71\,/\,0.685\,/\,0.296 \\
\cmidrule(lr){1-7}
AdaptiveSplat$^{\ast}$~\cite{adaptivesplat} (60\%) + AnyPrune & \tb{16.32}\,/\,\tb{0.498}\,/\,\tb{0.538} & \tb{22.08}\,/\,\tb{0.638}\,/\,\tb{0.418} & \tb{23.32}\,/\,\tb{0.665}\,/\,\tb{0.380} & \tb{17.87}\,/\,\tb{0.532}\,/\,\tb{0.461} & \tb{22.40}\,/\,0.633\,/\,\tb{0.351} & \tb{22.69}\,/\,\tb{0.644}\,/\,\tb{0.335} \\
\quad + best other & 13.36\textsuperscript{\,R}\,/\,0.369\textsuperscript{\,R}\,/\,0.603\textsuperscript{\,L} & 18.61\textsuperscript{\,R}\,/\,0.526\textsuperscript{\,R}\,/\,0.495\textsuperscript{\,R$'$} & 20.64\textsuperscript{\,R}\,/\,0.593\textsuperscript{\,R}\,/\,0.446\textsuperscript{\,R$'$} & 14.91\,/\,0.476\,/\,0.510\textsuperscript{\,R$'$} & 20.22\textsuperscript{\,M}\,/\,0.598\textsuperscript{\,R$'$}\,/\,0.397\textsuperscript{\,R$'$} & 21.45\textsuperscript{\,R}\,/\,0.625\textsuperscript{\,R}\,/\,0.364\textsuperscript{\,R$'$} \\
\quad + own pruning & -- & 21.41\,/\,0.610\,/\,0.477 & 21.41\,/\,0.617\,/\,0.449 & -- & 21.88\,/\,\tb{0.638}\,/\,0.391 & 21.82\,/\,0.637\,/\,0.374 \\
\bottomrule
\end{tabular}
}
\end{table*}

\begin{table*}[t]
\centering
\caption{\textbf{Pruning quality on DL3DV at 24 and 32 input views.}
Layout as in \cref{tab:pruning}, with 10\%, 30\% and 100\% added.
In parentheses, the share kept by the pipeline's internal pruning dial; pipelines without one run with it off.}
\label{tab:pruning-2432}
\scriptsize
\setlength{\tabcolsep}{1.5pt}
\adjustbox{max width=\textwidth}{\begin{tabular}{@{}l@{\hskip 5pt}cccccc@{}}
\toprule
 & \multicolumn{6}{c}{24 views} \\
\cmidrule(lr){2-7}
Method & 10\% & 20\% & 30\% & 50\% & 70\% & 100\% \\
\midrule
YoNoSplat~\cite{yonosplat} + AnyPrune & \tb{21.33}\,/\,\tb{0.699}\,/\,\tb{0.344} & \tb{23.76}\,/\,\tb{0.752}\,/\,\tb{0.276} & \tb{24.29}\,/\,\tb{0.769}\,/\,\tb{0.245} & \tb{24.40}\,/\,\tb{0.776}\,/\,\tb{0.225} & \tb{24.39}\,/\,\tb{0.777}\,/\,\tb{0.222} & \tb{24.34}\,/\,\tb{0.777}\,/\,\tb{0.222} \\
\quad + best other & 20.93\,/\,0.691\,/\,\tb{0.344}\textsuperscript{\,R} & 22.63\,/\,0.737\,/\,0.284\textsuperscript{\,R} & 23.29\textsuperscript{\,R}\,/\,0.757\textsuperscript{\,S}\,/\,0.256\textsuperscript{\,S} & 24.06\,/\,0.771\,/\,0.228\textsuperscript{\,S} & 24.10\textsuperscript{\,S}\,/\,0.772\textsuperscript{\,S}\,/\,0.225\textsuperscript{\,G} & 24.10\,/\,0.772\,/\,0.225 \\
\cmidrule(lr){1-7}
ReSplat$^{\ast}$~\cite{resplat} + AnyPrune & 19.43\,/\,0.688\,/\,\tb{0.398} & 23.14\,/\,\tb{0.780}\,/\,0.308 & \tb{25.16}\,/\,\tb{0.825}\,/\,0.258 & \tb{27.00}\,/\,\tb{0.863}\,/\,\tb{0.209} & \tb{27.59}\,/\,\tb{0.874}\,/\,\tb{0.190} & \tb{27.45}\,/\,\tb{0.875}\,/\,\tb{0.188} \\
\quad + best other & \tb{21.17}\,/\,\tb{0.690}\,/\,0.406\textsuperscript{\,R} & \tb{23.28}\,/\,0.775\,/\,\tb{0.302}\textsuperscript{\,R} & 24.39\,/\,0.817\,/\,\tb{0.251}\textsuperscript{\,R} & 25.81\textsuperscript{\,M}\,/\,0.853\textsuperscript{\,M}\,/\,0.212\textsuperscript{\,R} & 26.85\,/\,0.866\,/\,0.198\textsuperscript{\,M} & 27.04\,/\,0.870\,/\,0.192 \\
\cmidrule(lr){1-7}
EcoSplat$^{\ast}$~\cite{ecosplat} + AnyPrune & \tb{14.60}\,/\,0.423\,/\,0.605 & \tb{15.56}\,/\,\tb{0.455}\,/\,0.587 & \tb{15.74}\,/\,\tb{0.466}\,/\,0.578 & \tb{15.78}\,/\,\tb{0.471}\,/\,0.575 & \tb{15.79}\,/\,\tb{0.472}\,/\,0.575 & \tb{15.79}\,/\,\tb{0.472}\,/\,\tb{0.575} \\
\quad + best other & 14.37\textsuperscript{\,R$'$}\,/\,\tb{0.433\textsuperscript{\,R}}\,/\,\tb{0.588\textsuperscript{\,R$'$}} & 14.86\textsuperscript{\,M}\,/\,0.442\textsuperscript{\,R}\,/\,\tb{0.581\textsuperscript{\,R$'$}} & 15.20\textsuperscript{\,M}\,/\,0.447\textsuperscript{\,R}\,/\,\tb{0.577\textsuperscript{\,R$'$}} & 15.33\textsuperscript{\,S}\,/\,0.454\textsuperscript{\,S}\,/\,\tb{0.573\textsuperscript{\,R$'$}} & 15.34\textsuperscript{\,S}\,/\,0.455\textsuperscript{\,S}\,/\,\tb{0.572\textsuperscript{\,R$'$}} & 15.35\,/\,0.455\,/\,0.578 \\
\cmidrule(lr){1-7}
EcoSplat$^{\ast}$~\cite{ecosplat} (50\%) + AnyPrune & 15.09\,/\,0.462\,/\,0.587 & \tb{15.61}\,/\,\tb{0.474}\,/\,0.576 & \tb{15.73}\,/\,\tb{0.477}\,/\,\tb{0.571} & \tb{15.77}\,/\,\tb{0.480}\,/\,\tb{0.567} & \tb{15.78}\,/\,\tb{0.480}\,/\,\tb{0.566} & \tb{15.78}\,/\,\tb{0.480}\,/\,\tb{0.566} \\
\quad + best other & 14.48\,/\,0.447\,/\,0.590\textsuperscript{\,R} & 14.87\,/\,0.453\,/\,0.583\textsuperscript{\,R} & 15.12\textsuperscript{\,S}\,/\,0.456\textsuperscript{\,R}\,/\,0.576\textsuperscript{\,L} & 15.21\,/\,0.458\,/\,0.571\textsuperscript{\,S} & 15.22\textsuperscript{\,S}\,/\,0.460\textsuperscript{\,R}\,/\,0.569\textsuperscript{\,L} & 15.23\,/\,0.459\,/\,0.570 \\
\quad + own pruning & \tb{15.12}\,/\,\tb{0.473}\,/\,\tb{0.581} & 15.19\,/\,0.472\,/\,\tb{0.575} & 15.21\,/\,0.470\,/\,0.572 & 15.22\,/\,0.467\,/\,0.570 & 15.23\,/\,0.464\,/\,0.570 & 15.23\,/\,0.459\,/\,0.570 \\
\cmidrule(lr){1-7}
F4Splat$^{\ast}$~\cite{f4splat} + AnyPrune & \tb{11.29}\,/\,\tb{0.276}\,/\,\tb{0.603} & \tb{14.20}\,/\,\tb{0.396}\,/\,\tb{0.528} & \tb{16.02}\,/\,\tb{0.464}\,/\,\tb{0.480} & \tb{17.54}\,/\,\tb{0.522}\,/\,\tb{0.429} & \tb{17.86}\,/\,\tb{0.536}\,/\,\tb{0.416} & \tb{17.97}\,/\,\tb{0.541}\,/\,\tb{0.415} \\
\quad + best other & 10.12\textsuperscript{\,R}\,/\,0.237\textsuperscript{\,L}\,/\,0.620\textsuperscript{\,R} & 12.55\textsuperscript{\,R}\,/\,0.335\textsuperscript{\,L}\,/\,0.557\textsuperscript{\,R} & 14.07\,/\,0.402\,/\,0.516\textsuperscript{\,R} & 15.68\textsuperscript{\,R}\,/\,0.474\textsuperscript{\,L}\,/\,0.464\textsuperscript{\,L} & 16.47\,/\,0.498\,/\,0.442\textsuperscript{\,L} & 16.82\,/\,0.508\,/\,0.438 \\
\cmidrule(lr){1-7}
F4Splat~\cite{f4splat} (50\%) + AnyPrune & 13.64\,/\,0.383\,/\,0.557 & 16.15\,/\,0.469\,/\,0.501 & 17.35\,/\,0.510\,/\,0.465 & \tb{17.96}\,/\,\tb{0.539}\,/\,\tb{0.423} & \tb{18.04}\,/\,\tb{0.543}\,/\,\tb{0.412} & \tb{18.11}\,/\,\tb{0.546}\,/\,\tb{0.411} \\
\quad + best other & 11.49\textsuperscript{\,R}\,/\,0.319\textsuperscript{\,L}\,/\,0.585\textsuperscript{\,R} & 13.37\textsuperscript{\,R}\,/\,0.382\textsuperscript{\,L}\,/\,0.538\textsuperscript{\,R} & 14.58\textsuperscript{\,R}\,/\,0.421\textsuperscript{\,L}\,/\,0.509\textsuperscript{\,R} & 15.92\textsuperscript{\,R}\,/\,0.472\textsuperscript{\,L}\,/\,0.470\textsuperscript{\,R} & 16.78\textsuperscript{\,S}\,/\,0.501\textsuperscript{\,L}\,/\,0.447\textsuperscript{\,L} & 17.10\,/\,0.515\,/\,0.431 \\
\quad + own pruning & \tb{17.40}\,/\,\tb{0.511}\,/\,\tb{0.440} & \tb{17.39}\,/\,\tb{0.515}\,/\,\tb{0.436} & \tb{17.35}\,/\,\tb{0.516}\,/\,\tb{0.432} & 17.28\,/\,0.516\,/\,0.429 & 17.21\,/\,0.516\,/\,0.430 & 17.10\,/\,0.515\,/\,0.431 \\
\cmidrule(lr){1-7}
AdaptiveSplat~\cite{adaptivesplat} + AnyPrune & \tb{10.00}\,/\,0.227\,/\,0.623 & \tb{12.71}\,/\,0.315\,/\,\tb{0.561} & \tb{14.77}\,/\,\tb{0.388}\,/\,0.509 & \tb{17.49}\,/\,\tb{0.487}\,/\,\tb{0.431} & \tb{18.53}\,/\,\tb{0.529}\,/\,\tb{0.390} & \tb{18.65}\,/\,\tb{0.537}\,/\,\tb{0.381} \\
\quad + best other & 9.94\,/\,\tb{0.253}\,/\,\tb{0.617}\textsuperscript{\,M} & 12.45\textsuperscript{\,M}\,/\,\tb{0.330\textsuperscript{\,M}}\,/\,\tb{0.561\textsuperscript{\,R$'$}} & 14.51\textsuperscript{\,R}\,/\,0.385\textsuperscript{\,R$'$}\,/\,\tb{0.508\textsuperscript{\,R$'$}} & 16.71\,/\,0.474\,/\,\tb{0.431}\textsuperscript{\,R} & 17.36\textsuperscript{\,G}\,/\,0.506\textsuperscript{\,Rd}\,/\,0.406\textsuperscript{\,R} & 17.75\,/\,0.520\,/\,0.397 \\
\midrule[\heavyrulewidth]
 & \multicolumn{6}{c}{32 views} \\
\cmidrule(lr){2-7}
Method & 10\% & 20\% & 30\% & 50\% & 70\% & 100\% \\
\midrule
YoNoSplat~\cite{yonosplat} + AnyPrune & \tb{21.11}\,/\,\tb{0.690}\,/\,\tb{0.344} & \tb{23.16}\,/\,\tb{0.733}\,/\,\tb{0.284} & \tb{23.53}\,/\,\tb{0.746}\,/\,\tb{0.259} & \tb{23.59}\,/\,\tb{0.750}\,/\,\tb{0.245} & \tb{23.57}\,/\,\tb{0.751}\,/\,\tb{0.244} & \tb{23.52}\,/\,\tb{0.750}\,/\,\tb{0.244} \\
\quad + best other & 20.71\,/\,0.680\,/\,0.347\textsuperscript{\,R} & 22.13\,/\,0.719\,/\,0.294\textsuperscript{\,R} & 22.68\textsuperscript{\,R}\,/\,0.735\textsuperscript{\,S}\,/\,0.270\textsuperscript{\,S} & 23.32\,/\,0.747\,/\,0.247\textsuperscript{\,S} & 23.34\textsuperscript{\,S}\,/\,0.747\textsuperscript{\,S}\,/\,0.246\textsuperscript{\,G} & 23.34\,/\,0.747\,/\,0.246 \\
\cmidrule(lr){1-7}
ReSplat$^{\ast}$~\cite{resplat} + AnyPrune & 19.34\,/\,0.689\,/\,\tb{0.396} & 23.08\,/\,\tb{0.780}\,/\,0.307 & \tb{25.00}\,/\,\tb{0.823}\,/\,0.260 & \tb{26.67}\,/\,\tb{0.858}\,/\,\tb{0.215} & \tb{27.13}\,/\,\tb{0.867}\,/\,\tb{0.199} & \tb{26.85}\,/\,\tb{0.866}\,/\,\tb{0.200} \\
\quad + best other & \tb{21.19}\,/\,\tb{0.693}\,/\,0.402\textsuperscript{\,R} & \tb{23.27}\,/\,0.776\,/\,\tb{0.300}\textsuperscript{\,R} & 24.32\,/\,0.816\,/\,\tb{0.251}\textsuperscript{\,R} & 25.75\textsuperscript{\,M}\,/\,0.850\textsuperscript{\,M}\,/\,0.216\textsuperscript{\,R} & 26.54\,/\,0.860\,/\,0.206\textsuperscript{\,M} & 26.63\,/\,0.863\,/\,0.202 \\
\cmidrule(lr){1-7}
AdaptiveSplat~\cite{adaptivesplat} + AnyPrune & \tb{10.21}\,/\,0.227\,/\,0.628 & \tb{12.88}\,/\,0.309\,/\,0.570 & \tb{14.83}\,/\,0.377\,/\,0.522 & \tb{17.17}\,/\,\tb{0.467}\,/\,0.454 & \tb{17.94}\,/\,\tb{0.503}\,/\,\tb{0.420} & \tb{18.03}\,/\,\tb{0.510}\,/\,\tb{0.414} \\
\quad + best other & 10.15\,/\,\tb{0.253}\,/\,\tb{0.622}\textsuperscript{\,M} & 12.62\textsuperscript{\,M}\,/\,\tb{0.323\textsuperscript{\,M}}\,/\,\tb{0.567\textsuperscript{\,R$'$}} & 14.56\textsuperscript{\,R}\,/\,\tb{0.378\textsuperscript{\,R$'$}}\,/\,\tb{0.518\textsuperscript{\,R$'$}} & 16.48\,/\,0.455\,/\,\tb{0.452}\textsuperscript{\,R} & 17.04\textsuperscript{\,G}\,/\,0.484\textsuperscript{\,Rd}\,/\,0.433\textsuperscript{\,R} & 17.33\,/\,0.495\,/\,0.427 \\
\bottomrule
\end{tabular}
}
\end{table*}

\begin{table*}[t]
\centering
\caption{\textbf{Pruning quality on DL3DV with internal pruning disabled.}
EcoSplat, F4Splat and AdaptiveSplat with their internal pruning dial off, and AnyPrune trained on these fields. Layout as in \cref{tab:pruning}.}
\label{tab:pruning-dialoff}
\scriptsize
\setlength{\tabcolsep}{3pt}\renewcommand{\arraystretch}{0.9}
\adjustbox{max width=\textwidth}{\begin{tabular}{@{}l@{\hskip 5pt}ccc@{\hskip 5pt}ccc@{}}
\toprule
 & \multicolumn{3}{c}{2 views} & \multicolumn{3}{c}{4 views} \\
\cmidrule(lr){2-4} \cmidrule(lr){5-7}
Method & 20\% & 50\% & 70\% & 20\% & 50\% & 70\% \\
\midrule
EcoSplat$^{\ast}$~\cite{ecosplat} + AnyPrune & \tb{18.13}\,/\,\tb{0.557}\,/\,\tb{0.498} & \tb{20.97}\,/\,\tb{0.641}\,/\,\tb{0.416} & \tb{21.22}\,/\,\tb{0.652}\,/\,\tb{0.397} & \tb{18.91}\,/\,\tb{0.572}\,/\,\tb{0.458} & \tb{21.76}\,/\,\tb{0.664}\,/\,\tb{0.361} & \tb{21.92}\,/\,\tb{0.672}\,/\,\tb{0.345} \\
\quad + best other & 13.00\,/\,0.399\,/\,0.554\textsuperscript{\,R} & 17.99\textsuperscript{\,M}\,/\,0.574\textsuperscript{\,R}\,/\,0.455\textsuperscript{\,R} & 20.08\textsuperscript{\,M}\,/\,0.618\textsuperscript{\,R}\,/\,0.410\textsuperscript{\,L} & 15.19\,/\,0.482\,/\,0.501\textsuperscript{\,R} & 20.56\textsuperscript{\,M}\,/\,0.617\textsuperscript{\,R}\,/\,0.404\textsuperscript{\,R$'$} & 21.37\,/\,0.660\,/\,0.346\textsuperscript{\,Rd} \\
\cmidrule(lr){1-7}
F4Splat$^{\ast}$~\cite{f4splat} + AnyPrune & \tb{14.93}\,/\,\tb{0.483}\,/\,\tb{0.527} & \tb{20.83}\,/\,\tb{0.653}\,/\,0.420 & \tb{22.37}\,/\,\tb{0.695}\,/\,0.379 & \tb{16.14}\,/\,\tb{0.509}\,/\,\tb{0.482} & \tb{21.95}\,/\,\tb{0.681}\,/\,0.361 & 23.17\,/\,0.718\,/\,0.324 \\
\quad + best other & 10.04\textsuperscript{\,C}\,/\,0.254\textsuperscript{\,L}\,/\,0.560\textsuperscript{\,S} & 16.11\,/\,0.587\,/\,\tb{0.357}\textsuperscript{\,R} & 18.84\textsuperscript{\,R}\,/\,0.686\textsuperscript{\,R}\,/\,\tb{0.268\textsuperscript{\,L}} & 12.96\textsuperscript{\,C}\,/\,0.411\textsuperscript{\,M}\,/\,0.483\textsuperscript{\,M} & 20.12\textsuperscript{\,M}\,/\,0.648\textsuperscript{\,Rd}\,/\,\tb{0.320\textsuperscript{\,R}} & \tb{23.37\textsuperscript{\,M}}\,/\,\tb{0.738\textsuperscript{\,R}}\,/\,\tb{0.232\textsuperscript{\,R}} \\
\cmidrule(lr){1-7}
AdaptiveSplat~\cite{adaptivesplat} + AnyPrune & 9.45\,/\,\tb{0.246}\,/\,0.616 & 15.52\,/\,0.538\,/\,0.450 & \tb{20.54}\,/\,\tb{0.686}\,/\,\tb{0.329} & \tb{10.75}\,/\,0.307\,/\,0.579 & \tb{17.84}\,/\,\tb{0.585}\,/\,\tb{0.392} & \tb{22.03}\,/\,\tb{0.693}\,/\,\tb{0.285} \\
\quad + best other & \tb{9.83\textsuperscript{\,R}}\,/\,0.226\textsuperscript{\,G}\,/\,\tb{0.593\textsuperscript{\,S}} & \tb{16.42\textsuperscript{\,R}}\,/\,\tb{0.581\textsuperscript{\,M}}\,/\,\tb{0.414\textsuperscript{\,M}} & 19.14\textsuperscript{\,M}\,/\,0.669\textsuperscript{\,R}\,/\,0.331\textsuperscript{\,R} & 10.71\textsuperscript{\,R}\,/\,\tb{0.343\textsuperscript{\,M}}\,/\,\tb{0.562\textsuperscript{\,M}} & 17.75\,/\,0.574\,/\,0.394\textsuperscript{\,R} & 19.87\,/\,0.670\,/\,0.296\textsuperscript{\,R} \\
\midrule[\heavyrulewidth]
 & \multicolumn{3}{c}{8 views} & \multicolumn{3}{c}{16 views} \\
\cmidrule(lr){2-4} \cmidrule(lr){5-7}
Method & 20\% & 50\% & 70\% & 20\% & 50\% & 70\% \\
\midrule
EcoSplat$^{\ast}$~\cite{ecosplat} + AnyPrune & \tb{18.97}\,/\,\tb{0.574}\,/\,\tb{0.463} & \tb{20.97}\,/\,\tb{0.642}\,/\,\tb{0.384} & \tb{21.02}\,/\,\tb{0.645}\,/\,0.378 & \tb{17.60}\,/\,\tb{0.520}\,/\,0.518 & \tb{18.20}\,/\,\tb{0.546}\,/\,\tb{0.489} & \tb{18.20}\,/\,\tb{0.547}\,/\,0.488 \\
\quad + best other & 17.59\textsuperscript{\,R}\,/\,0.554\textsuperscript{\,R}\,/\,0.465\textsuperscript{\,R$'$} & 20.51\,/\,0.628\,/\,0.397\textsuperscript{\,M} & 20.63\textsuperscript{\,G}\,/\,0.638\textsuperscript{\,G}\,/\,\tb{0.374\textsuperscript{\,Rd}} & 16.78\textsuperscript{\,M}\,/\,0.510\textsuperscript{\,R$'$}\,/\,\tb{0.514\textsuperscript{\,R$'$}} & 17.77\textsuperscript{\,S}\,/\,0.532\textsuperscript{\,G}\,/\,0.491\textsuperscript{\,R$'$} & 17.79\textsuperscript{\,M}\,/\,0.535\textsuperscript{\,S}\,/\,\tb{0.485\textsuperscript{\,R$'$}} \\
\cmidrule(lr){1-7}
F4Splat$^{\ast}$~\cite{f4splat} + AnyPrune & \tb{14.40}\,/\,\tb{0.437}\,/\,\tb{0.492} & \tb{20.60}\,/\,\tb{0.633}\,/\,\tb{0.330} & \tb{21.54}\,/\,\tb{0.660}\,/\,\tb{0.292} & \tb{14.52}\,/\,\tb{0.416}\,/\,\tb{0.502} & \tb{19.08}\,/\,\tb{0.571}\,/\,\tb{0.371} & \tb{19.56}\,/\,\tb{0.588}\,/\,\tb{0.350} \\
\quad + best other & 11.87\textsuperscript{\,R}\,/\,0.318\textsuperscript{\,L}\,/\,0.540\textsuperscript{\,R} & 16.77\textsuperscript{\,R}\,/\,0.545\textsuperscript{\,L}\,/\,0.382\textsuperscript{\,L} & 18.26\,/\,0.600\,/\,0.334\textsuperscript{\,L} & 12.51\textsuperscript{\,R}\,/\,0.335\textsuperscript{\,L}\,/\,0.538\textsuperscript{\,R} & 16.38\textsuperscript{\,R}\,/\,0.509\textsuperscript{\,L}\,/\,0.418\textsuperscript{\,L} & 17.44\,/\,0.542\,/\,0.386\textsuperscript{\,L} \\
\cmidrule(lr){1-7}
AdaptiveSplat~\cite{adaptivesplat} + AnyPrune & \tb{11.86}\,/\,0.325\,/\,0.555 & \tb{18.51}\,/\,\tb{0.564}\,/\,\tb{0.376} & \tb{20.98}\,/\,\tb{0.635}\,/\,\tb{0.299} & \tb{12.42}\,/\,0.318\,/\,\tb{0.554} & \tb{17.86}\,/\,\tb{0.512}\,/\,\tb{0.404} & \tb{19.31}\,/\,\tb{0.562}\,/\,\tb{0.353} \\
\quad + best other & 11.77\,/\,\tb{0.362}\,/\,\tb{0.538}\textsuperscript{\,M} & 17.75\,/\,0.552\,/\,0.379\textsuperscript{\,R} & 19.04\,/\,0.607\,/\,0.318\textsuperscript{\,R} & 12.24\textsuperscript{\,M}\,/\,\tb{0.339\textsuperscript{\,M}}\,/\,0.558\textsuperscript{\,R$'$} & 16.98\,/\,0.498\,/\,0.406\textsuperscript{\,R} & 17.79\textsuperscript{\,G}\,/\,0.536\textsuperscript{\,Rd}\,/\,0.372\textsuperscript{\,R} \\
\bottomrule
\end{tabular}
}
\end{table*}

\begin{table*}[t]
\centering
\caption{\textbf{F4Splat with its internal pruning dial at 20\% on DL3DV.}
Keeping 10\% and 20\% of F4Splat's densest field with its own pruning, the AnyPrune model of the main paper (internal pruning dial at 50\%), and AnyPrune trained on the field at 20\%, which at 20\% only adjusts it.
Best highlighted. From 8 views on, F4Splat's fields have holes at our resolution (\cref{sec:suppl-f4splat}).}
\label{tab:f4splat-dial20}
\scriptsize
\setlength{\tabcolsep}{3pt}
\adjustbox{max width=\textwidth}{\begin{tabular}{@{}lcccccc@{}}
\toprule
 & \multicolumn{2}{c}{2 views} & \multicolumn{2}{c}{4 views} & \multicolumn{2}{c}{8 views} \\
\cmidrule(l){2-3} \cmidrule(l){4-5} \cmidrule(l){6-7}
Method & 10\% & 20\% & 10\% & 20\% & 10\% & 20\% \\
\midrule
Own pruning & \tb{23.77}\,/\,\tb{0.755}\,/\,\tb{0.251} & 23.62\,/\,0.775\,/\,\tb{0.219} & 22.61\,/\,0.731\,/\,\tb{0.294} & 23.23\,/\,\tb{0.755}\,/\,\tb{0.250} & 20.43\,/\,0.623\,/\,0.321 & 20.23\,/\,0.627\,/\,\tb{0.308} \\
AnyPrune, dial at 50\% & 15.00\,/\,0.497\,/\,0.463 & 21.89\,/\,0.710\,/\,0.338 & 19.04\,/\,0.614\,/\,0.404 & 23.64\,/\,0.738\,/\,0.304 & 16.83\,/\,0.516\,/\,0.442 & 20.86\,/\,0.638\,/\,0.341 \\
AnyPrune, dial at 20\% & 22.19\,/\,0.720\,/\,0.342 & \tb{24.85}\,/\,\tb{0.782}\,/\,0.278 & \tb{23.63}\,/\,\tb{0.737}\,/\,0.319 & \tb{23.72}\,/\,0.740\,/\,0.308 & \tb{21.77}\,/\,\tb{0.667}\,/\,\tb{0.320} & \tb{21.91}\,/\,\tb{0.671}\,/\,0.309 \\
\midrule[\heavyrulewidth]
 & \multicolumn{2}{c}{16 views} & \multicolumn{2}{c}{24 views} \\
\cmidrule(l){2-3} \cmidrule(l){4-5}
Method & 10\% & 20\% & 10\% & 20\% \\
\midrule
Own pruning & 18.84\,/\,0.561\,/\,0.377 & 18.72\,/\,0.562\,/\,0.369 & 17.39\,/\,0.515\,/\,0.436 & 17.32\,/\,0.516\,/\,0.430 &  &  \\
AnyPrune, dial at 50\% & 16.82\,/\,0.494\,/\,0.467 & 19.34\,/\,0.578\,/\,0.385 & 16.15\,/\,0.469\,/\,0.501 & 17.80\,/\,0.530\,/\,0.439 &  &  \\
AnyPrune, dial at 20\% & \tb{19.82}\,/\,\tb{0.597}\,/\,\tb{0.372} & \tb{19.97}\,/\,\tb{0.603}\,/\,\tb{0.364} & \tb{18.08}\,/\,\tb{0.544}\,/\,\tb{0.432} & \tb{18.20}\,/\,\tb{0.550}\,/\,\tb{0.427} &  &  \\
\bottomrule
\end{tabular}
}
\end{table*}

\begin{table*}[t]
\centering
\caption{\textbf{Transfer across pipelines.}
PSNR on ReSplat's fields at 8 views of AnyPrune trained on ReSplat and of the two ablation models trained on YoNoSplat.
Ranking only: keep logits on the unadjusted field. Own cut: PSNR and share kept. Bold: best.}
\label{tab:transfer}
\small
\setlength{\tabcolsep}{4pt}
\begin{tabular}{@{}lccccc@{}}
\toprule
Kept & 20\% & 50\% & 70\% & 100\% & Own cut \\
\midrule
Unpruned field & -- & -- & -- & 25.64 & -- \\
\midrule
Opacity & 17.99 & 22.93 & 24.16 & -- & -- \\
Speedy-Splat & 13.14 & 18.86 & 22.85 & -- & -- \\
\midrule
\multicolumn{6}{@{}l}{\textit{Trained on ReSplat}} \\
AnyPrune & \textbf{21.45} & \textbf{26.71} & \textbf{27.26} & \textbf{27.36} & \textbf{27.03 (59\%)} \\
\quad ranking only & 21.22 & 25.27 & 25.59 & -- & -- \\
\midrule
\multicolumn{6}{@{}l}{\textit{Trained on YoNoSplat}} \\
AnyPrune, seed 1 & 14.81 & 21.00 & 23.69 & 24.50 & 23.76 (75\%) \\
\quad ranking only & 14.67 & 20.89 & 24.13 & -- & -- \\
AnyPrune, seed 2 & 15.28 & 21.69 & 24.02 & 24.50 & 24.13 (76\%) \\
\quad ranking only & 15.07 & 21.55 & 24.50 & -- & -- \\
\bottomrule
\end{tabular}

\end{table*}

\suppltop{\begin{table*}[t]
\centering
\caption{\textbf{Pruning ratio suggested by AnyPrune at every view count.}
\Cref{tab:own-cut} extended to every view count and to the internal pruning dial off; notation otherwise as in \cref{tab:pruning}.
In parentheses, the setting of the pipeline's internal pruning dial: the share of its output it keeps, or off.}
\label{tab:own-cut-full}
\small\renewcommand{\arraystretch}{0.92}
\setlength{\tabcolsep}{4pt}
\adjustbox{max width=\textwidth}{\begin{tabular}{@{}l@{\hskip 6pt}rrrr@{\hskip 6pt}rrrr@{\hskip 6pt}rrrr@{}}
\toprule
 & \multicolumn{4}{c}{2 views} & \multicolumn{4}{c}{4 views} & \multicolumn{4}{c}{8 views} \\
\cmidrule(lr){2-5} \cmidrule(lr){6-9} \cmidrule(lr){10-13}
 & Kept & PSNR & SSIM & LPIPS & Kept & PSNR & SSIM & LPIPS & Kept & PSNR & SSIM & LPIPS \\
\midrule
YoNoSplat & 44\% & +2.01 & +0.007 & +0.023 & 35\% & +1.07 & +0.000 & +0.041 & 31\% & +0.43 & -0.004 & +0.043 \\
ReSplat$^{\ast}$ & 75\% & +0.07 & -0.001 & +0.020 & 70\% & +0.78 & +0.005 & +0.010 & 63\% & +0.79 & +0.005 & +0.009 \\
TokenGS & 18\% & +2.05 & +0.009 & -0.005 & 19\% & +2.14 & +0.011 & -0.009 & 20\% & +1.96 & +0.011 & -0.009 \\
Off The Grid$^{\ast\dagger}$ & 54\% & +0.70 & -0.006 & +0.076 & 41\% & +0.79 & +0.004 & +0.047 & 34\% & +0.49 & +0.007 & +0.028 \\
EcoSplat$^{\ast}$ (off) & 75\% & +0.54 & +0.004 & +0.027 & 60\% & +0.42 & +0.004 & +0.018 & 43\% & +0.25 & -0.001 & +0.018 \\
EcoSplat$^{\ast}$ (50\%) & 47\% & +2.29 & +0.086 & -0.015 & 37\% & +1.97 & +0.069 & -0.020 & 26\% & +0.19 & +0.010 & +0.020 \\
F4Splat$^{\ast}$ (off) & 89\% & +0.85 & -0.060 & +0.170 & 65\% & -1.06 & -0.061 & +0.143 & 63\% & +2.47 & +0.039 & -0.023 \\
F4Splat (50\%) & 60\% & +2.01 & +0.003 & +0.083 & 40\% & -0.26 & -0.030 & +0.092 & 47\% & +1.83 & +0.032 & +0.006 \\
AdaptiveSplat (off) & 90\% & +2.34 & +0.015 & -0.004 & 83\% & +1.96 & +0.015 & -0.008 & 76\% & +1.46 & +0.015 & -0.009 \\
AdaptiveSplat$^{\ast}$ (60\%) & 85\% & +1.87 & +0.024 & +0.014 & 72\% & +1.61 & +0.021 & +0.003 & 60\% & +1.27 & +0.018 & +0.001 \\
\midrule
 & \multicolumn{4}{c}{16 views} & \multicolumn{4}{c}{24 views} & \multicolumn{4}{c}{32 views} \\
\cmidrule(lr){2-5} \cmidrule(lr){6-9} \cmidrule(lr){10-13}
 & Kept & PSNR & SSIM & LPIPS & Kept & PSNR & SSIM & LPIPS & Kept & PSNR & SSIM & LPIPS \\
\midrule
YoNoSplat & 28\% & +0.15 & -0.006 & +0.036 & 26\% & +0.11 & -0.006 & +0.030 & 24\% & +0.10 & -0.005 & +0.025 \\
ReSplat$^{\ast}$ & 57\% & +0.33 & -0.001 & +0.015 & 53\% & +0.07 & -0.003 & +0.014 & 50\% & -0.01 & -0.004 & +0.014 \\
TokenGS & 21\% & +1.55 & +0.011 & -0.009 & \multicolumn{4}{c}{--} & \multicolumn{4}{c}{--} \\
Off The Grid$^{\ast\dagger}$ & 31\% & +0.30 & +0.005 & +0.022 & \multicolumn{4}{c}{--} & \multicolumn{4}{c}{--} \\
EcoSplat$^{\ast}$ (off) & 29\% & +0.27 & +0.002 & +0.010 & 23\% & +0.29 & +0.004 & +0.006 & \multicolumn{4}{c}{--} \\
EcoSplat$^{\ast}$ (50\%) & 17\% & +0.07 & +0.004 & +0.019 & 14\% & +0.19 & +0.010 & +0.011 & \multicolumn{4}{c}{--} \\
F4Splat$^{\ast}$ (off) & 49\% & +1.20 & +0.015 & -0.006 & 40\% & +0.29 & -0.008 & +0.009 & \multicolumn{4}{c}{--} \\
F4Splat (50\%) & 37\% & +0.88 & +0.011 & +0.025 & 31\% & +0.32 & -0.003 & +0.031 & \multicolumn{4}{c}{--} \\
AdaptiveSplat (off) & 72\% & +1.04 & +0.012 & -0.008 & 70\% & +0.77 & +0.009 & -0.005 & 68\% & +0.56 & +0.006 & -0.002 \\
AdaptiveSplat$^{\ast}$ (60\%) & 51\% & +0.85 & +0.011 & +0.005 & \multicolumn{4}{c}{--} & \multicolumn{4}{c}{--} \\
\bottomrule
\end{tabular}
}
\end{table*}
}
\subsection{Pruning cost}
\label{sec:suppl-cost}

\Cref{tab:timing} gives the runtime numbers plotted in \cref{fig:timing}, with AnyPrune's time split into its stages.
We prune YoNoSplat fields from 2 to 32 views on five DL3DV-Benchmark scenes and report the median, after a warm-up pass; peak memory is what each method allocates beyond the field and the views, including its own weights.
Methods that need a kept ratio keep 30\%, and Speedy-Splat\,+\,optimization runs 3{,}000 iterations of 3DGS optimization after the cut.
AnyPrune's cost grows sub-linearly with the field size, and it is faster than NanoGS, GHAP and Speedy-Splat with optimization at every view count.

\subsection{Transfer across pipelines}
\label{sec:suppl-transfer}

\Cref{tab:transfer} tests whether AnyPrune trained on one pipeline can prune the fields of another without retraining.
We test the two reference models of the ablation studies, trained on YoNoSplat, on ReSplat's fields at 8 views of the same 50 held-out scenes.
As the reference, a third model is trained on ReSplat by the same recipe, scenes and view counts, on the fields of ReSplat's 8-view weights, which it also runs at test time.
We also apply each model's keep logits to the unadjusted field, to separate the ranking from the adjustments.
The models trained on YoNoSplat do not transfer: they degrade the unpruned field, their adjustments do not help, and their ranking falls behind the opacity baseline, so AnyPrune has to be retrained for each pipeline.

\begin{figure}[t]
\centering
\includegraphics[width=\linewidth]{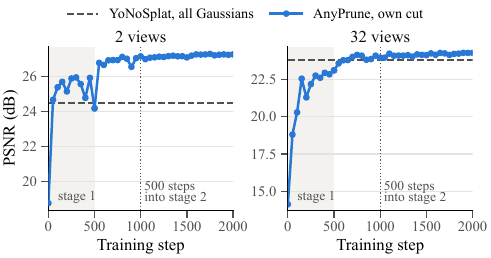}
\caption{\textbf{Validation PSNR during training.}
AnyPrune running on YoNoSplat, with the second training stage extended from 500 to 1500 steps, validated at its own cut on 16 held-out DL3DV scenes. Dashed: the unpruned field.}
\label{fig:train-length}
\end{figure}

\begin{table}[t]
\centering
\caption{\textbf{Training length.}
Change of performance on DL3DV when AnyPrune on YoNoSplat trains its second stage for 1500 steps instead of 500: mean over the kept ratios from 10\% to 100\%, and at the own cut, with the share it keeps.}
\label{tab:train-length}
\footnotesize
\setlength{\tabcolsep}{4pt}
\begin{tabular}{lccccc}
\toprule
 & \multicolumn{3}{c}{10--100\% kept} & \multicolumn{2}{c}{Own cut} \\
\cmidrule(lr){2-4}\cmidrule(lr){5-6}
Views & PSNR$\uparrow$ & SSIM$\uparrow$ & LPIPS$\downarrow$ & Kept & PSNR$\uparrow$ \\
\midrule
2 & +0.10 & +0.005 & -0.008 & 44\% $\to$ 42\% & +0.15 \\
4 & +0.05 & +0.003 & -0.003 & 35\% $\to$ 35\% & +0.07 \\
8 & +0.04 & +0.002 & -0.002 & 31\% $\to$ 32\% & +0.07 \\
16 & +0.02 & +0.001 & -0.001 & 28\% $\to$ 29\% & +0.06 \\
\bottomrule
\end{tabular}

\end{table}

\subsection{Training length}
\label{sec:suppl-train-length}

Each stage of our training schedule lasts 500 steps (\cref{sec:method-training}).
To check whether this is enough, we retrain AnyPrune on YoNoSplat with the two-stage schedule, extending the second stage to 1500 steps, over which its learning rate decays linearly to zero, and validate it every 50 steps on 16 held-out DL3DV scenes.
\Cref{fig:train-length} shows that the PSNR of AnyPrune's own cut rises quickly during the first stage and the first few hundred steps of the second, then flattens: the 1000 steps after the first 500 of the second stage add about 0.1\,dB at 2 views and 0.3\,dB at 32 views.
On the DL3DV test scenes (\cref{tab:train-length}), the longer schedule is slightly better at every view count from 2 to 16, by at most 0.1\,dB PSNR on average over the kept ratios, with small gains in SSIM and LPIPS too.
We therefore stop after 500 steps, since convergence is mostly reached by then, but one can train for longer to squeeze some more performance out of the model.

\section{Qualitative results}
\label{sec:suppl-qualitative}

\begin{figure*}[t]
\centering
\setlength{\unitlength}{\hsize}\divide\unitlength by 1000 %
\newcommand{\qf}{\fontsize{6.5}{7.3}\selectfont}%
\newcommand{\qi}[4]{\put(#1,#2){\includegraphics[width=#3\unitlength]{#4}}}%
\newcommand{\qt}[3]{\put(#1,#2){\makebox(0,0)[t]{\qf\shortstack{#3}}}}%
\newcommand{\qT}[3]{\put(#1,#2){\makebox(0,0)[t]{\footnotesize #3}}}%
\noindent\begin{picture}(1000,1138)
\put(504,1138){\makebox(0,0)[t]{\footnotesize 70\%}}
\put(614,1138){\makebox(0,0)[t]{\footnotesize 50\%}}
\put(726,1138){\makebox(0,0)[t]{\footnotesize 30\%}}
\put(836,1138){\makebox(0,0)[t]{\footnotesize 20\%}}
\put(948,1138){\makebox(0,0)[t]{\footnotesize 10\%}}
\put(445,1064){\makebox(0,0)[r]{\rotatebox{90}{\makebox[0pt]{\footnotesize\shortstack{\textbf{AnyPrune}\\\textbf{(ours)}}}}}}
\qi{451}{1011}{105}{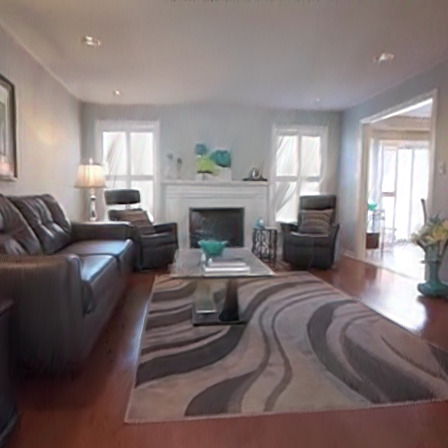}
\qt{504}{1009}{\textbf{26.77}\,/\,\textbf{.891}\,/\,\textbf{.229}}
\qi{562}{1011}{105}{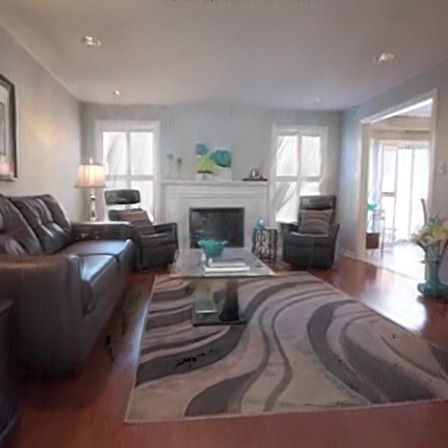}
\qt{614}{1009}{\textbf{26.23}\,/\,\textbf{.876}\,/\,\textbf{.251}}
\qi{673}{1011}{105}{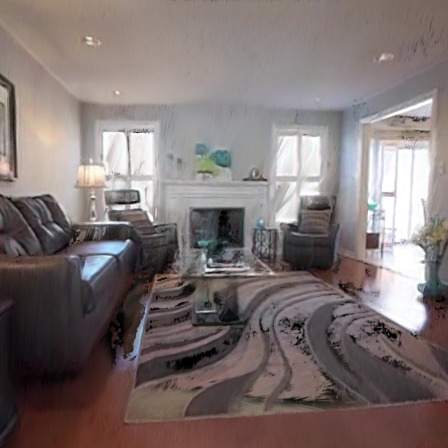}
\qt{726}{1009}{\textbf{22.71}\,/\,\textbf{.799}\,/\,\textbf{.335}}
\qi{784}{1011}{105}{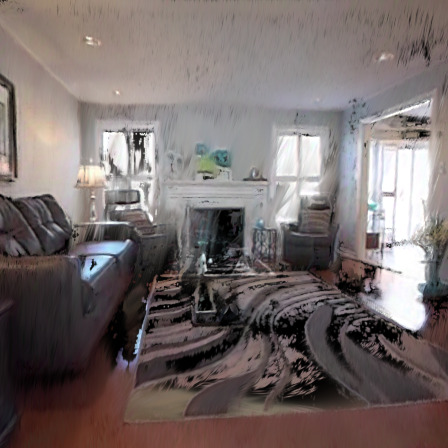}
\qt{836}{1009}{\textbf{18.38}\,/\,\textbf{.696}\,/\,\textbf{.428}}
\qi{895}{1011}{105}{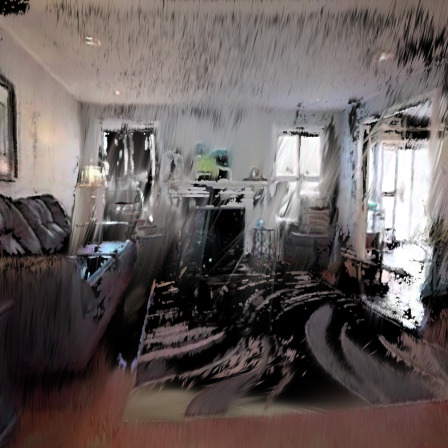}
\qt{948}{1009}{\textbf{14.02}\,/\,\textbf{.505}\,/\,\textbf{.541}}
\put(445,940){\makebox(0,0)[r]{\rotatebox{90}{\makebox[0pt]{\footnotesize\shortstack{RAP\\retrained}}}}}
\qi{451}{887}{105}{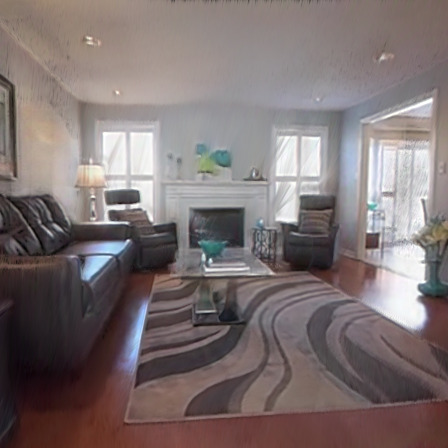}
\qt{504}{885}{24.51\,/\,.872\,/\,.266}
\qi{562}{887}{105}{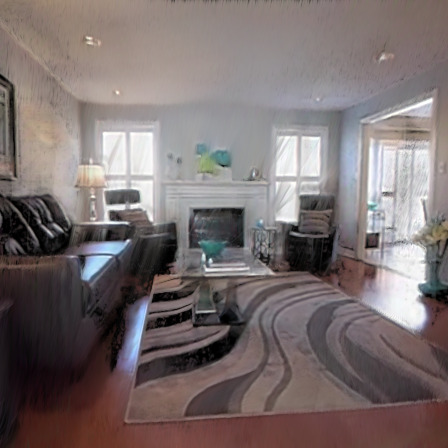}
\qt{614}{885}{22.61\,/\,.818\,/\,.330}
\qi{673}{887}{105}{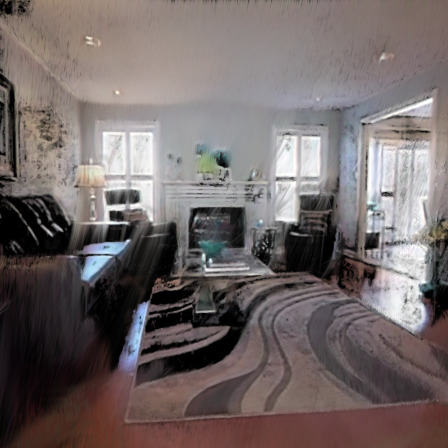}
\qt{726}{885}{18.98\,/\,.704\,/\,.417}
\qi{784}{887}{105}{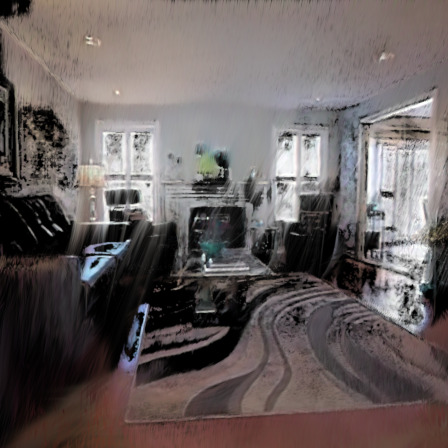}
\qt{836}{885}{15.51\,/\,.611\,/\,.477}
\qi{895}{887}{105}{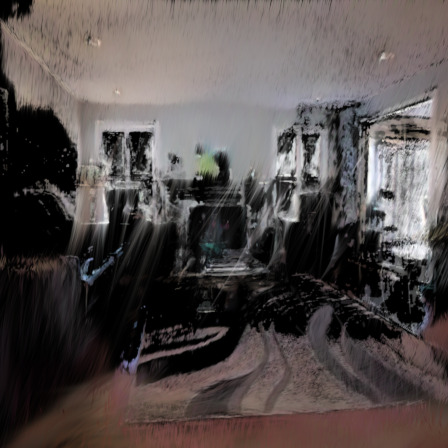}
\qt{948}{885}{11.20\,/\,.473\,/\,.561}
\put(445,816){\makebox(0,0)[r]{\rotatebox{90}{\makebox[0pt]{\footnotesize\shortstack{RAP\\released}}}}}
\qi{451}{763}{105}{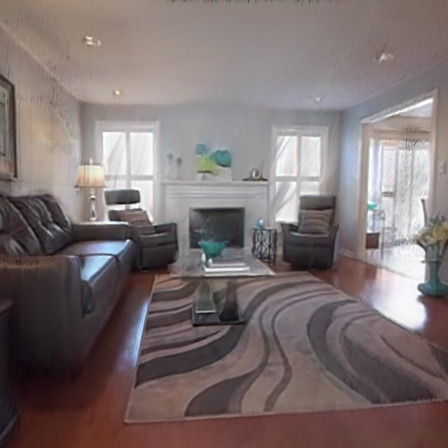}
\qt{504}{761}{23.64\,/\,.863\,/\,.272}
\qi{562}{763}{105}{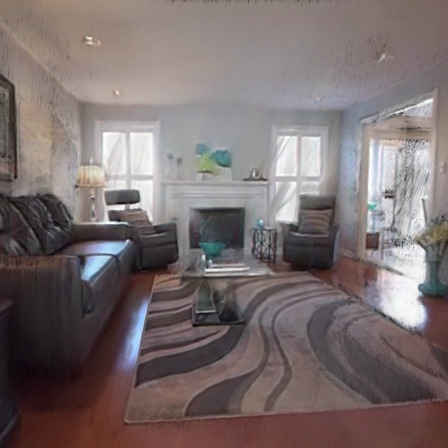}
\qt{614}{761}{21.15\,/\,.814\,/\,.318}
\qi{673}{763}{105}{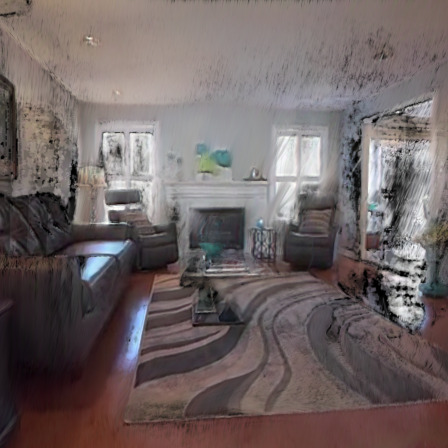}
\qt{726}{761}{15.96\,/\,.678\,/\,.417}
\qi{784}{763}{105}{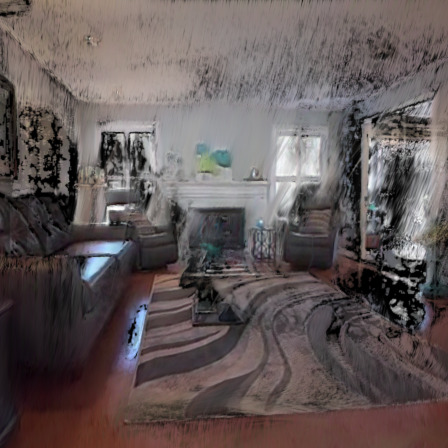}
\qt{836}{761}{12.88\,/\,.552\,/\,.495}
\qi{895}{763}{105}{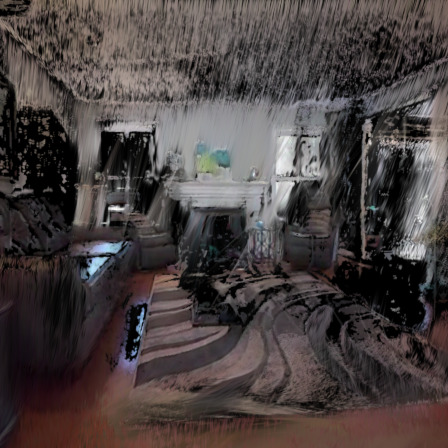}
\qt{948}{761}{9.91\,/\,.361\,/\,.590}
\put(445,692){\makebox(0,0)[r]{\rotatebox{90}{\makebox[0pt]{\footnotesize\shortstack{GaussianPOP}}}}}
\qi{451}{639}{105}{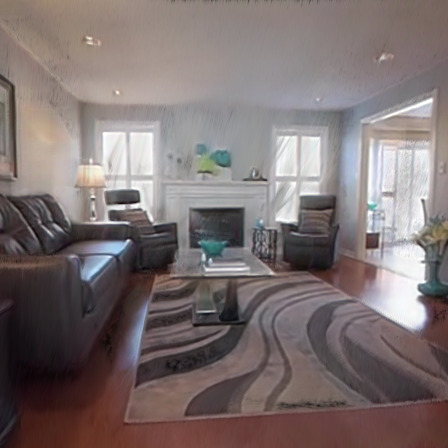}
\qt{504}{637}{24.53\,/\,.869\,/\,.279}
\qi{562}{639}{105}{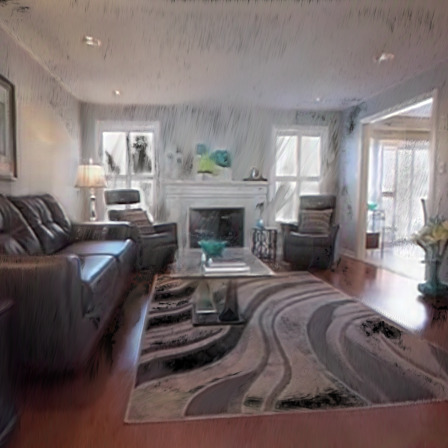}
\qt{614}{637}{21.70\,/\,.799\,/\,.359}
\qi{673}{639}{105}{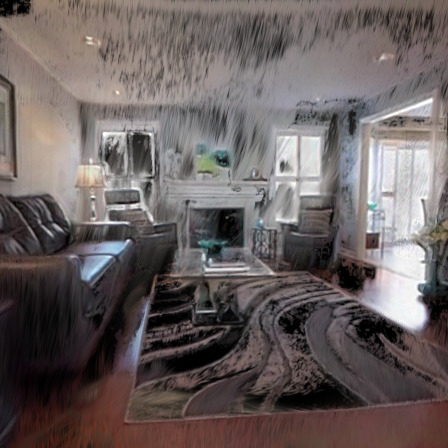}
\qt{726}{637}{16.54\,/\,.586\,/\,.478}
\qi{784}{639}{105}{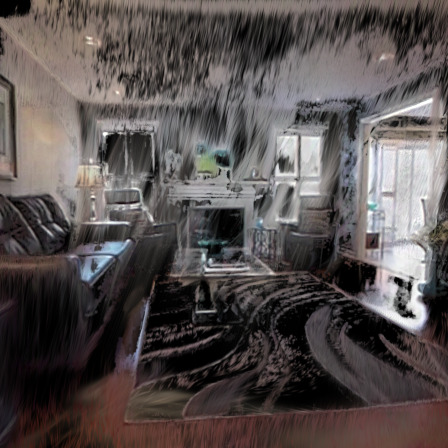}
\qt{836}{637}{13.04\,/\,.425\,/\,.556}
\qi{895}{639}{105}{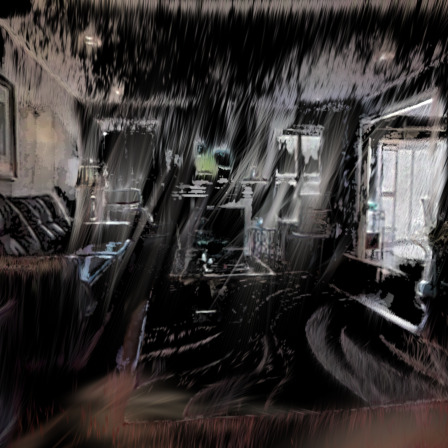}
\qt{948}{637}{9.31\,/\,.246\,/\,.632}
\put(445,568){\makebox(0,0)[r]{\rotatebox{90}{\makebox[0pt]{\footnotesize\shortstack{Mini-\\Splatting}}}}}
\qi{451}{515}{105}{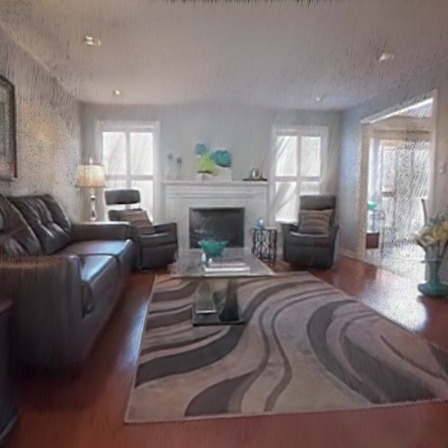}
\qt{504}{513}{22.83\,/\,.847\,/\,.313}
\qi{562}{515}{105}{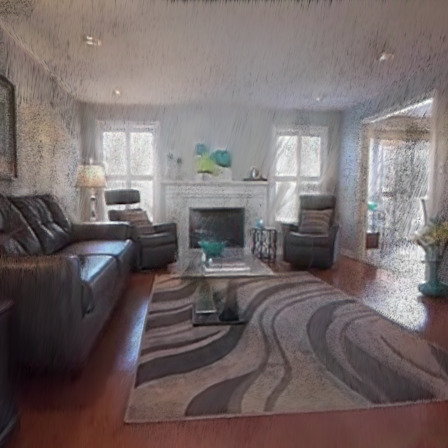}
\qt{614}{513}{20.30\,/\,.783\,/\,.388}
\qi{673}{515}{105}{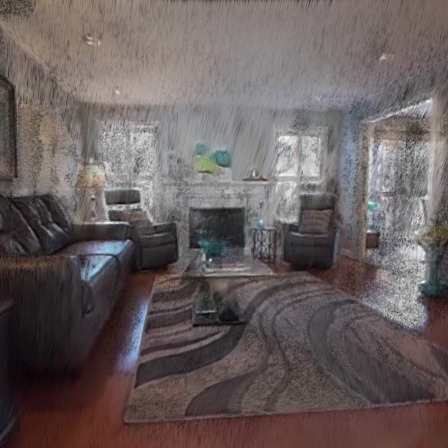}
\qt{726}{513}{16.46\,/\,.650\,/\,.489}
\qi{784}{515}{105}{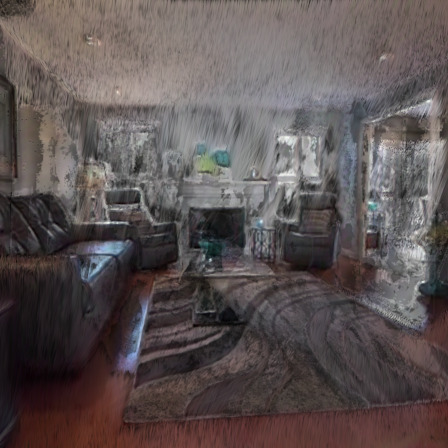}
\qt{836}{513}{13.89\,/\,.565\,/\,.521}
\qi{895}{515}{105}{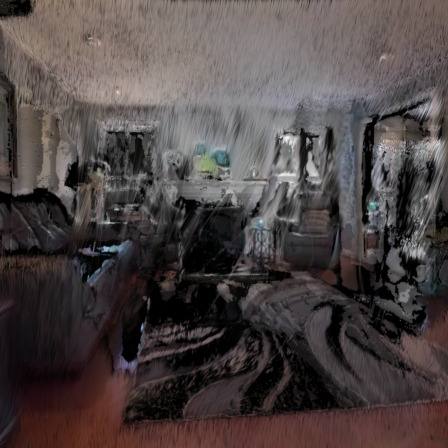}
\qt{948}{513}{10.75\,/\,.428\,/\,.575}
\put(445,444){\makebox(0,0)[r]{\rotatebox{90}{\makebox[0pt]{\footnotesize\shortstack{Light-\\Gaussian}}}}}
\qi{451}{391}{105}{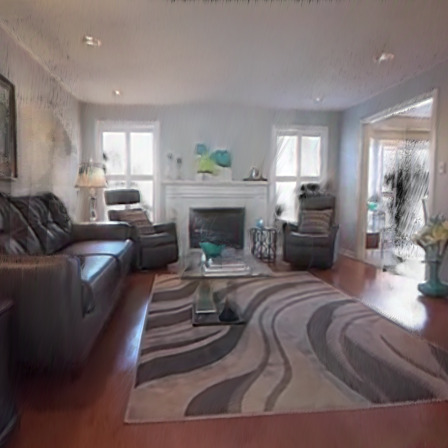}
\qt{504}{389}{20.39\,/\,.837\,/\,.280}
\qi{562}{391}{105}{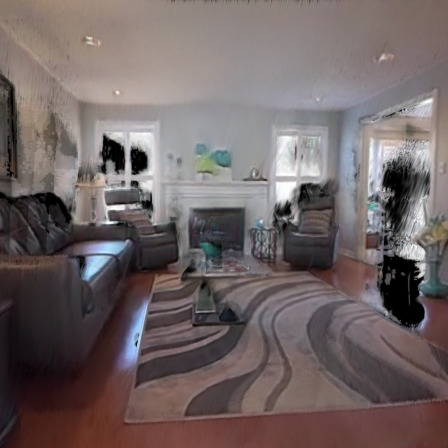}
\qt{614}{389}{14.34\,/\,.754\,/\,.330}
\qi{673}{391}{105}{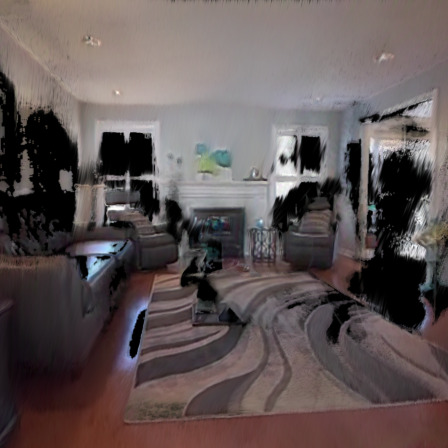}
\qt{726}{389}{10.68\,/\,.636\,/\,.412}
\qi{784}{391}{105}{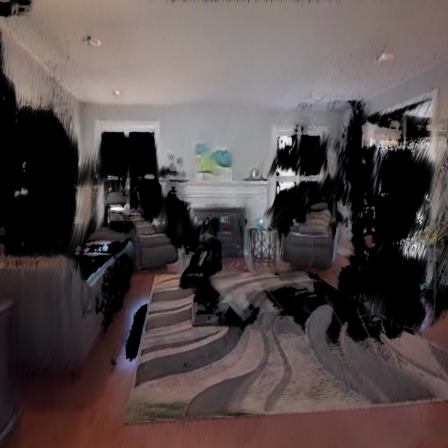}
\qt{836}{389}{9.31\,/\,.560\,/\,.469}
\qi{895}{391}{105}{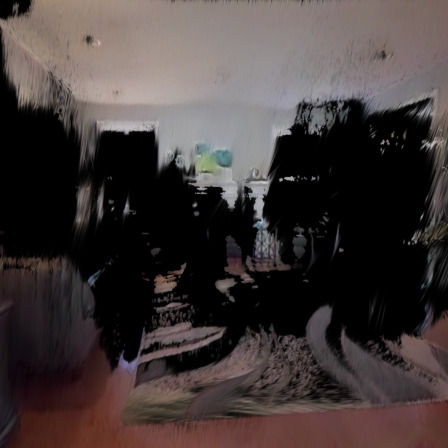}
\qt{948}{389}{8.15\,/\,.432\,/\,.550}
\put(445,320){\makebox(0,0)[r]{\rotatebox{90}{\makebox[0pt]{\footnotesize\shortstack{Speedy-\\Splat}}}}}
\qi{451}{267}{105}{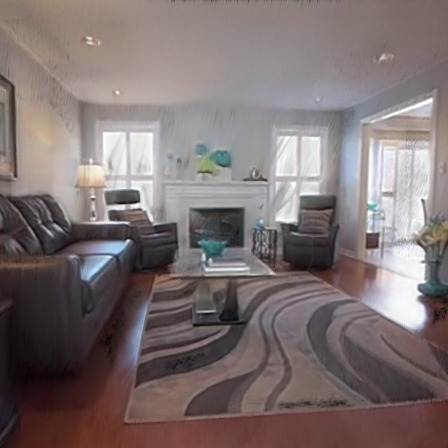}
\qt{504}{265}{24.47\,/\,.870\,/\,.275}
\qi{562}{267}{105}{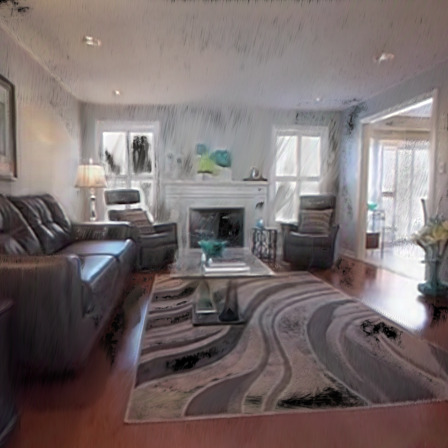}
\qt{614}{265}{21.26\,/\,.806\,/\,.349}
\qi{673}{267}{105}{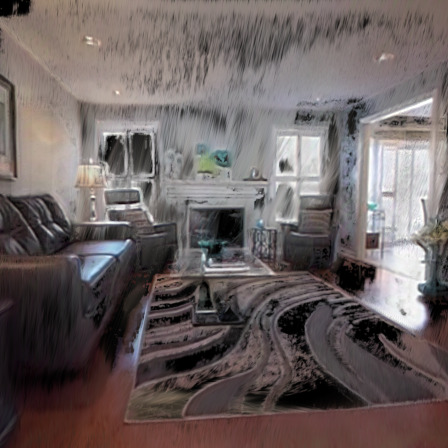}
\qt{726}{265}{16.42\,/\,.620\,/\,.467}
\qi{784}{267}{105}{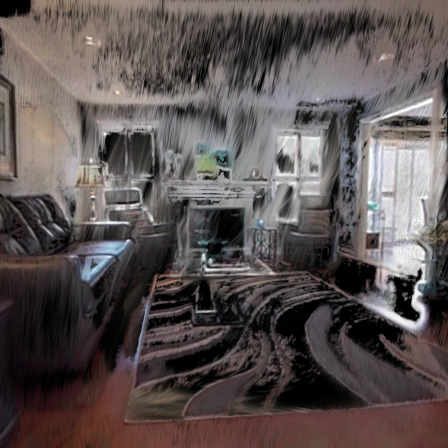}
\qt{836}{265}{13.09\,/\,.468\,/\,.539}
\qi{895}{267}{105}{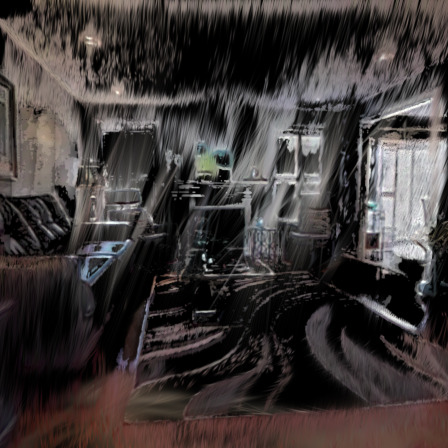}
\qt{948}{265}{9.65\,/\,.288\,/\,.611}
\put(445,196){\makebox(0,0)[r]{\rotatebox{90}{\makebox[0pt]{\footnotesize\shortstack{RadSplat}}}}}
\qi{451}{143}{105}{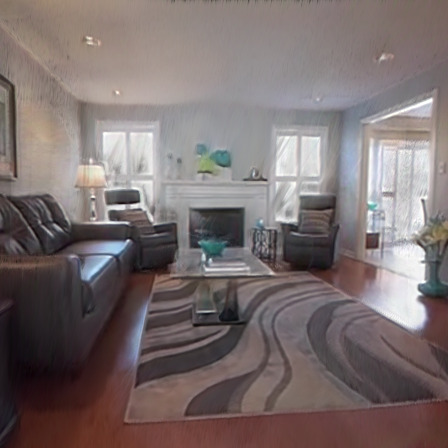}
\qt{504}{141}{23.13\,/\,.865\,/\,.277}
\qi{562}{143}{105}{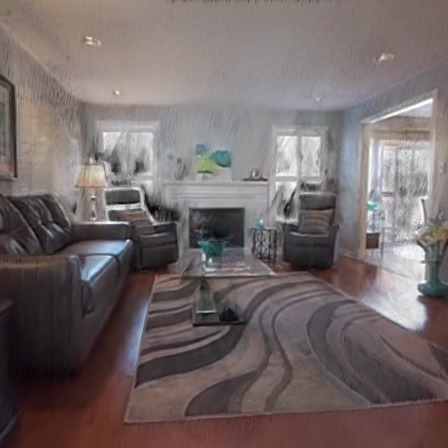}
\qt{614}{141}{19.15\,/\,.802\,/\,.348}
\qi{673}{143}{105}{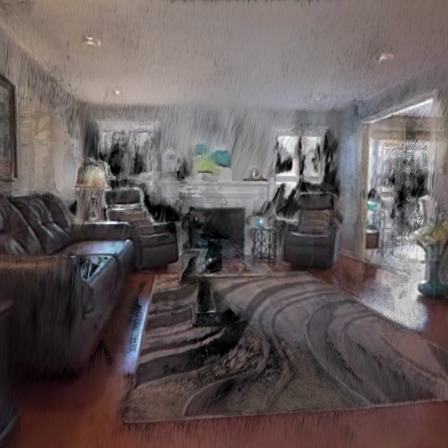}
\qt{726}{141}{14.49\,/\,.650\,/\,.467}
\qi{784}{143}{105}{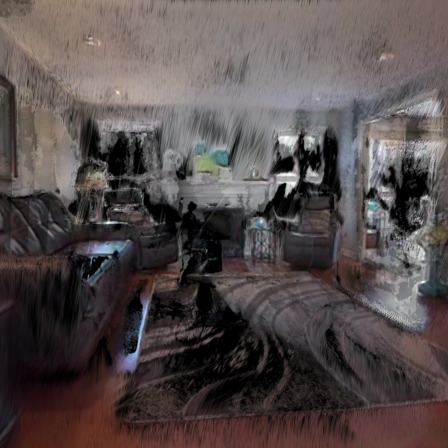}
\qt{836}{141}{11.67\,/\,.502\,/\,.547}
\qi{895}{143}{105}{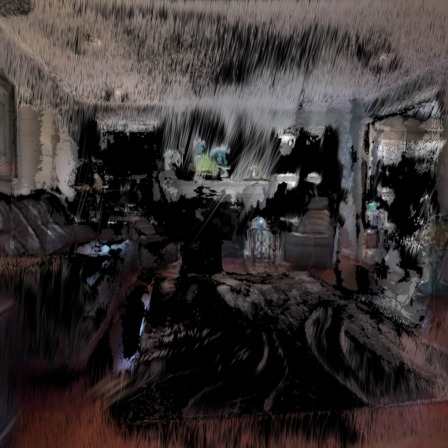}
\qt{948}{141}{8.83\,/\,.267\,/\,.642}
\put(445,72){\makebox(0,0)[r]{\rotatebox{90}{\makebox[0pt]{\footnotesize\shortstack{REFINE}}}}}
\qi{451}{19}{105}{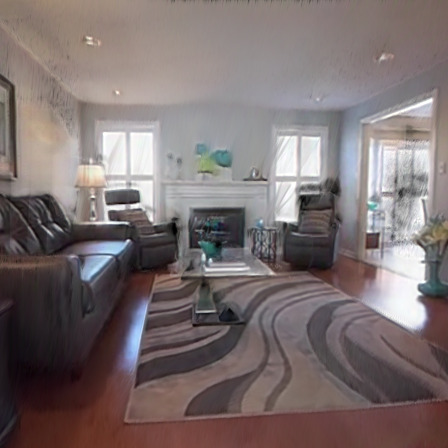}
\qt{504}{17}{22.49\,/\,.853\,/\,.272}
\qi{562}{19}{105}{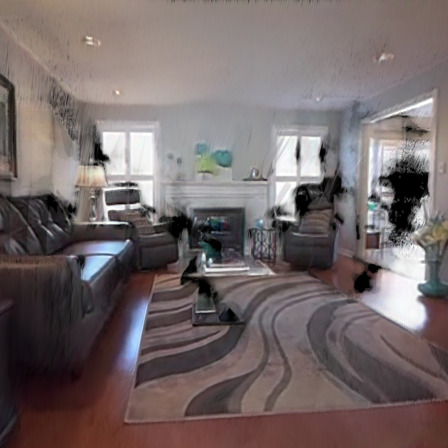}
\qt{614}{17}{16.51\,/\,.782\,/\,.335}
\qi{673}{19}{105}{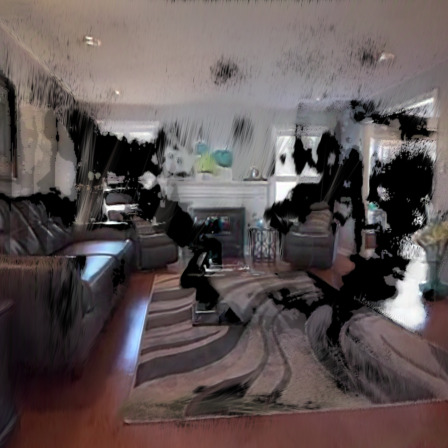}
\qt{726}{17}{11.24\,/\,.607\,/\,.462}
\qi{784}{19}{105}{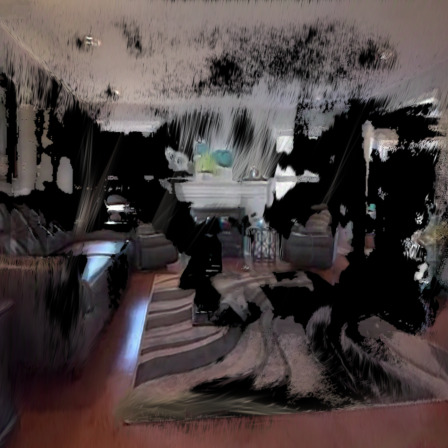}
\qt{836}{17}{8.91\,/\,.436\,/\,.551}
\qi{895}{19}{105}{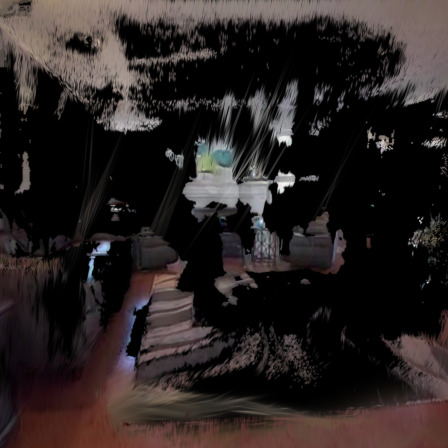}
\qt{948}{17}{7.00\,/\,.257\,/\,.640}
\put(236,1138){\makebox(0,0)[t]{\footnotesize GT}}
\qi{70}{783}{333}{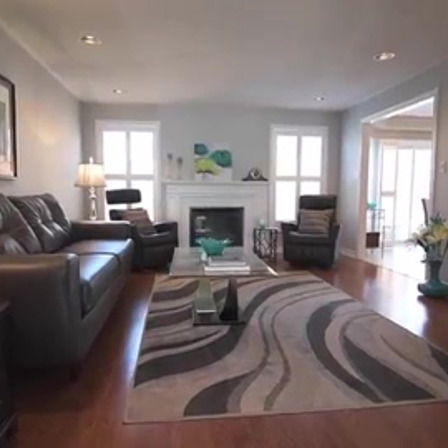}
\put(236,759){\makebox(0,0)[t]{\footnotesize Unpruned}}
\qi{70}{403}{333}{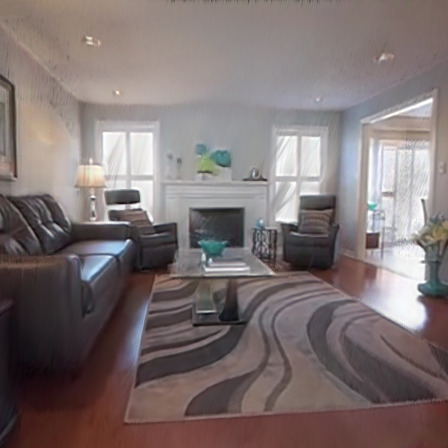}
\qT{236}{401}{24.84\,/\,.884\,/\,.250}
\put(236,379){\makebox(0,0)[t]{\footnotesize \textbf{AnyPrune (ours)}, 100\%}}
\qi{70}{24}{333}{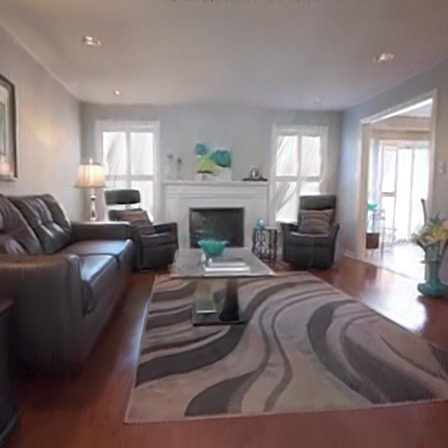}
\qT{236}{22}{\textbf{26.83}\,/\,\textbf{.894}\,/\,\textbf{.223}}
\end{picture}%
\caption{Every optimization-free pruning method on AdaptiveSplat (60\%), on RealEstate10K scene 42 at 16 views (1{,}926{,}784 Gaussians). At 100\%, AnyPrune only adjusts the Gaussians. Under each render: PSNR\,/\,SSIM\,/\,LPIPS of the shown view. In parentheses: the share kept by the pipeline's internal pruning dial, or off.}
\label{fig:qual-full-1}
\end{figure*}

\begin{figure*}[t]
\centering
\setlength{\unitlength}{\hsize}\divide\unitlength by 1000 %
\newcommand{\qf}{\fontsize{6.5}{7.3}\selectfont}%
\newcommand{\qi}[4]{\put(#1,#2){\includegraphics[width=#3\unitlength]{#4}}}%
\newcommand{\qt}[3]{\put(#1,#2){\makebox(0,0)[t]{\qf\shortstack{#3}}}}%
\newcommand{\qT}[3]{\put(#1,#2){\makebox(0,0)[t]{\footnotesize #3}}}%
\noindent\begin{picture}(1000,1138)
\put(504,1138){\makebox(0,0)[t]{\footnotesize 70\%}}
\put(614,1138){\makebox(0,0)[t]{\footnotesize 50\%}}
\put(726,1138){\makebox(0,0)[t]{\footnotesize 30\%}}
\put(836,1138){\makebox(0,0)[t]{\footnotesize 20\%}}
\put(948,1138){\makebox(0,0)[t]{\footnotesize 10\%}}
\put(445,1064){\makebox(0,0)[r]{\rotatebox{90}{\makebox[0pt]{\footnotesize\shortstack{\textbf{AnyPrune}\\\textbf{(ours)}}}}}}
\qi{451}{1011}{105}{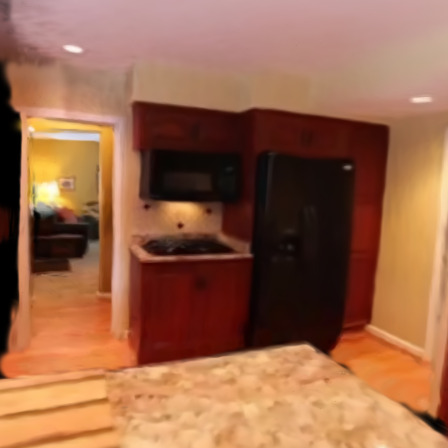}
\qt{504}{1009}{\textbf{20.10}\,/\,\textbf{.781}\,/\,\textbf{.270}}
\qi{562}{1011}{105}{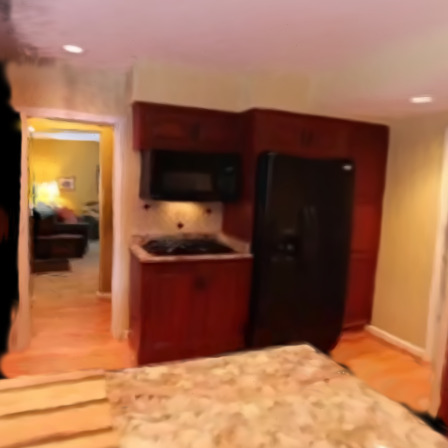}
\qt{614}{1009}{\textbf{20.14}\,/\,\textbf{.784}\,/\,\textbf{.268}}
\qi{673}{1011}{105}{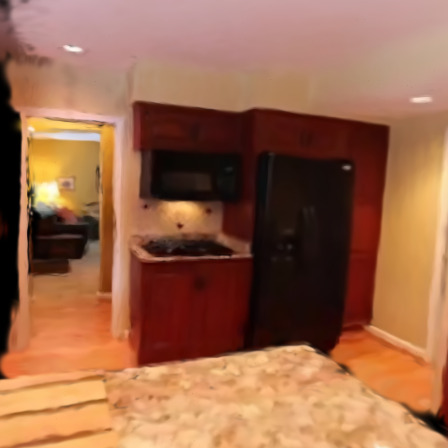}
\qt{726}{1009}{\textbf{19.92}\,/\,\textbf{.784}\,/\,\textbf{.280}}
\qi{784}{1011}{105}{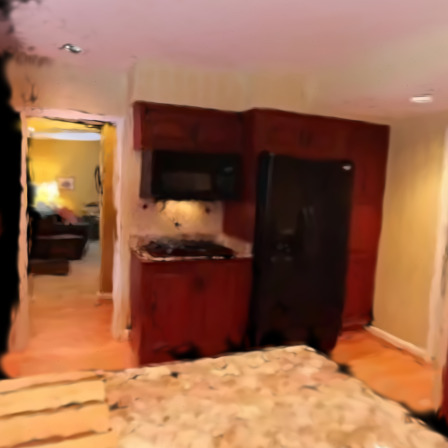}
\qt{836}{1009}{\textbf{19.33}\,/\,\textbf{.767}\,/\,\textbf{.314}}
\qi{895}{1011}{105}{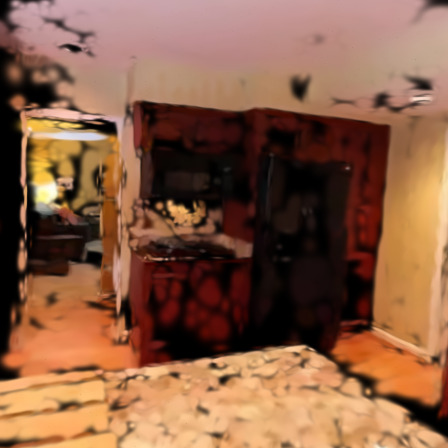}
\qt{948}{1009}{\textbf{15.94}\,/\,\textbf{.614}\,/\,\textbf{.468}}
\put(445,940){\makebox(0,0)[r]{\rotatebox{90}{\makebox[0pt]{\footnotesize\shortstack{RAP\\retrained}}}}}
\qi{451}{887}{105}{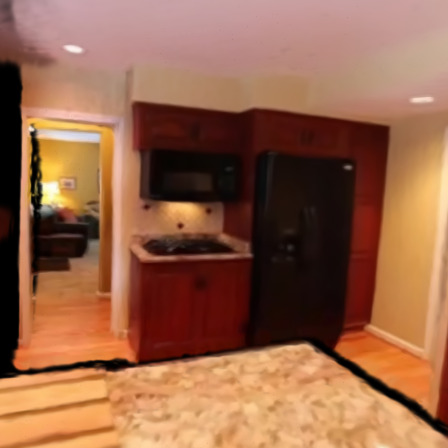}
\qt{504}{885}{18.07\,/\,.758\,/\,.274}
\qi{562}{887}{105}{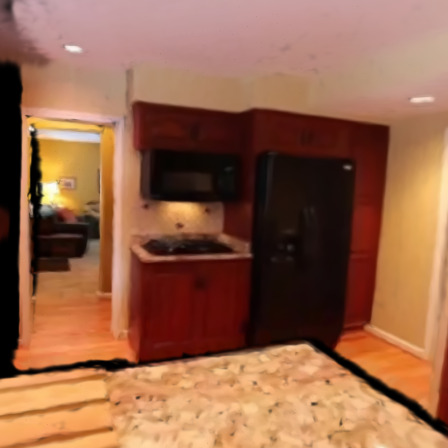}
\qt{614}{885}{17.93\,/\,.762\,/\,.291}
\qi{673}{887}{105}{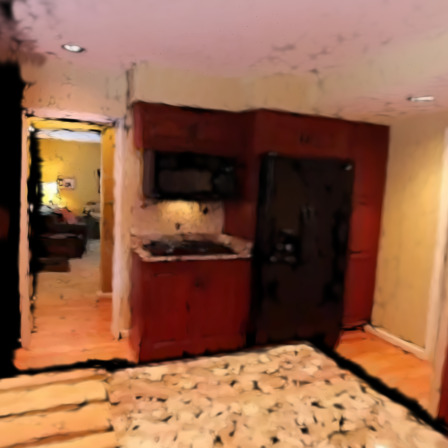}
\qt{726}{885}{17.19\,/\,.727\,/\,.368}
\qi{784}{887}{105}{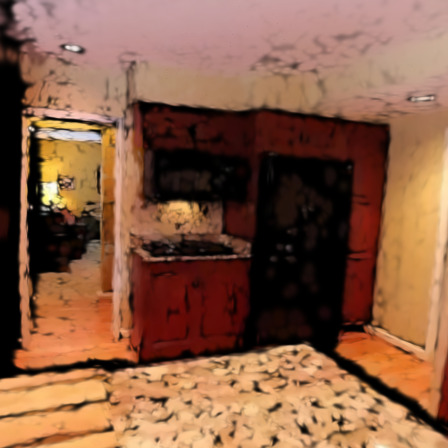}
\qt{836}{885}{16.07\,/\,.615\,/\,.453}
\qi{895}{887}{105}{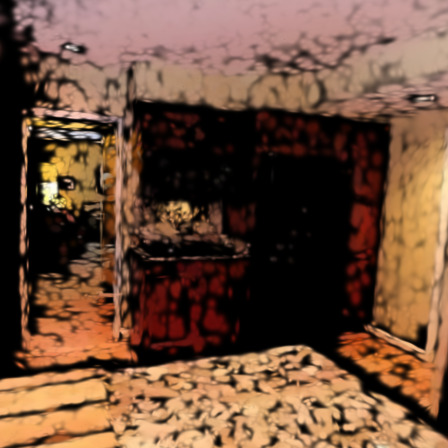}
\qt{948}{885}{13.04\,/\,.395\,/\,.583}
\put(445,816){\makebox(0,0)[r]{\rotatebox{90}{\makebox[0pt]{\footnotesize\shortstack{RAP\\released}}}}}
\qi{451}{763}{105}{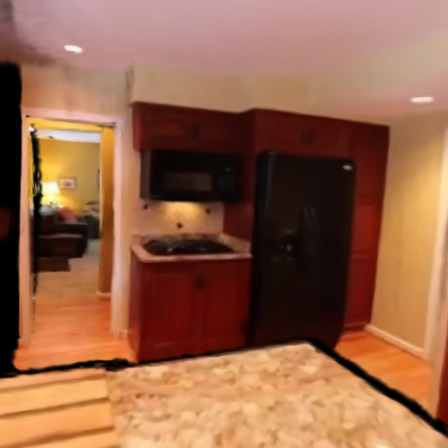}
\qt{504}{761}{18.11\,/\,.755\,/\,.280}
\qi{562}{763}{105}{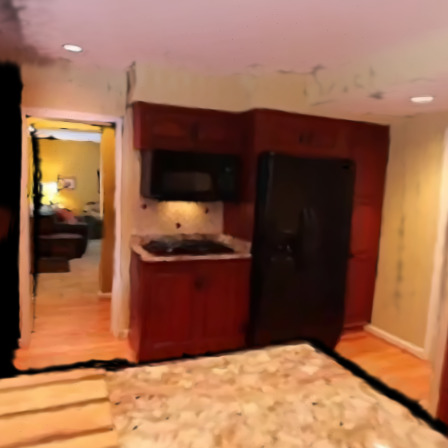}
\qt{614}{761}{17.91\,/\,.747\,/\,.313}
\qi{673}{763}{105}{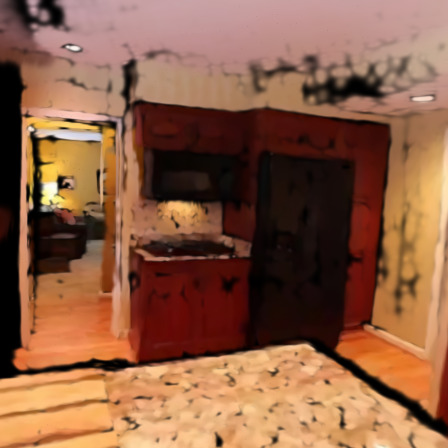}
\qt{726}{761}{15.85\,/\,.658\,/\,.430}
\qi{784}{763}{105}{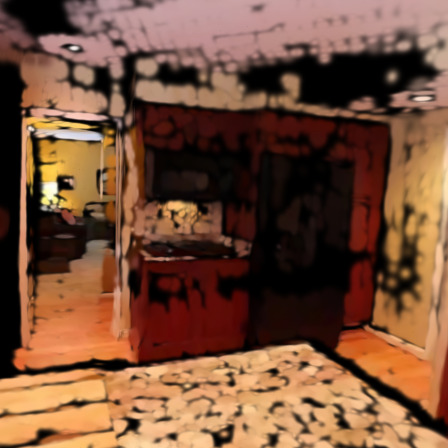}
\qt{836}{761}{12.40\,/\,.485\,/\,.541}
\qi{895}{763}{105}{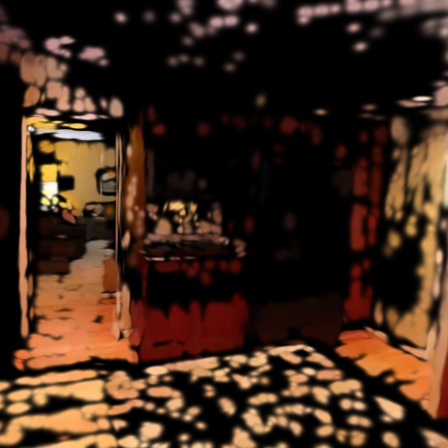}
\qt{948}{761}{7.89\,/\,.226\,/\,.667}
\put(445,692){\makebox(0,0)[r]{\rotatebox{90}{\makebox[0pt]{\footnotesize\shortstack{GaussianPOP}}}}}
\qi{451}{639}{105}{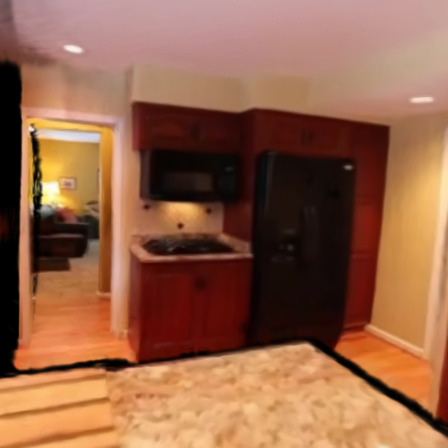}
\qt{504}{637}{18.07\,/\,.753\,/\,.280}
\qi{562}{639}{105}{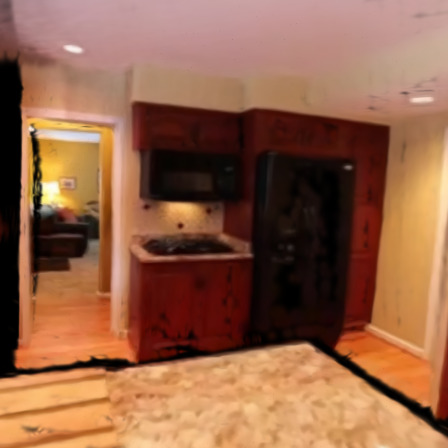}
\qt{614}{637}{17.88\,/\,.730\,/\,.320}
\qi{673}{639}{105}{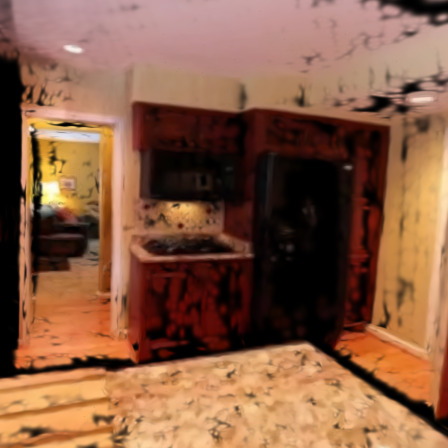}
\qt{726}{637}{16.18\,/\,.601\,/\,.443}
\qi{784}{639}{105}{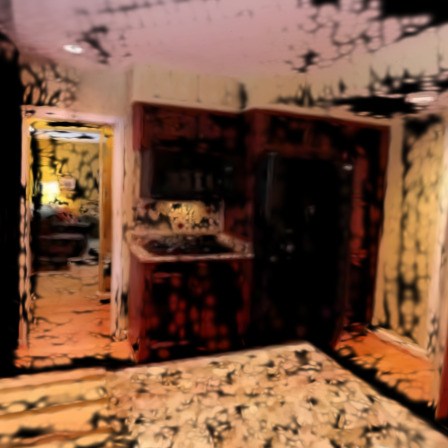}
\qt{836}{637}{13.94\,/\,.454\,/\,.530}
\qi{895}{639}{105}{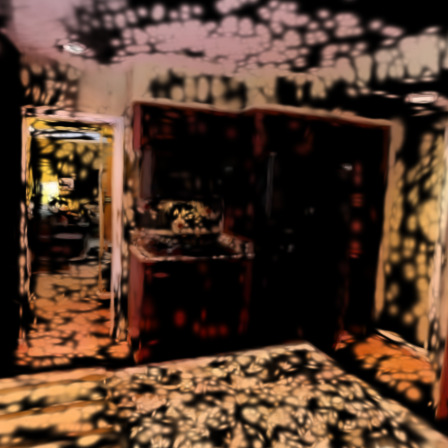}
\qt{948}{637}{10.07\,/\,.232\,/\,.636}
\put(445,568){\makebox(0,0)[r]{\rotatebox{90}{\makebox[0pt]{\footnotesize\shortstack{Mini-\\Splatting}}}}}
\qi{451}{515}{105}{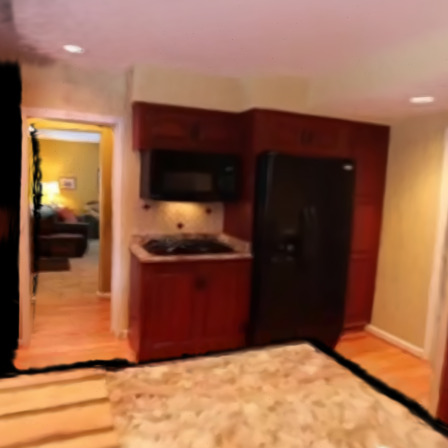}
\qt{504}{513}{18.10\,/\,.759\,/\,.278}
\qi{562}{515}{105}{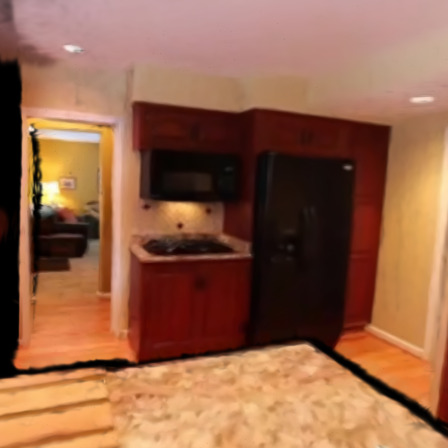}
\qt{614}{513}{18.01\,/\,.761\,/\,.283}
\qi{673}{515}{105}{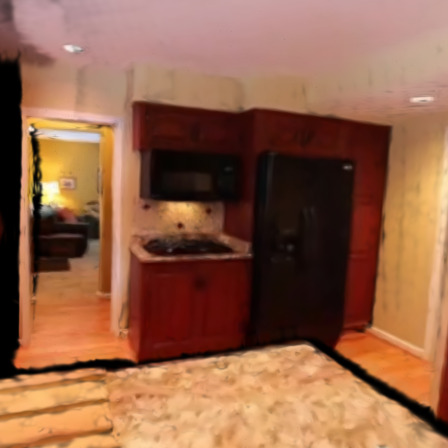}
\qt{726}{513}{17.76\,/\,.752\,/\,.311}
\qi{784}{515}{105}{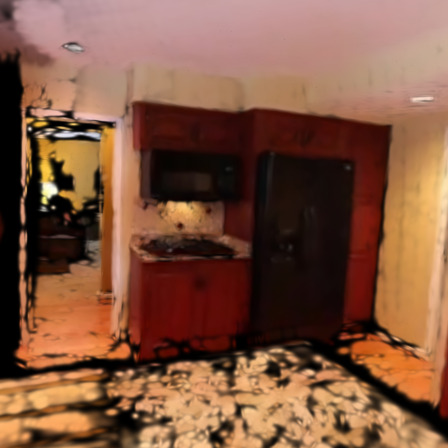}
\qt{836}{513}{13.95\,/\,.647\,/\,.398}
\qi{895}{515}{105}{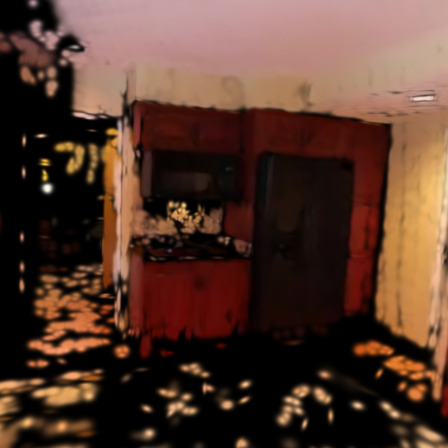}
\qt{948}{513}{9.05\,/\,.491\,/\,.536}
\put(445,444){\makebox(0,0)[r]{\rotatebox{90}{\makebox[0pt]{\footnotesize\shortstack{Light-\\Gaussian}}}}}
\qi{451}{391}{105}{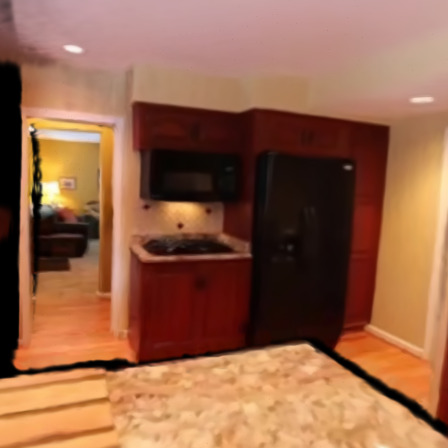}
\qt{504}{389}{18.05\,/\,.754\,/\,.277}
\qi{562}{391}{105}{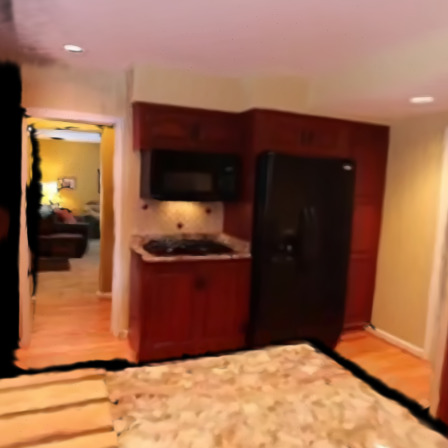}
\qt{614}{389}{17.63\,/\,.751\,/\,.283}
\qi{673}{391}{105}{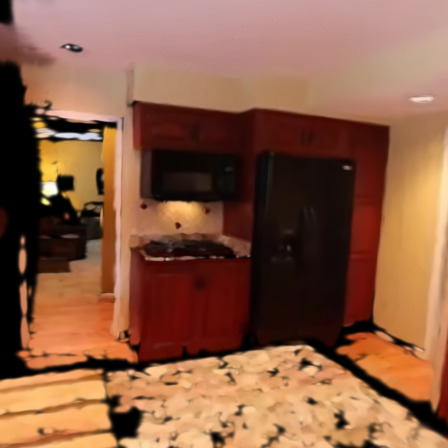}
\qt{726}{389}{15.02\,/\,.707\,/\,.346}
\qi{784}{391}{105}{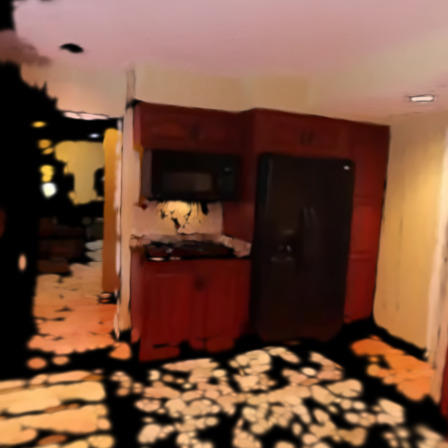}
\qt{836}{389}{11.94\,/\,.617\,/\,.417}
\qi{895}{391}{105}{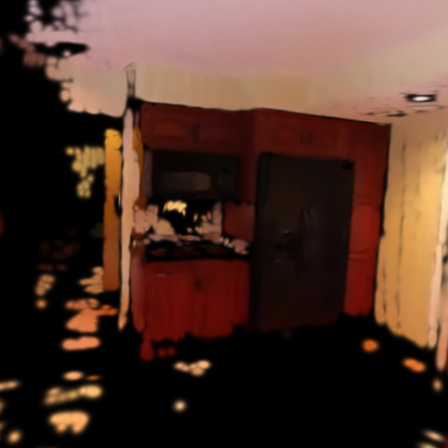}
\qt{948}{389}{8.37\,/\,.480\,/\,.536}
\put(445,320){\makebox(0,0)[r]{\rotatebox{90}{\makebox[0pt]{\footnotesize\shortstack{Speedy-\\Splat}}}}}
\qi{451}{267}{105}{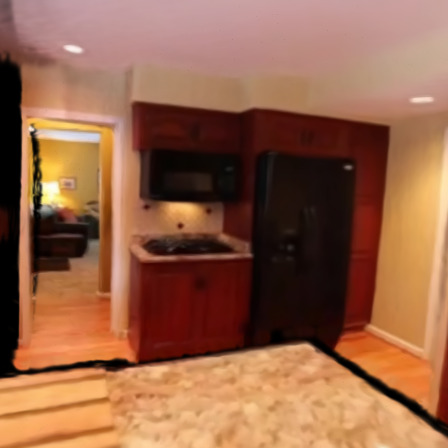}
\qt{504}{265}{18.09\,/\,.752\,/\,.277}
\qi{562}{267}{105}{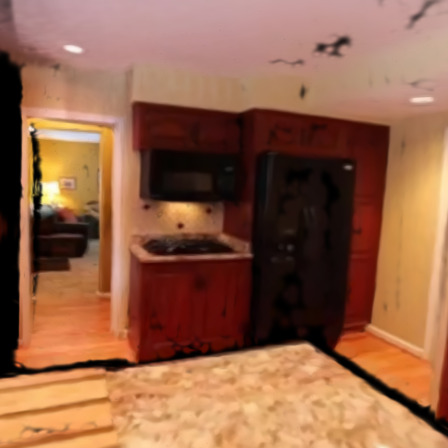}
\qt{614}{265}{17.78\,/\,.729\,/\,.325}
\qi{673}{267}{105}{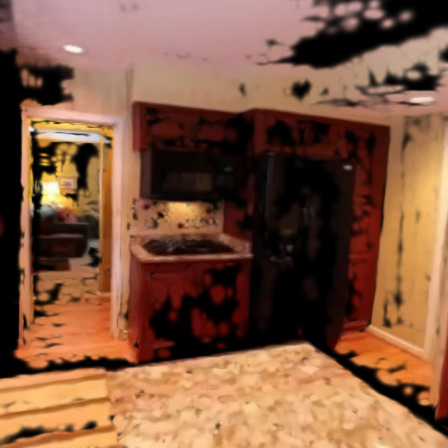}
\qt{726}{265}{13.98\,/\,.556\,/\,.466}
\qi{784}{267}{105}{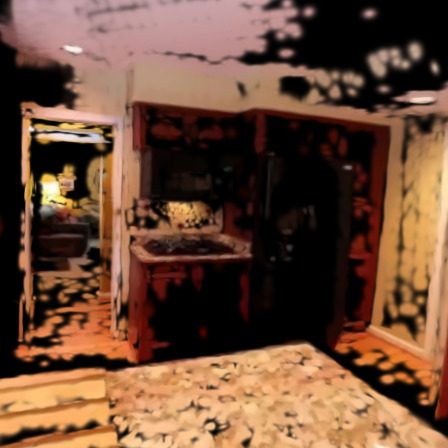}
\qt{836}{265}{11.48\,/\,.405\,/\,.549}
\qi{895}{267}{105}{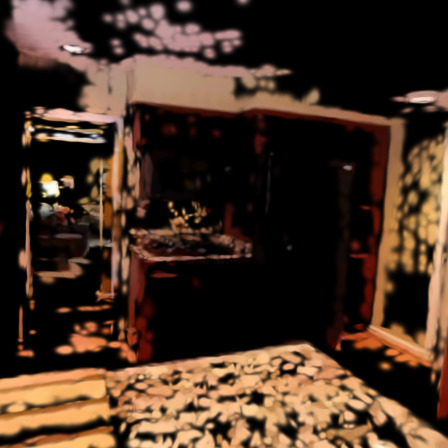}
\qt{948}{265}{8.54\,/\,.218\,/\,.646}
\put(445,196){\makebox(0,0)[r]{\rotatebox{90}{\makebox[0pt]{\footnotesize\shortstack{RadSplat}}}}}
\qi{451}{143}{105}{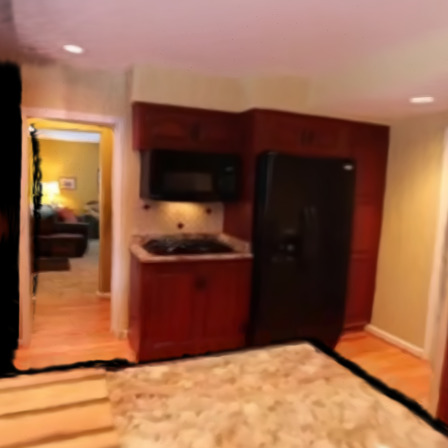}
\qt{504}{141}{18.12\,/\,.754\,/\,.277}
\qi{562}{143}{105}{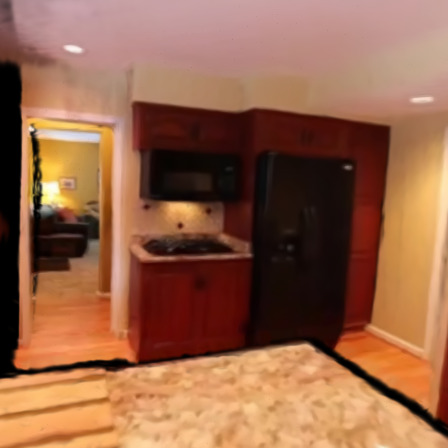}
\qt{614}{141}{18.12\,/\,.759\,/\,.281}
\qi{673}{143}{105}{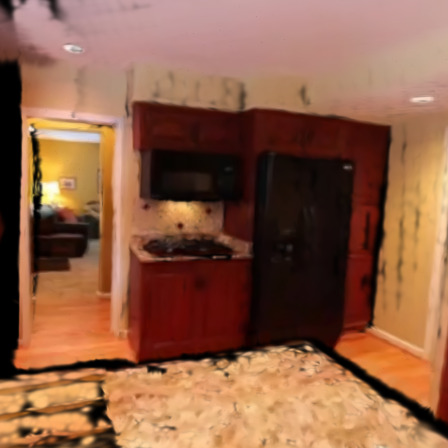}
\qt{726}{141}{16.74\,/\,.717\,/\,.354}
\qi{784}{143}{105}{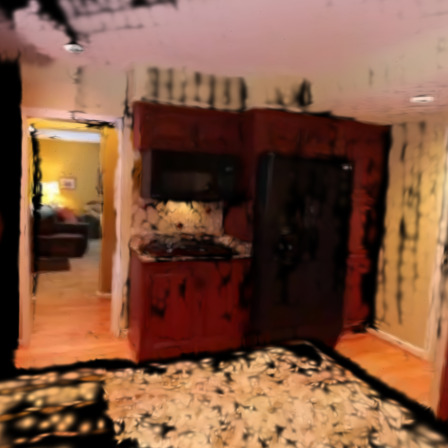}
\qt{836}{141}{14.09\,/\,.596\,/\,.449}
\qi{895}{143}{105}{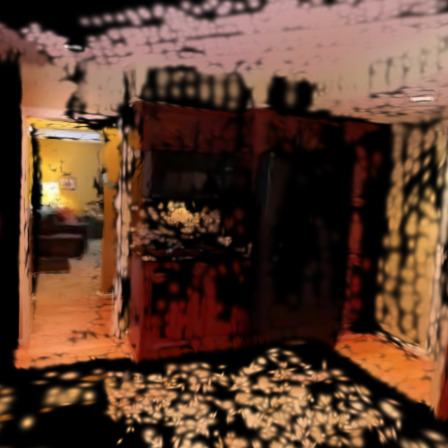}
\qt{948}{141}{10.16\,/\,.348\,/\,.606}
\put(445,72){\makebox(0,0)[r]{\rotatebox{90}{\makebox[0pt]{\footnotesize\shortstack{REFINE}}}}}
\qi{451}{19}{105}{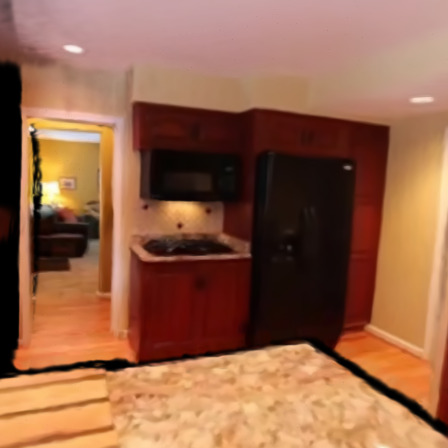}
\qt{504}{17}{18.08\,/\,.753\,/\,.276}
\qi{562}{19}{105}{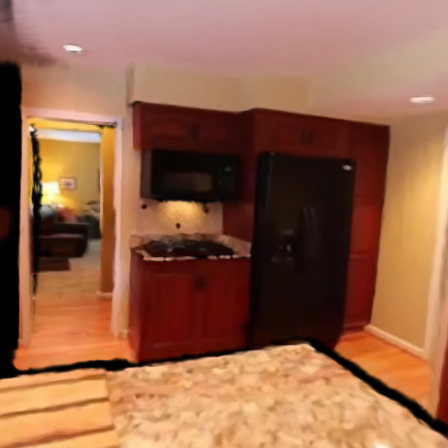}
\qt{614}{17}{17.91\,/\,.751\,/\,.285}
\qi{673}{19}{105}{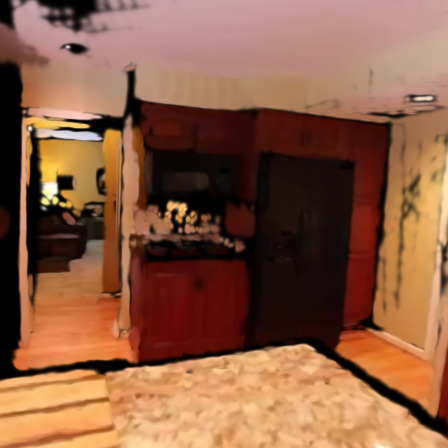}
\qt{726}{17}{15.21\,/\,.661\,/\,.400}
\qi{784}{19}{105}{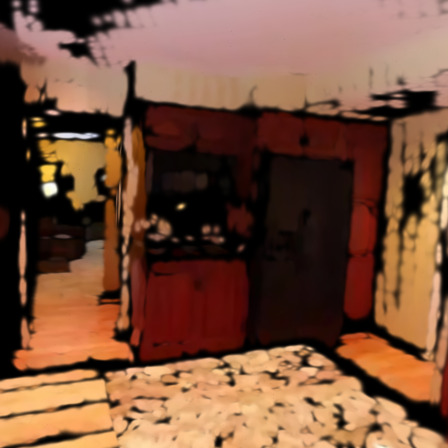}
\qt{836}{17}{12.27\,/\,.574\,/\,.500}
\qi{895}{19}{105}{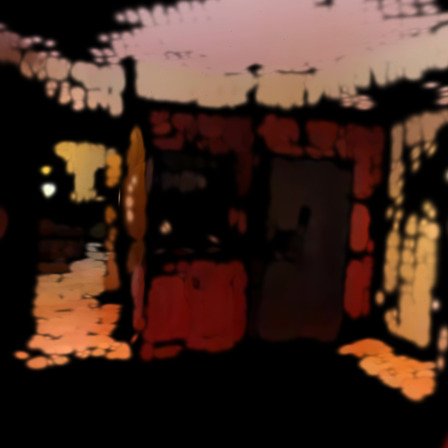}
\qt{948}{17}{7.54\,/\,.348\,/\,.644}
\put(236,1138){\makebox(0,0)[t]{\footnotesize GT}}
\qi{70}{783}{333}{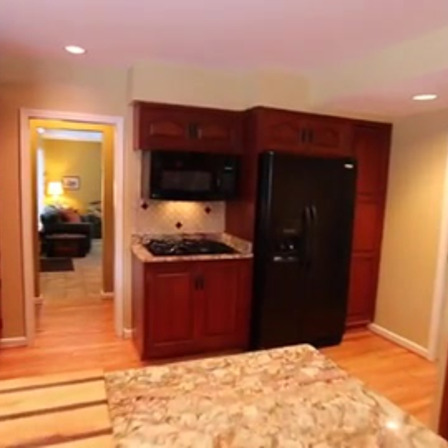}
\put(236,759){\makebox(0,0)[t]{\footnotesize Unpruned}}
\qi{70}{403}{333}{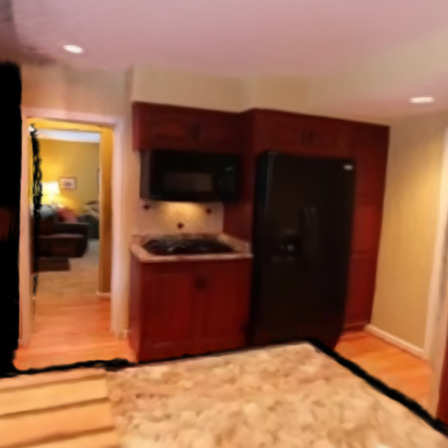}
\qT{236}{401}{18.13\,/\,.753\,/\,.277}
\put(236,379){\makebox(0,0)[t]{\footnotesize \textbf{AnyPrune (ours)}, 100\%}}
\qi{70}{24}{333}{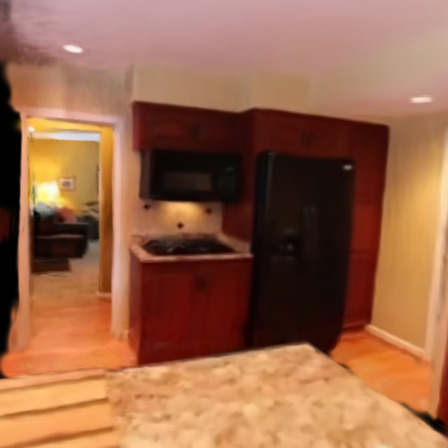}
\qT{236}{22}{\textbf{20.05}\,/\,\textbf{.780}\,/\,\textbf{.270}}
\end{picture}%
\caption{Every optimization-free pruning method on Off The Grid, on RealEstate10K scene 26 at 8 views (307{,}200 Gaussians). At 100\%, AnyPrune only adjusts the Gaussians. Under each render: PSNR\,/\,SSIM\,/\,LPIPS of the shown view.}
\label{fig:qual-full-2}
\end{figure*}

\begin{figure*}[t]
\centering
\setlength{\unitlength}{\hsize}\divide\unitlength by 1000 %
\newcommand{\qf}{\fontsize{6.5}{7.3}\selectfont}%
\newcommand{\qi}[4]{\put(#1,#2){\includegraphics[width=#3\unitlength]{#4}}}%
\newcommand{\qt}[3]{\put(#1,#2){\makebox(0,0)[t]{\qf\shortstack{#3}}}}%
\newcommand{\qT}[3]{\put(#1,#2){\makebox(0,0)[t]{\footnotesize #3}}}%
\noindent\begin{picture}(1000,1158)
\put(226,1158){\makebox(0,0)[t]{\footnotesize GT}}
\put(436,1158){\makebox(0,0)[t]{\footnotesize 100\%}}
\put(540,1158){\makebox(0,0)[t]{\footnotesize 70\%}}
\put(642,1158){\makebox(0,0)[t]{\footnotesize 50\%}}
\put(746,1158){\makebox(0,0)[t]{\footnotesize 30\%}}
\put(848,1158){\makebox(0,0)[t]{\footnotesize 20\%}}
\put(952,1158){\makebox(0,0)[t]{\footnotesize 10\%}}
\put(112,1136){\makebox(0,0)[lt]{\footnotesize\textit{YoNoSplat, DL3DV, 16 views, 802{,}816 Gaussians}}}
\qi{112}{886}{228}{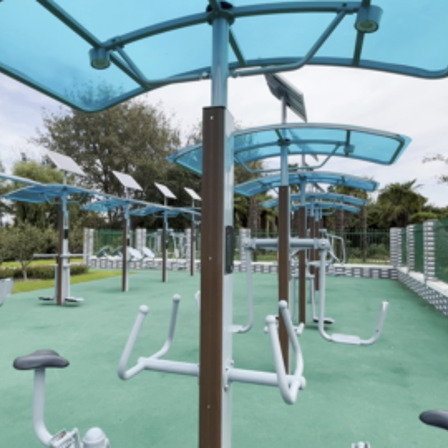}
\put(382,1066){\makebox(0,0)[r]{\rotatebox{90}{\makebox[0pt]{\footnotesize\shortstack{\textbf{AnyPrune}\\\textbf{(ours)}}}}}}
\put(382,934){\makebox(0,0)[r]{\rotatebox{90}{\makebox[0pt]{\footnotesize\shortstack{Best\\other}}}}}
\qi{388}{1017}{97}{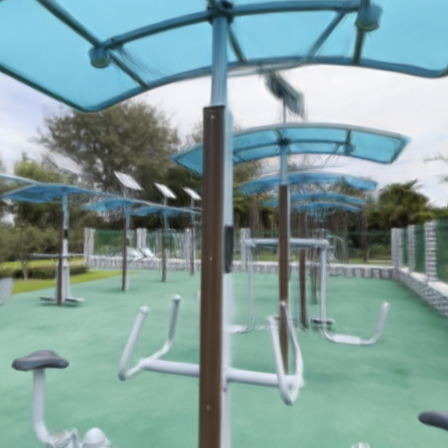}
\qt{436}{1015}{\textbf{25.03}\,/\,\textbf{.849}\,/\,\textbf{.168}}
\qi{388}{886}{97}{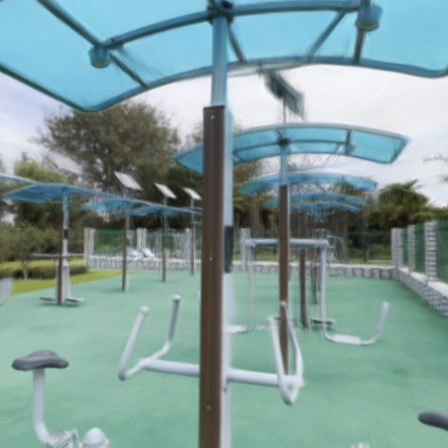}
\qt{436}{884}{24.29\,/\,.839\,/\,.173\\\textit{Unpruned}}
\qi{491}{1017}{97}{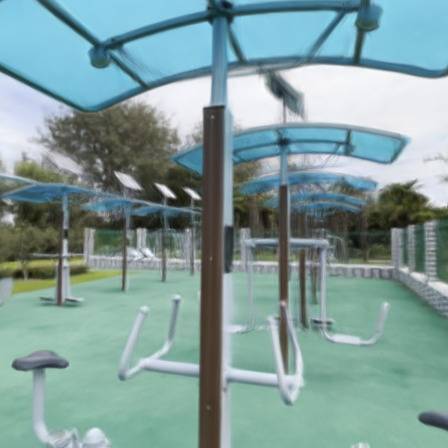}
\qt{540}{1015}{\textbf{25.04}\,/\,\textbf{.849}\,/\,\textbf{.168}}
\qi{491}{886}{97}{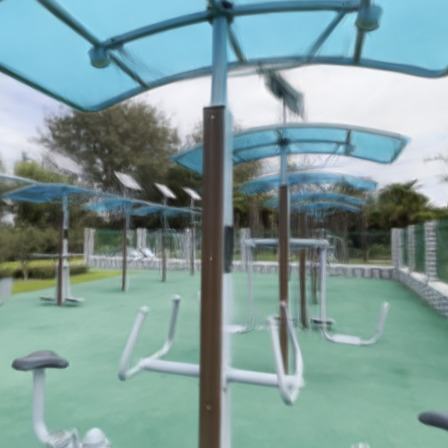}
\qt{540}{884}{24.30\,/\,.839\,/\,.173\\\textit{LightGaussian}}
\qi{594}{1017}{97}{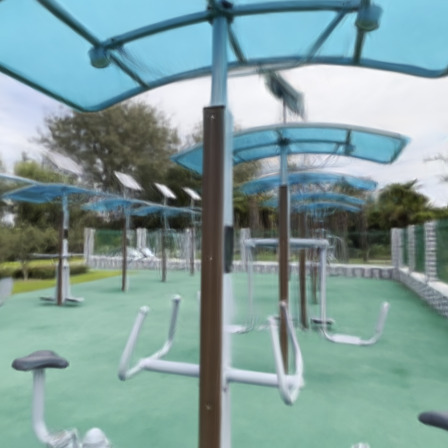}
\qt{642}{1015}{\textbf{25.05}\,/\,\textbf{.849}\,/\,\textbf{.167}}
\qi{594}{886}{97}{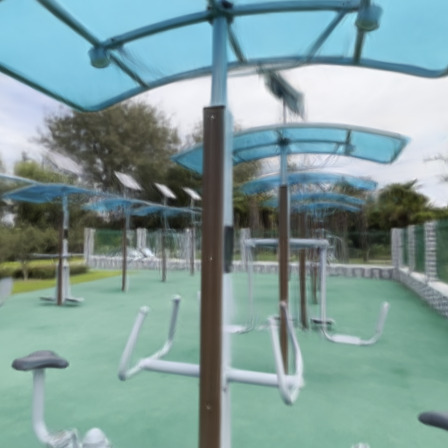}
\qt{642}{884}{24.29\,/\,.839\,/\,.173\\\textit{Speedy-Splat}}
\qi{697}{1017}{97}{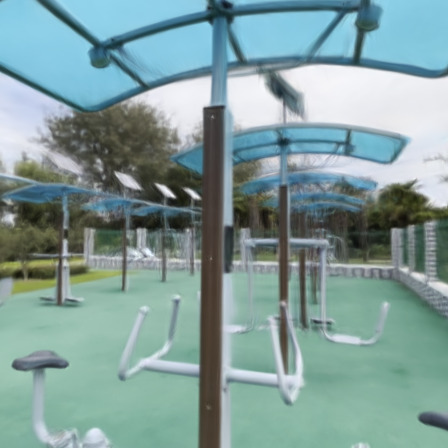}
\qt{746}{1015}{\textbf{25.00}\,/\,\textbf{.846}\,/\,\textbf{.183}}
\qi{697}{886}{97}{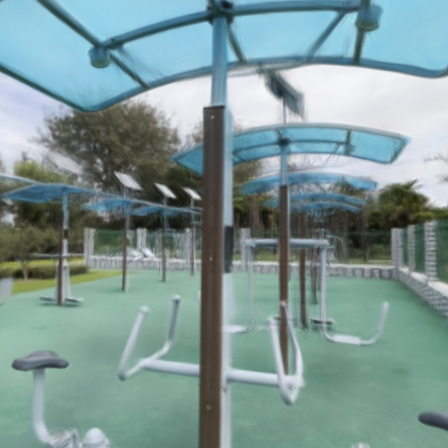}
\qt{746}{884}{23.03\,/\,.831\,/\,.193\\\textit{Speedy-Splat}}
\qi{800}{1017}{97}{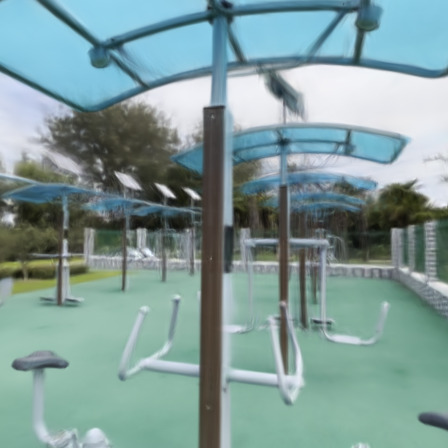}
\qt{848}{1015}{\textbf{24.47}\,/\,\textbf{.833}\,/\,\textbf{.207}}
\qi{800}{886}{97}{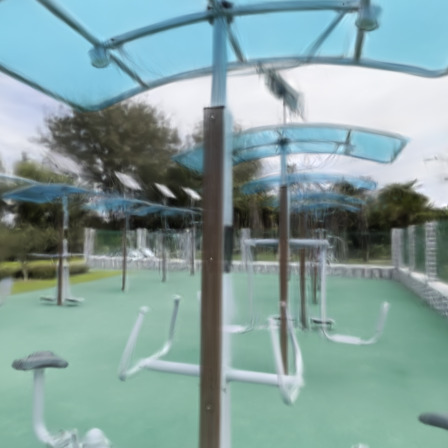}
\qt{848}{884}{21.77\,/\,.798\,/\,.234\\\textit{RAP}}
\qi{903}{1017}{97}{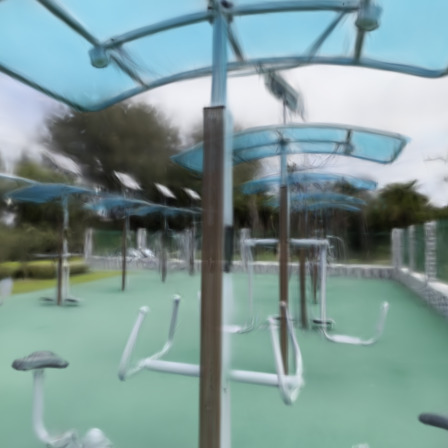}
\qt{952}{1015}{\textbf{21.42}\,/\,\textbf{.779}\,/\,\textbf{.265}}
\qi{903}{886}{97}{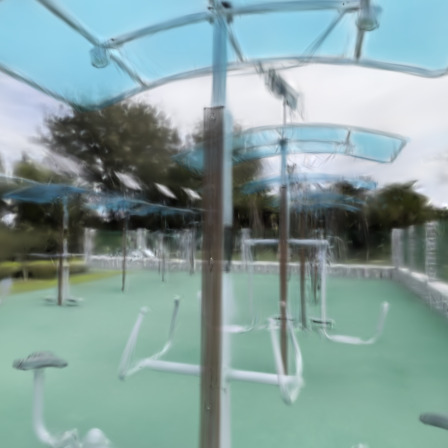}
\qt{952}{884}{18.96\,/\,.739\,/\,.300\\\textit{RAP}}
\put(112,852){\makebox(0,0)[lt]{\footnotesize\textit{ReSplat, RealEstate10K, 16 views, 200{,}704 Gaussians}}}
\qi{112}{602}{228}{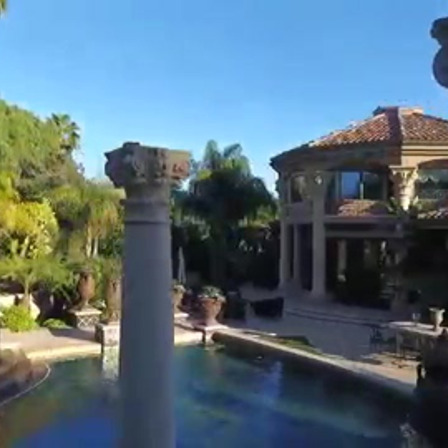}
\put(382,782){\makebox(0,0)[r]{\rotatebox{90}{\makebox[0pt]{\footnotesize\shortstack{\textbf{AnyPrune}\\\textbf{(ours)}}}}}}
\put(382,650){\makebox(0,0)[r]{\rotatebox{90}{\makebox[0pt]{\footnotesize\shortstack{Best\\other}}}}}
\qi{388}{733}{97}{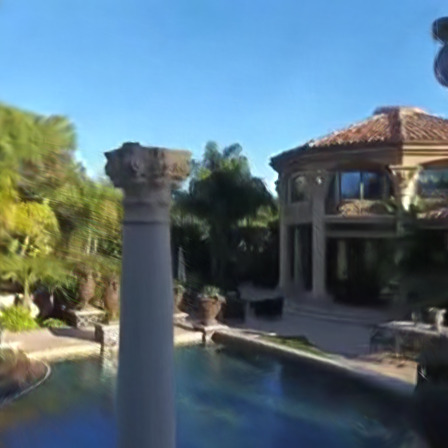}
\qt{436}{731}{\textbf{28.16}\,/\,\textbf{.889}\,/\,\textbf{.230}}
\qi{388}{602}{97}{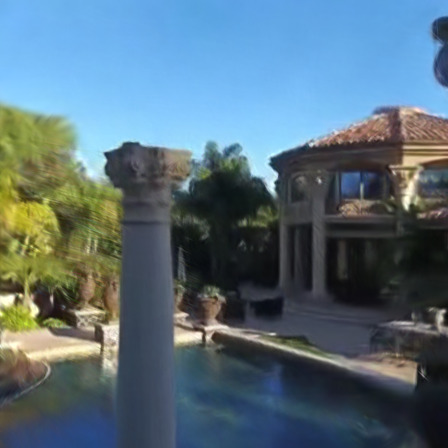}
\qt{436}{600}{27.29\,/\,.878\,/\,.239\\\textit{Unpruned}}
\qi{491}{733}{97}{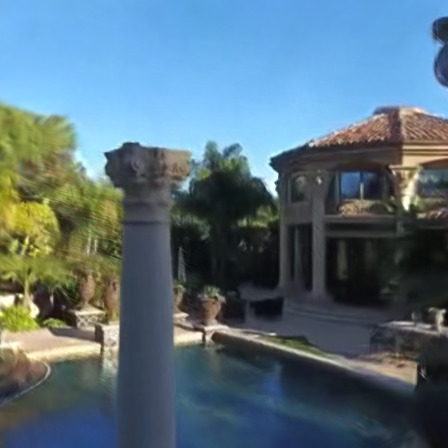}
\qt{540}{731}{\textbf{28.06}\,/\,\textbf{.889}\,/\,\textbf{.229}}
\qi{491}{602}{97}{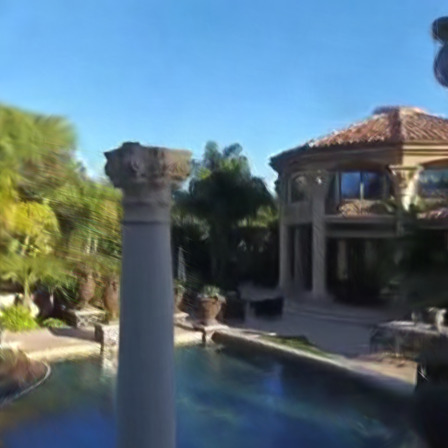}
\qt{540}{600}{27.23\,/\,.877\,/\,.242\\\textit{Speedy-Splat}}
\qi{594}{733}{97}{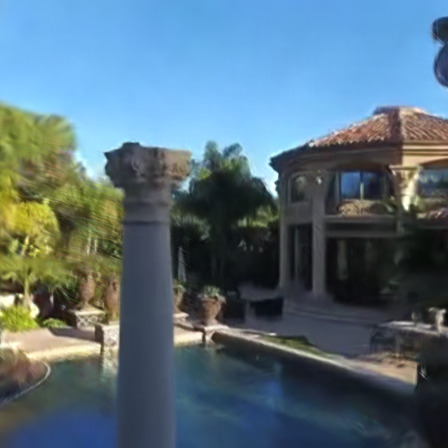}
\qt{642}{731}{\textbf{27.88}\,/\,\textbf{.888}\,/\,\textbf{.235}}
\qi{594}{602}{97}{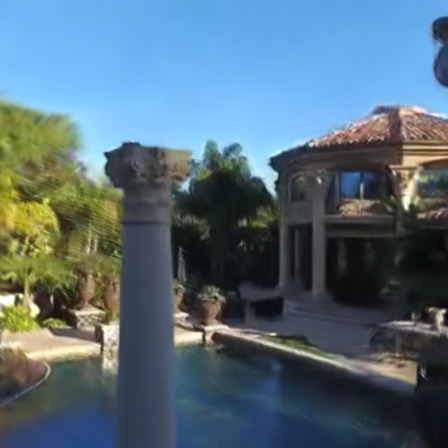}
\qt{642}{600}{26.54\,/\,.872\,/\,.244\\\textit{RAP}}
\qi{697}{733}{97}{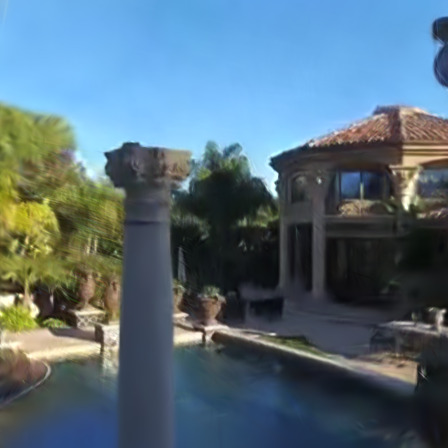}
\qt{746}{731}{\textbf{27.27}\,/\,\textbf{.875}\,/\,\textbf{.251}}
\qi{697}{602}{97}{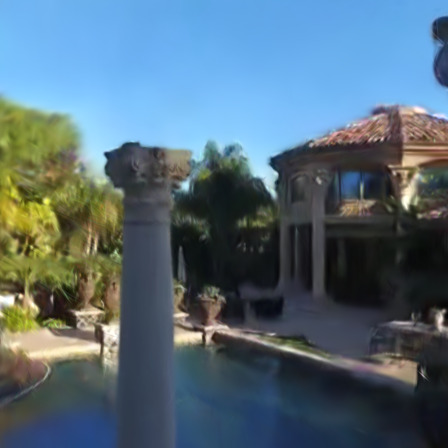}
\qt{746}{600}{25.49\,/\,.858\,/\,.272\\\textit{RAP}}
\qi{800}{733}{97}{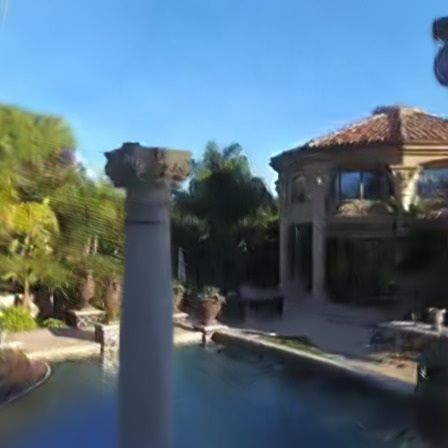}
\qt{848}{731}{\textbf{26.32}\,/\,\textbf{.855}\,/\,\textbf{.276}}
\qi{800}{602}{97}{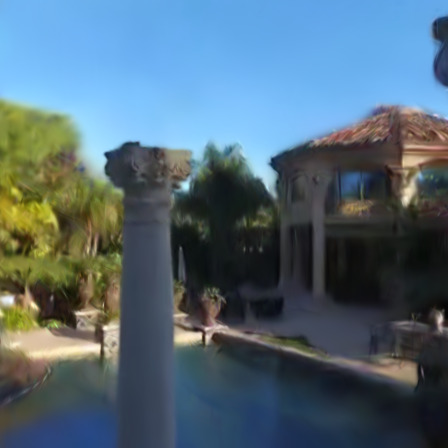}
\qt{848}{600}{23.84\,/\,.827\,/\,.300\\\textit{RAP}}
\qi{903}{733}{97}{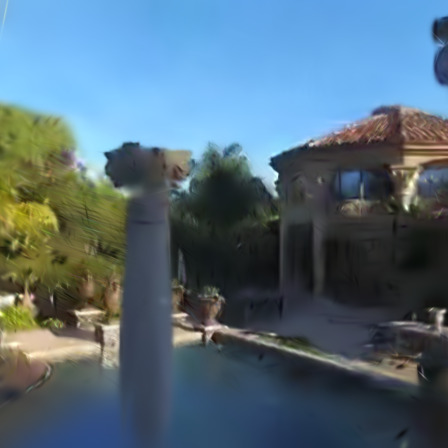}
\qt{952}{731}{\textbf{23.78}\,/\,\textbf{.805}\,/\,\textbf{.329}}
\qi{903}{602}{97}{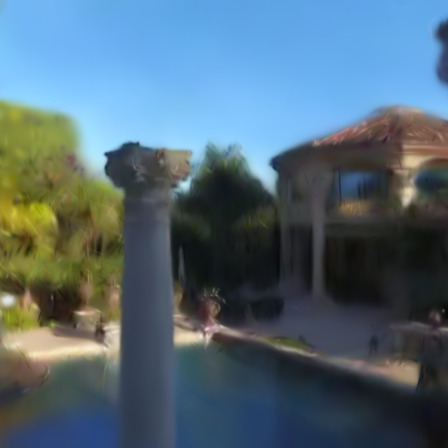}
\qt{952}{600}{22.75\,/\,.783\,/\,.348\\\textit{RAP}}
\put(112,568){\makebox(0,0)[lt]{\footnotesize\textit{TokenGS, DL3DV, 4 views, 262{,}144 Gaussians}}}
\qi{112}{318}{228}{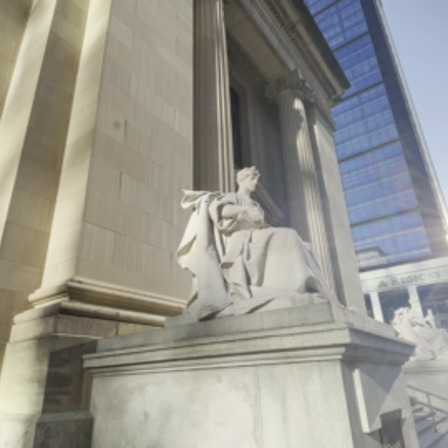}
\put(382,498){\makebox(0,0)[r]{\rotatebox{90}{\makebox[0pt]{\footnotesize\shortstack{\textbf{AnyPrune}\\\textbf{(ours)}}}}}}
\put(382,366){\makebox(0,0)[r]{\rotatebox{90}{\makebox[0pt]{\footnotesize\shortstack{Best\\other}}}}}
\qi{388}{449}{97}{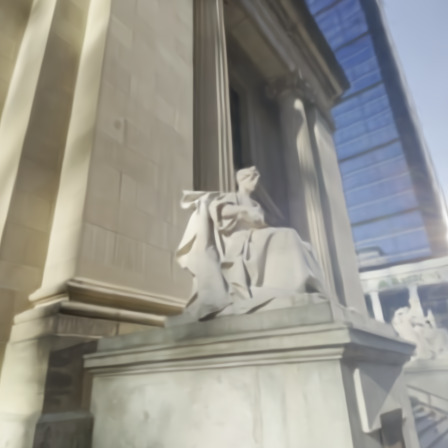}
\qt{436}{447}{\textbf{35.80}\,/\,\textbf{.959}\,/\,\textbf{.155}}
\qi{388}{318}{97}{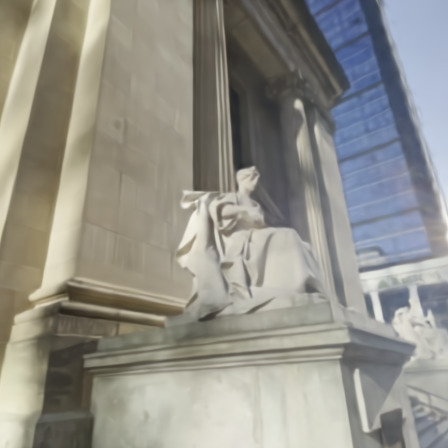}
\qt{436}{316}{30.94\,/\,.949\,/\,.174\\\textit{Unpruned}}
\qi{491}{449}{97}{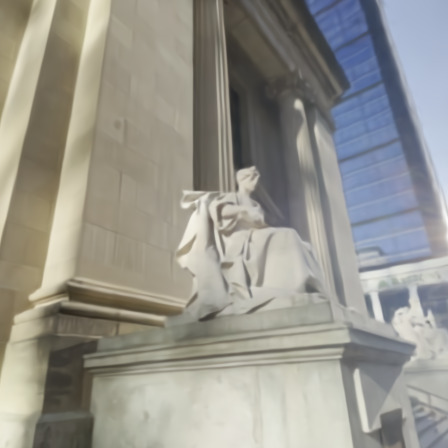}
\qt{540}{447}{\textbf{35.79}\,/\,\textbf{.959}\,/\,\textbf{.155}}
\qi{491}{318}{97}{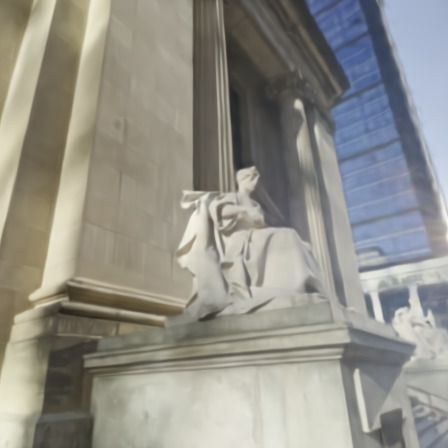}
\qt{540}{316}{30.94\,/\,.949\,/\,.174\\\textit{RAP, released}}
\qi{594}{449}{97}{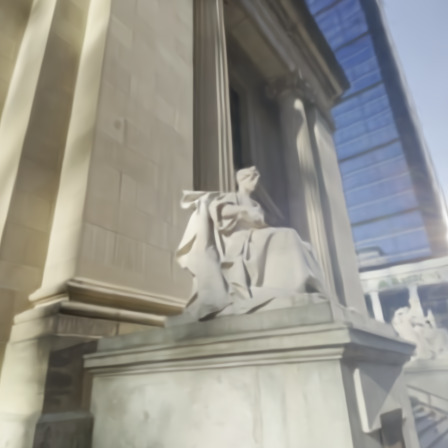}
\qt{642}{447}{\textbf{35.79}\,/\,\textbf{.959}\,/\,\textbf{.155}}
\qi{594}{318}{97}{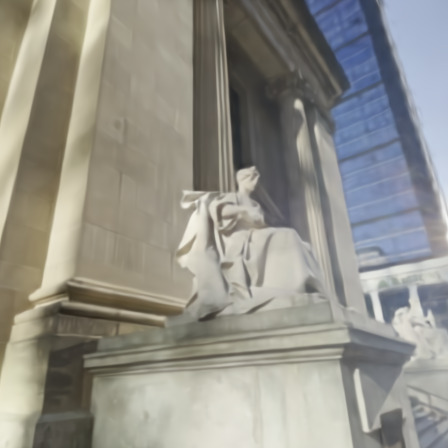}
\qt{642}{316}{30.94\,/\,.949\,/\,.176\\\textit{RAP, released}}
\qi{697}{449}{97}{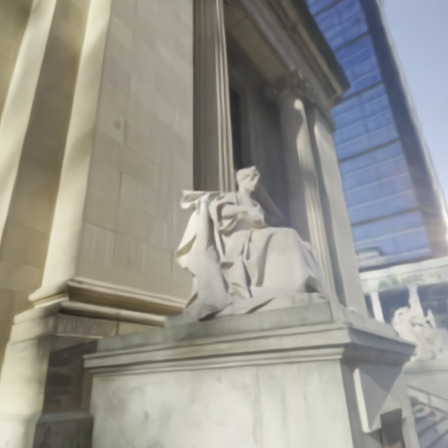}
\qt{746}{447}{\textbf{35.79}\,/\,\textbf{.959}\,/\,\textbf{.155}}
\qi{697}{318}{97}{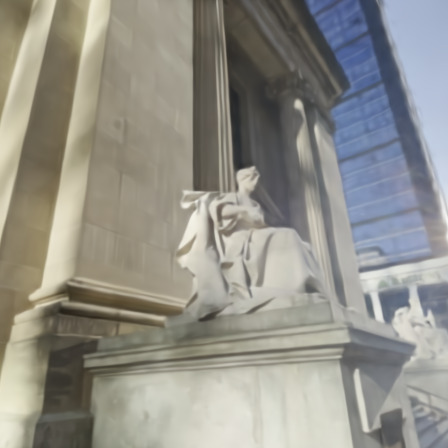}
\qt{746}{316}{31.00\,/\,.948\,/\,.177\\\textit{RAP}}
\qi{800}{449}{97}{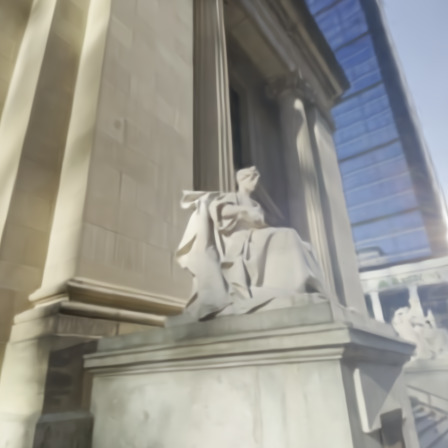}
\qt{848}{447}{\textbf{35.81}\,/\,\textbf{.959}\,/\,\textbf{.158}}
\qi{800}{318}{97}{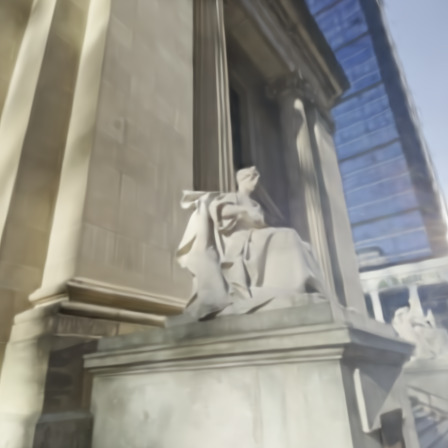}
\qt{848}{316}{31.06\,/\,.946\,/\,.185\\\textit{RAP}}
\qi{903}{449}{97}{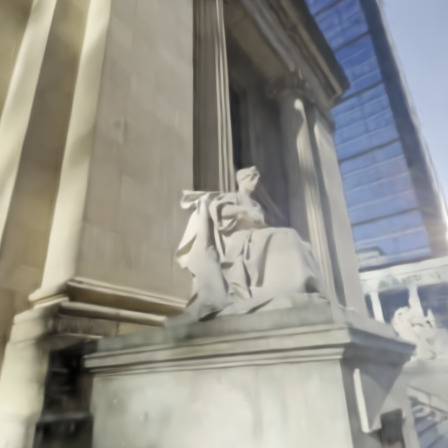}
\qt{952}{447}{\textbf{30.15}\,/\,\textbf{.938}\,/\,\textbf{.214}}
\qi{903}{318}{97}{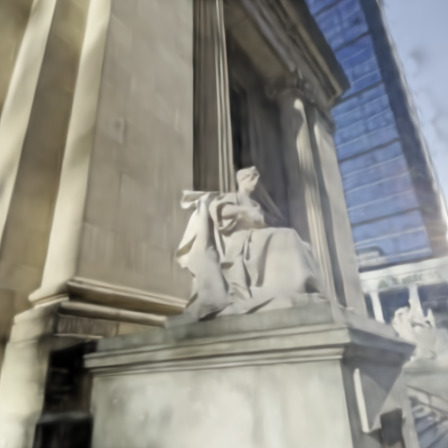}
\qt{952}{316}{26.21\,/\,.921\,/\,.226\\\textit{GaussianPOP}}
\put(112,284){\makebox(0,0)[lt]{\footnotesize\textit{EcoSplat (50\%), ACID, 4 views, 124{,}520 Gaussians}}}
\qi{112}{34}{228}{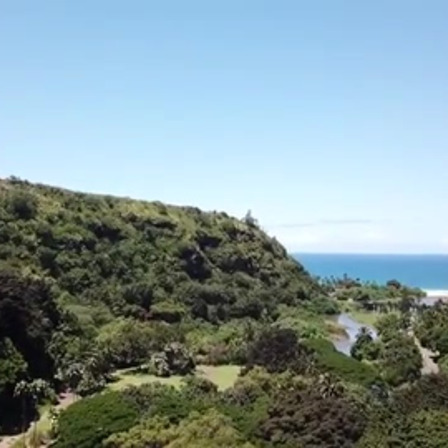}
\put(382,214){\makebox(0,0)[r]{\rotatebox{90}{\makebox[0pt]{\footnotesize\shortstack{\textbf{AnyPrune}\\\textbf{(ours)}}}}}}
\put(382,82){\makebox(0,0)[r]{\rotatebox{90}{\makebox[0pt]{\footnotesize\shortstack{Best\\other}}}}}
\qi{388}{165}{97}{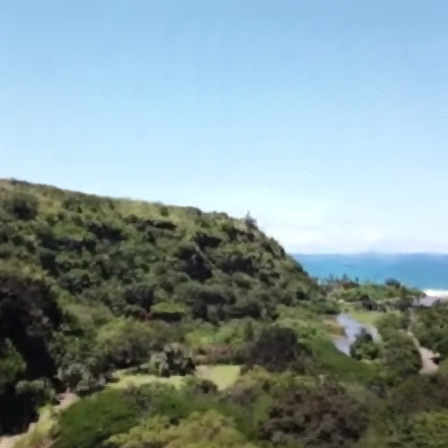}
\qt{436}{163}{\textbf{28.18}\,/\,\textbf{.856}\,/\,\textbf{.281}}
\qi{388}{34}{97}{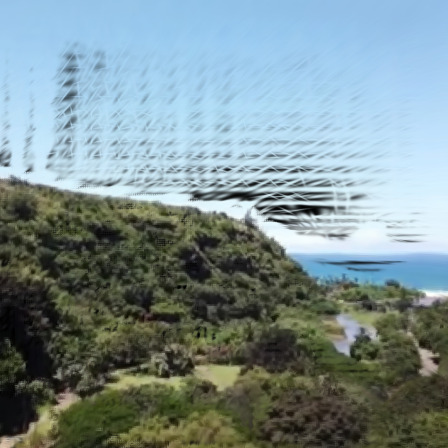}
\qt{436}{32}{19.76\,/\,.725\,/\,.374\\\textit{Unpruned}}
\qi{491}{165}{97}{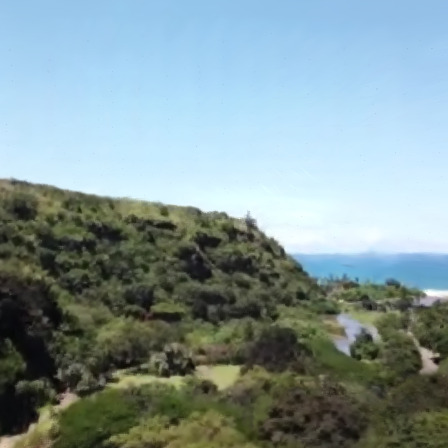}
\qt{540}{163}{\textbf{28.04}\,/\,\textbf{.851}\,/\,\textbf{.283}}
\qi{491}{34}{97}{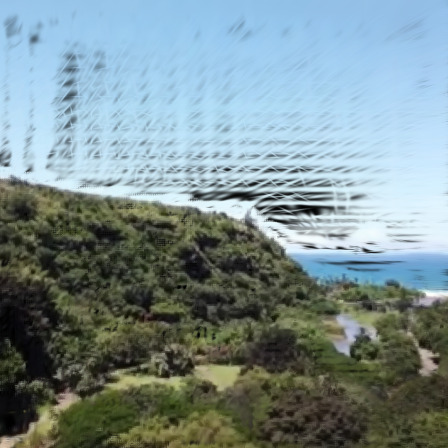}
\qt{540}{32}{19.56\,/\,.708\,/\,.405\\\textit{GaussianPOP}}
\qi{594}{165}{97}{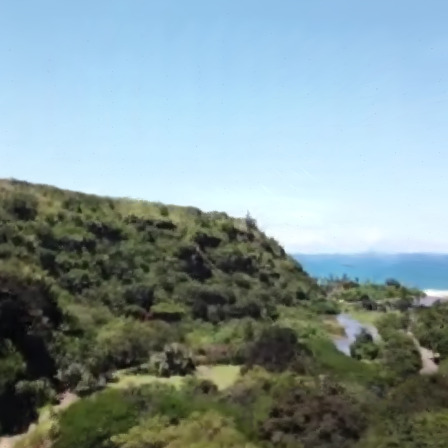}
\qt{642}{163}{\textbf{27.73}\,/\,\textbf{.840}\,/\,\textbf{.291}}
\qi{594}{34}{97}{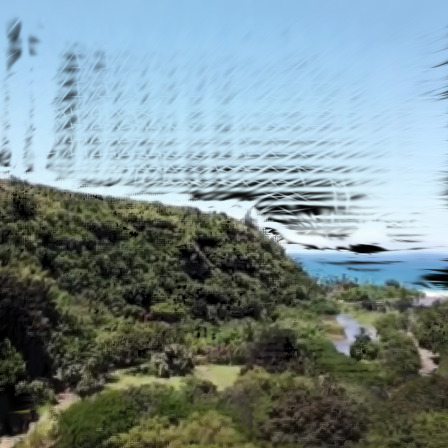}
\qt{642}{32}{18.72\,/\,.673\,/\,.427\\\textit{Speedy-Splat}}
\qi{697}{165}{97}{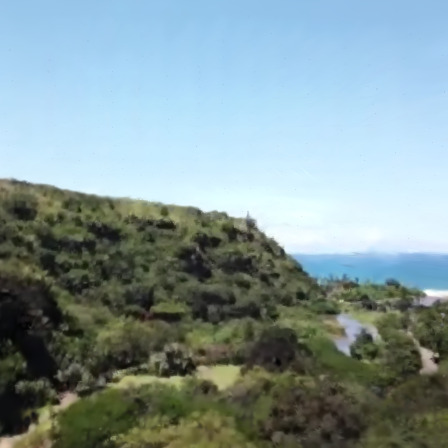}
\qt{746}{163}{\textbf{26.96}\,/\,\textbf{.815}\,/\,\textbf{.307}}
\qi{697}{34}{97}{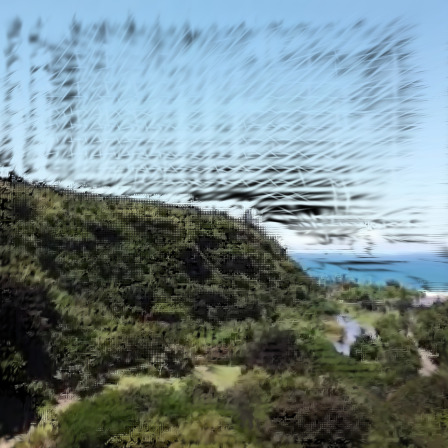}
\qt{746}{32}{17.68\,/\,.555\,/\,.512\\\textit{Mini-Splatting}}
\qi{800}{165}{97}{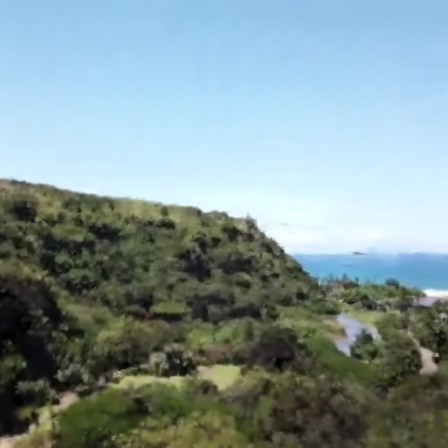}
\qt{848}{163}{\textbf{25.80}\,/\,\textbf{.784}\,/\,\textbf{.336}}
\qi{800}{34}{97}{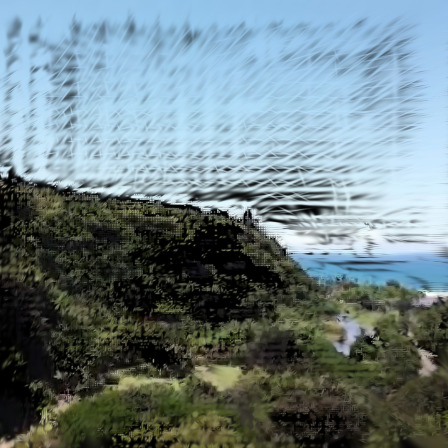}
\qt{848}{32}{16.28\,/\,.477\,/\,.544\\\textit{Mini-Splatting}}
\qi{903}{165}{97}{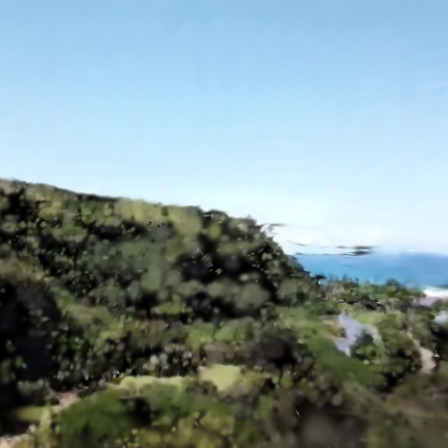}
\qt{952}{163}{\textbf{22.09}\,/\,\textbf{.704}\,/\,\textbf{.396}}
\qi{903}{34}{97}{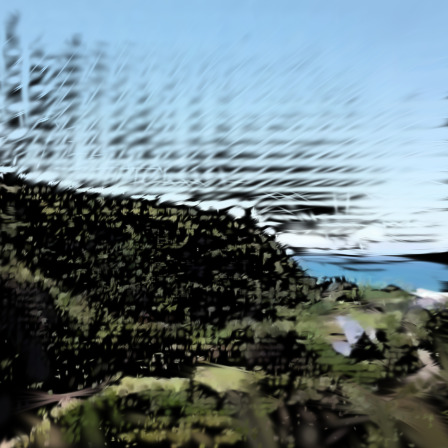}
\qt{952}{32}{14.29\,/\,.422\,/\,.559\\\textit{RAP}}
\end{picture}%
\caption{AnyPrune against the optimization-free pruning method with the highest PSNR on the shown view at each kept share. At 100\%, AnyPrune only adjusts the Gaussians. Under each render: PSNR\,/\,SSIM\,/\,LPIPS of the shown view, in bold where better. In parentheses: the share kept by the pipeline's internal pruning dial, or off.}
\label{fig:qual-best-1}
\end{figure*}

\begin{figure*}[t]
\centering
\setlength{\unitlength}{\hsize}\divide\unitlength by 1000 %
\newcommand{\qf}{\fontsize{6.5}{7.3}\selectfont}%
\newcommand{\qi}[4]{\put(#1,#2){\includegraphics[width=#3\unitlength]{#4}}}%
\newcommand{\qt}[3]{\put(#1,#2){\makebox(0,0)[t]{\qf\shortstack{#3}}}}%
\newcommand{\qT}[3]{\put(#1,#2){\makebox(0,0)[t]{\footnotesize #3}}}%
\noindent\begin{picture}(1000,1158)
\put(226,1158){\makebox(0,0)[t]{\footnotesize GT}}
\put(436,1158){\makebox(0,0)[t]{\footnotesize 100\%}}
\put(540,1158){\makebox(0,0)[t]{\footnotesize 70\%}}
\put(642,1158){\makebox(0,0)[t]{\footnotesize 50\%}}
\put(746,1158){\makebox(0,0)[t]{\footnotesize 30\%}}
\put(848,1158){\makebox(0,0)[t]{\footnotesize 20\%}}
\put(952,1158){\makebox(0,0)[t]{\footnotesize 10\%}}
\put(112,1136){\makebox(0,0)[lt]{\footnotesize\textit{F4Splat (50\%), DL3DV, 16 views, 524{,}287 Gaussians}}}
\qi{112}{886}{228}{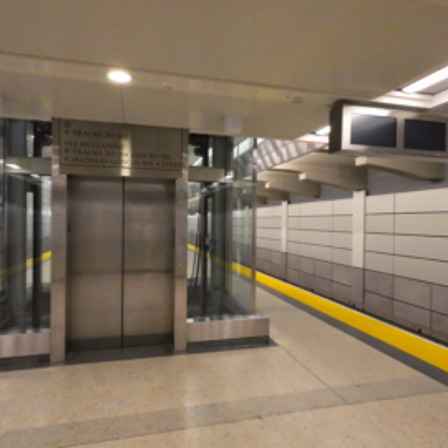}
\put(382,1066){\makebox(0,0)[r]{\rotatebox{90}{\makebox[0pt]{\footnotesize\shortstack{\textbf{AnyPrune}\\\textbf{(ours)}}}}}}
\put(382,934){\makebox(0,0)[r]{\rotatebox{90}{\makebox[0pt]{\footnotesize\shortstack{Best\\other}}}}}
\qi{388}{1017}{97}{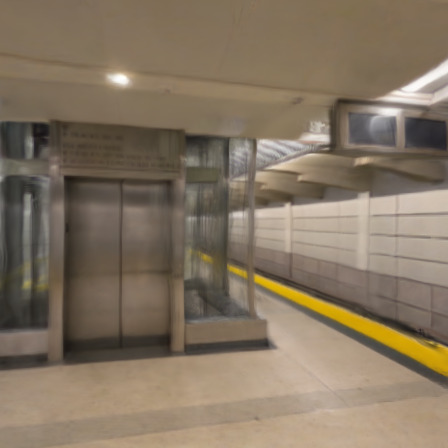}
\qt{436}{1015}{\textbf{22.50}\,/\,\textbf{.767}\,/\,\textbf{.298}}
\qi{388}{886}{97}{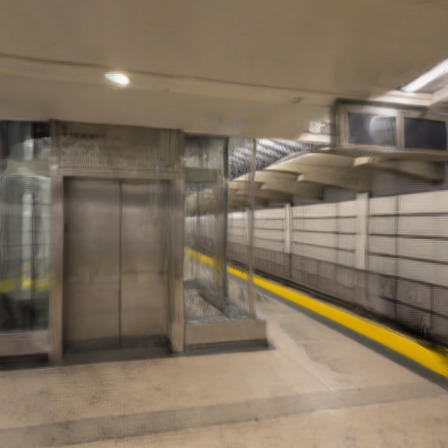}
\qt{436}{884}{20.08\,/\,.708\,/\,.342\\\textit{Unpruned}}
\qi{491}{1017}{97}{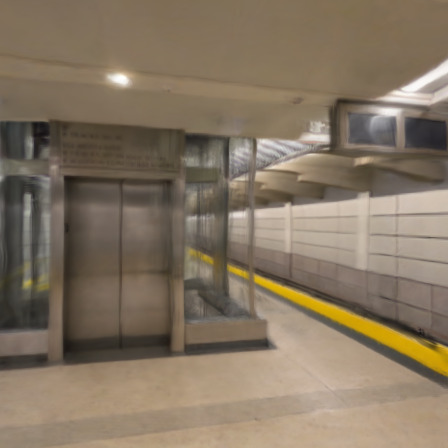}
\qt{540}{1015}{\textbf{22.29}\,/\,\textbf{.762}\,/\,\textbf{.302}}
\qi{491}{886}{97}{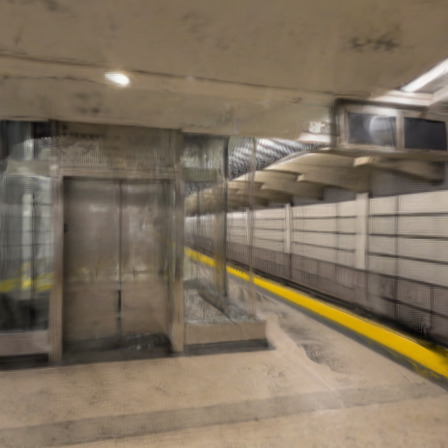}
\qt{540}{884}{19.78\,/\,.682\,/\,.393\\\textit{Speedy-Splat}}
\qi{594}{1017}{97}{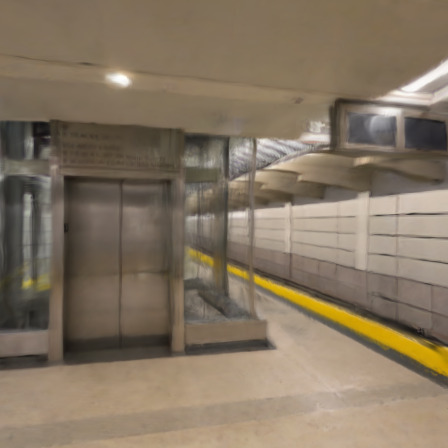}
\qt{642}{1015}{\textbf{21.99}\,/\,\textbf{.754}\,/\,\textbf{.325}}
\qi{594}{886}{97}{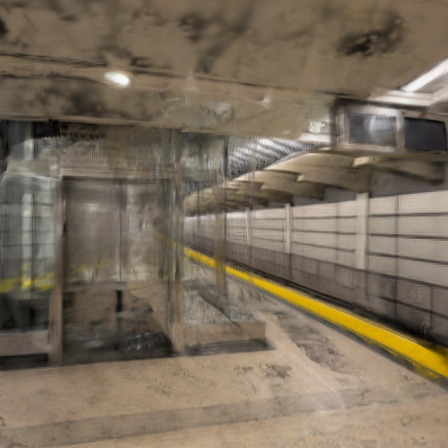}
\qt{642}{884}{18.50\,/\,.605\,/\,.461\\\textit{Speedy-Splat}}
\qi{697}{1017}{97}{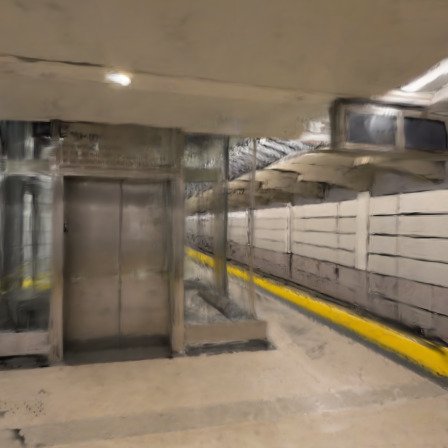}
\qt{746}{1015}{\textbf{21.15}\,/\,\textbf{.719}\,/\,\textbf{.383}}
\qi{697}{886}{97}{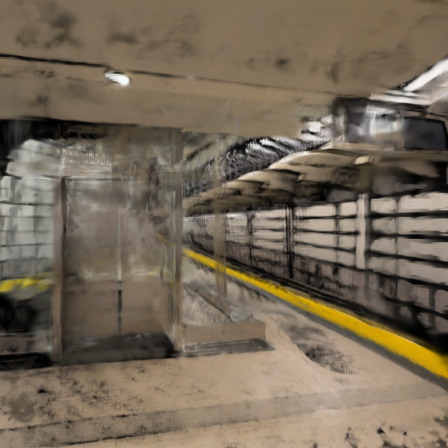}
\qt{746}{884}{15.76\,/\,.543\,/\,.496\\\textit{RAP, released}}
\qi{800}{1017}{97}{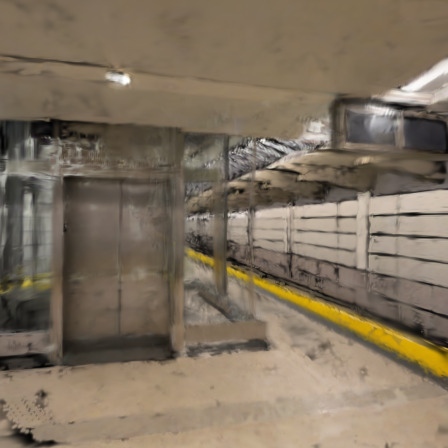}
\qt{848}{1015}{\textbf{19.65}\,/\,\textbf{.666}\,/\,\textbf{.439}}
\qi{800}{886}{97}{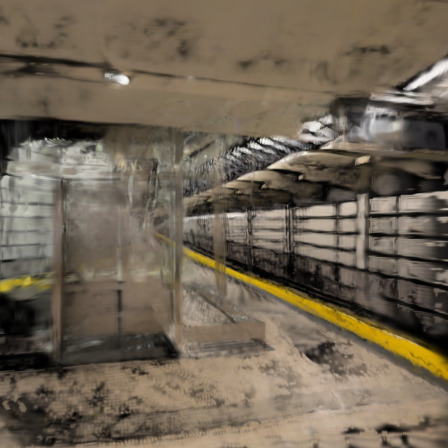}
\qt{848}{884}{14.47\,/\,.484\,/\,.530\\\textit{RAP}}
\qi{903}{1017}{97}{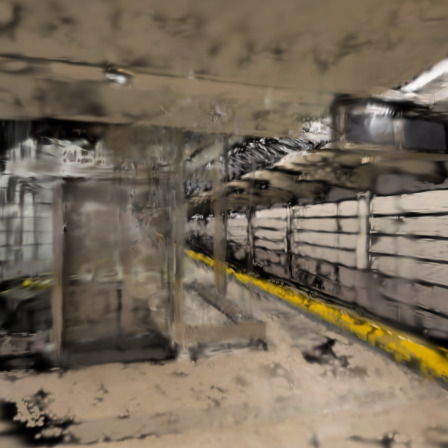}
\qt{952}{1015}{\textbf{16.37}\,/\,\textbf{.546}\,/\,\textbf{.524}}
\qi{903}{886}{97}{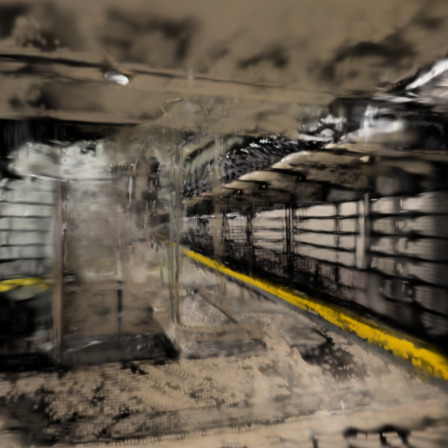}
\qt{952}{884}{12.98\,/\,.409\,/\,.574\\\textit{RAP}}
\put(112,852){\makebox(0,0)[lt]{\footnotesize\textit{EcoSplat (off), ACID, 8 views, 498{,}072 Gaussians}}}
\qi{112}{602}{228}{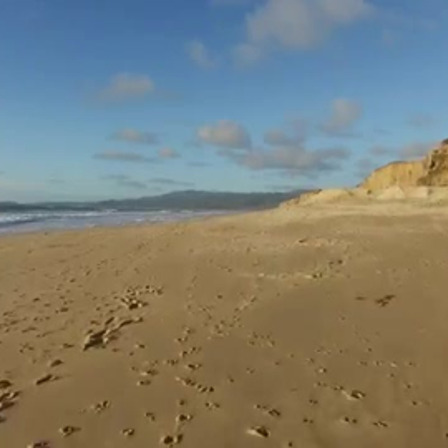}
\put(382,782){\makebox(0,0)[r]{\rotatebox{90}{\makebox[0pt]{\footnotesize\shortstack{\textbf{AnyPrune}\\\textbf{(ours)}}}}}}
\put(382,650){\makebox(0,0)[r]{\rotatebox{90}{\makebox[0pt]{\footnotesize\shortstack{Best\\other}}}}}
\qi{388}{733}{97}{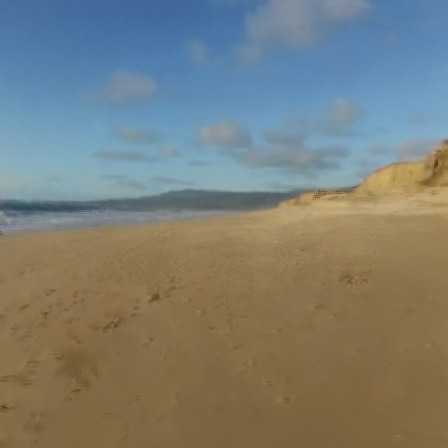}
\qt{436}{731}{\textbf{29.84}\,/\,\textbf{.872}\,/\,.339}
\qi{388}{602}{97}{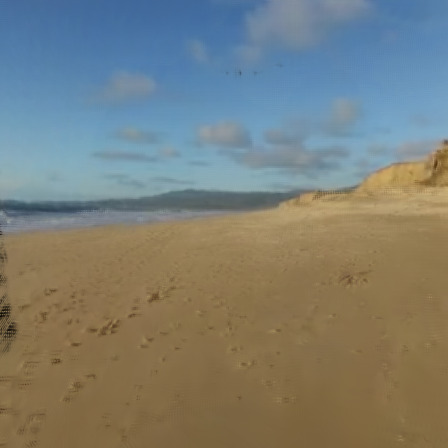}
\qt{436}{600}{29.27\,/\,.868\,/\,\textbf{.338}\\\textit{Unpruned}}
\qi{491}{733}{97}{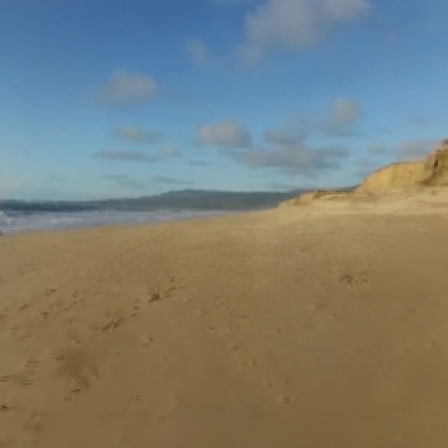}
\qt{540}{731}{\textbf{29.84}\,/\,\textbf{.872}\,/\,.339}
\qi{491}{602}{97}{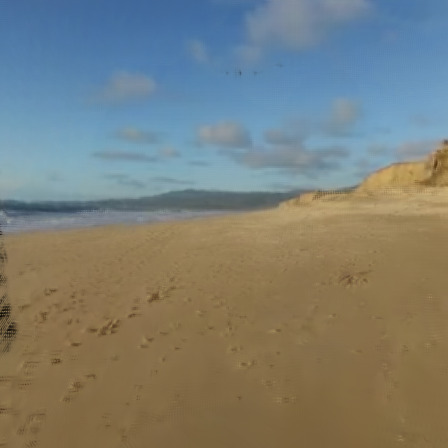}
\qt{540}{600}{29.27\,/\,.868\,/\,\textbf{.338}\\\textit{GaussianPOP}}
\qi{594}{733}{97}{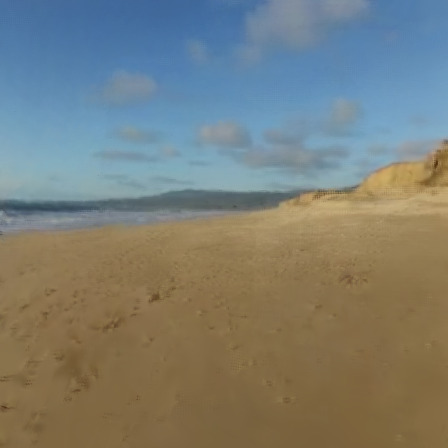}
\qt{642}{731}{\textbf{29.81}\,/\,\textbf{.871}\,/\,\textbf{.339}}
\qi{594}{602}{97}{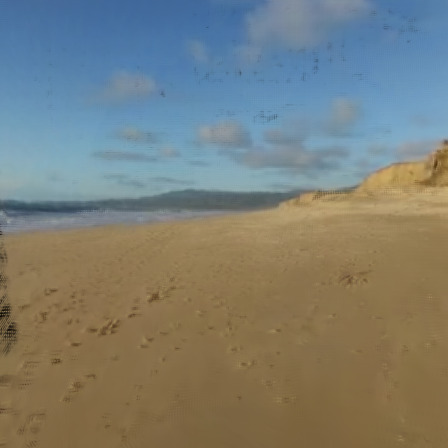}
\qt{642}{600}{29.15\,/\,.861\,/\,.389\\\textit{GaussianPOP}}
\qi{697}{733}{97}{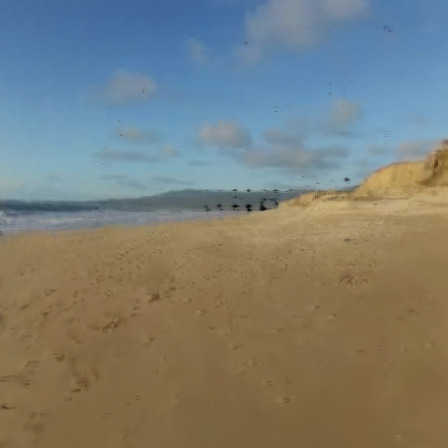}
\qt{746}{731}{\textbf{29.33}\,/\,\textbf{.859}\,/\,\textbf{.371}}
\qi{697}{602}{97}{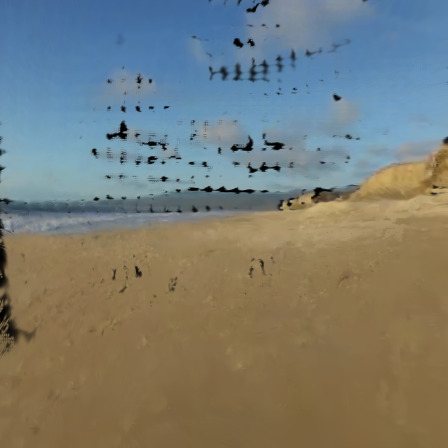}
\qt{746}{600}{23.62\,/\,.796\,/\,.451\\\textit{RAP, released}}
\qi{800}{733}{97}{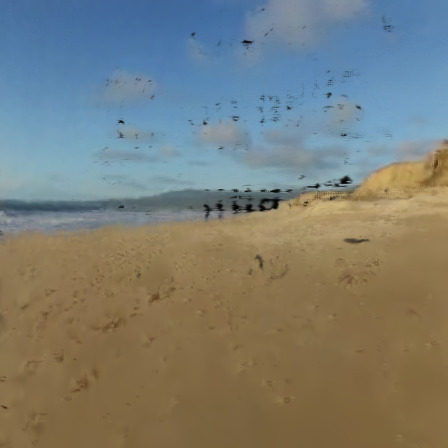}
\qt{848}{731}{\textbf{27.48}\,/\,\textbf{.832}\,/\,\textbf{.411}}
\qi{800}{602}{97}{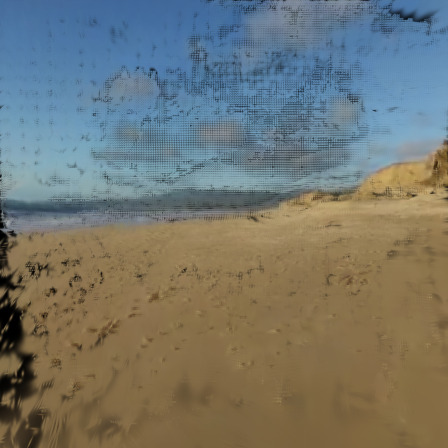}
\qt{848}{600}{21.95\,/\,.666\,/\,.555\\\textit{Mini-Splatting}}
\qi{903}{733}{97}{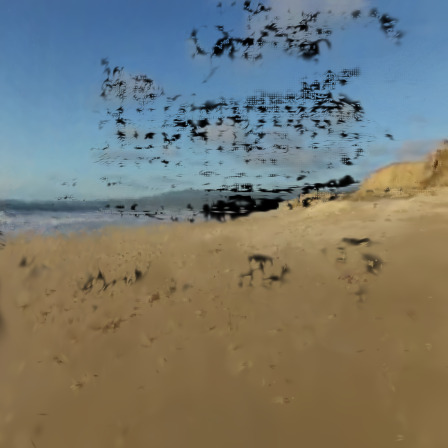}
\qt{952}{731}{\textbf{21.24}\,/\,\textbf{.726}\,/\,\textbf{.471}}
\qi{903}{602}{97}{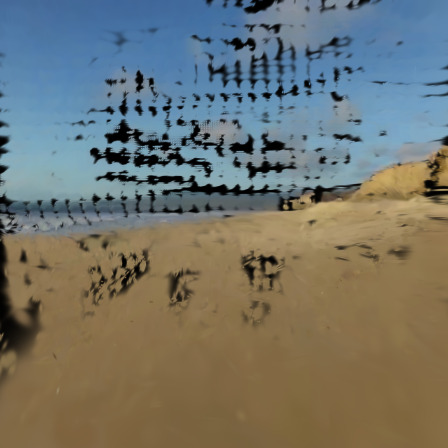}
\qt{952}{600}{18.00\,/\,.667\,/\,.523\\\textit{RAP, released}}
\put(112,568){\makebox(0,0)[lt]{\footnotesize\textit{F4Splat (off), DL3DV, 4 views, 262{,}144 Gaussians}}}
\qi{112}{318}{228}{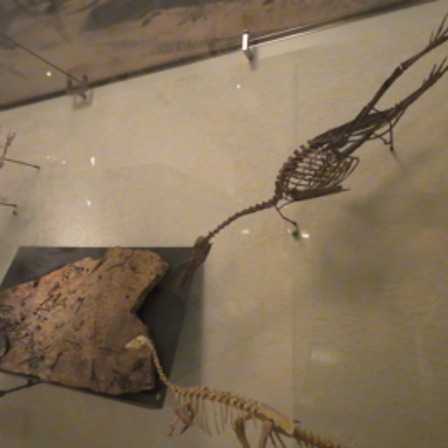}
\put(382,498){\makebox(0,0)[r]{\rotatebox{90}{\makebox[0pt]{\footnotesize\shortstack{\textbf{AnyPrune}\\\textbf{(ours)}}}}}}
\put(382,366){\makebox(0,0)[r]{\rotatebox{90}{\makebox[0pt]{\footnotesize\shortstack{Best\\other}}}}}
\qi{388}{449}{97}{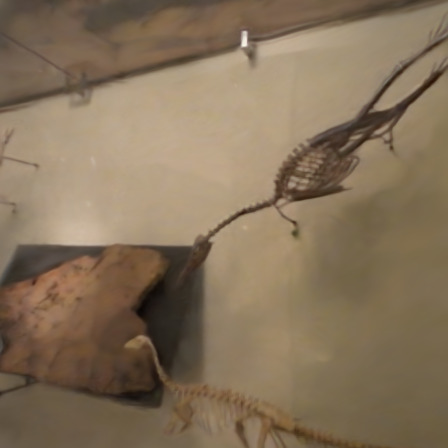}
\qt{436}{447}{\textbf{29.89}\,/\,.864\,/\,.338}
\qi{388}{318}{97}{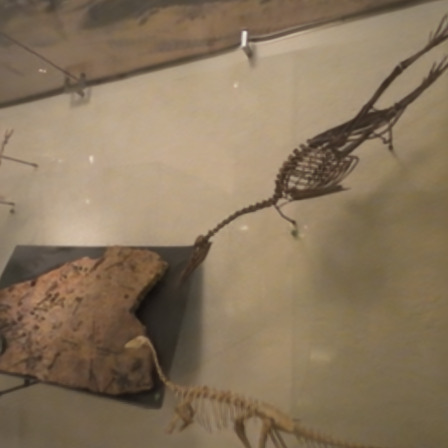}
\qt{436}{316}{29.16\,/\,\textbf{.873}\,/\,\textbf{.245}\\\textit{Unpruned}}
\qi{491}{449}{97}{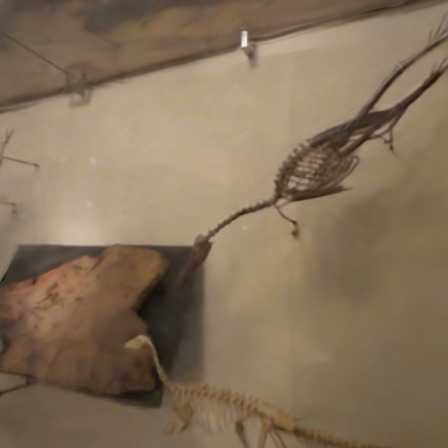}
\qt{540}{447}{\textbf{29.86}\,/\,\textbf{.863}\,/\,.342}
\qi{491}{318}{97}{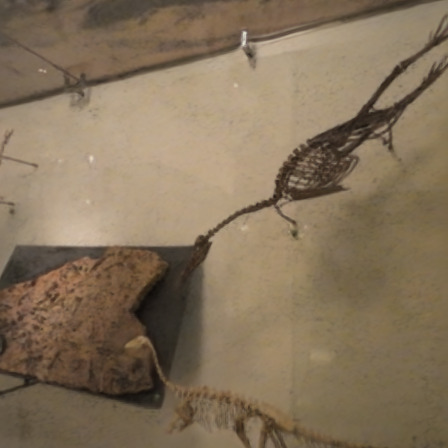}
\qt{540}{316}{29.20\,/\,.850\,/\,\textbf{.306}\\\textit{Mini-Splatting}}
\qi{594}{449}{97}{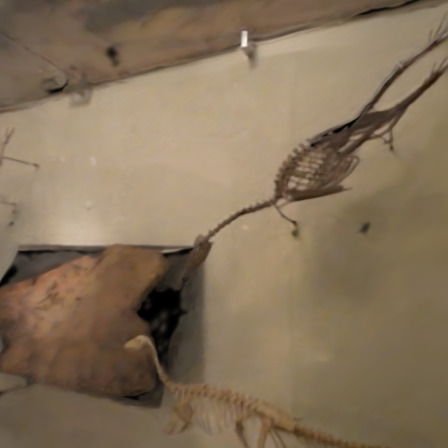}
\qt{642}{447}{\textbf{28.74}\,/\,\textbf{.846}\,/\,\textbf{.367}}
\qi{594}{318}{97}{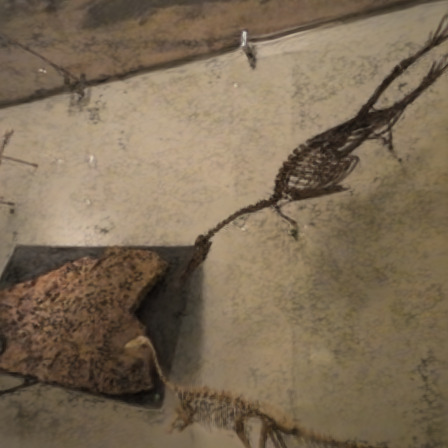}
\qt{642}{316}{25.83\,/\,.769\,/\,.400\\\textit{Mini-Splatting}}
\qi{697}{449}{97}{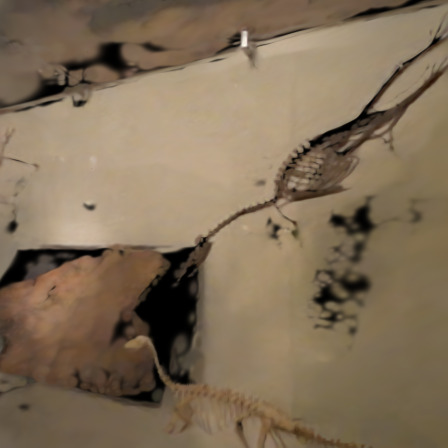}
\qt{746}{447}{\textbf{23.66}\,/\,\textbf{.785}\,/\,\textbf{.423}}
\qi{697}{318}{97}{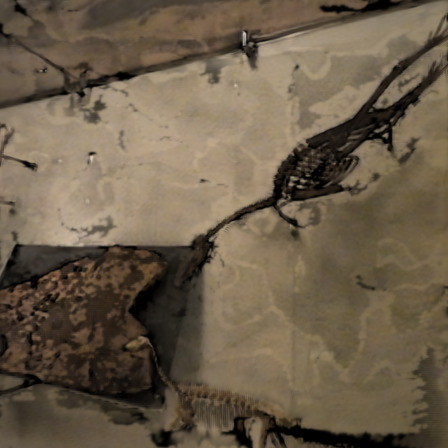}
\qt{746}{316}{19.13\,/\,.703\,/\,.431\\\textit{Mini-Splatting}}
\qi{800}{449}{97}{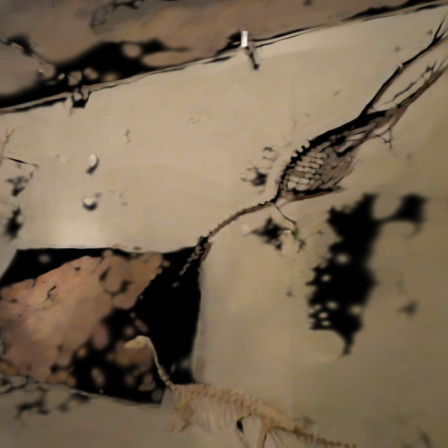}
\qt{848}{447}{\textbf{18.82}\,/\,\textbf{.695}\,/\,\textbf{.469}}
\qi{800}{318}{97}{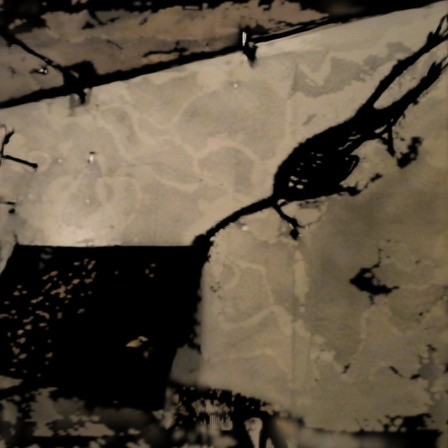}
\qt{848}{316}{14.97\,/\,.546\,/\,.518\\\textit{Provable coresets}}
\qi{903}{449}{97}{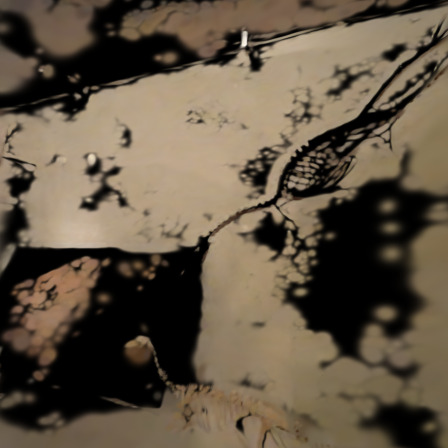}
\qt{952}{447}{\textbf{14.13}\,/\,\textbf{.492}\,/\,\textbf{.560}}
\qi{903}{318}{97}{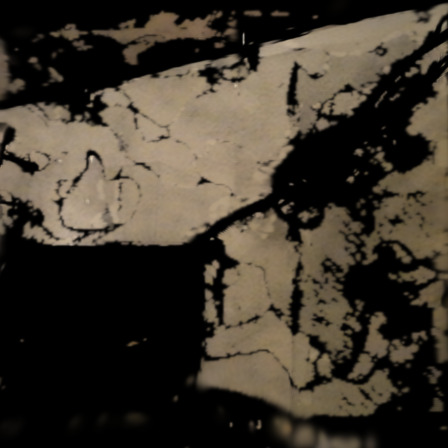}
\qt{952}{316}{11.15\,/\,.289\,/\,.648\\\textit{Provable coresets}}
\put(112,284){\makebox(0,0)[lt]{\footnotesize\textit{AdaptiveSplat (off), RealEstate10K, 16 views, 3{,}211{,}264 Gaussians}}}
\qi{112}{34}{228}{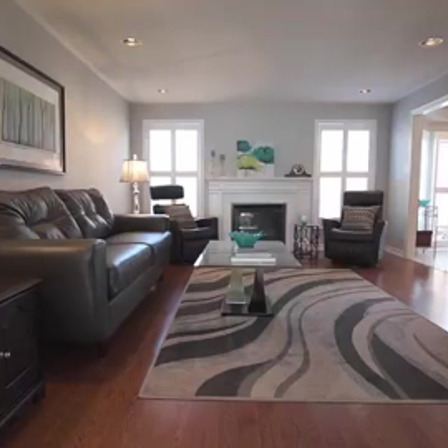}
\put(382,214){\makebox(0,0)[r]{\rotatebox{90}{\makebox[0pt]{\footnotesize\shortstack{\textbf{AnyPrune}\\\textbf{(ours)}}}}}}
\put(382,82){\makebox(0,0)[r]{\rotatebox{90}{\makebox[0pt]{\footnotesize\shortstack{Best\\other}}}}}
\qi{388}{165}{97}{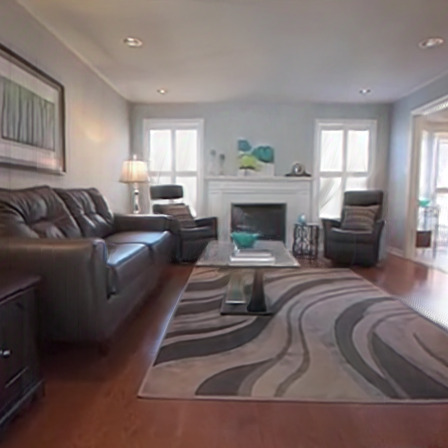}
\qt{436}{163}{\textbf{27.51}\,/\,\textbf{.892}\,/\,\textbf{.185}}
\qi{388}{34}{97}{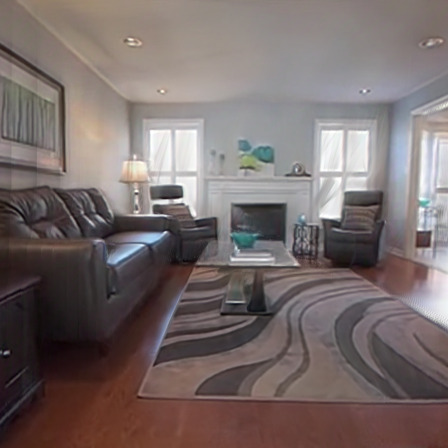}
\qt{436}{32}{25.39\,/\,.882\,/\,.201\\\textit{Unpruned}}
\qi{491}{165}{97}{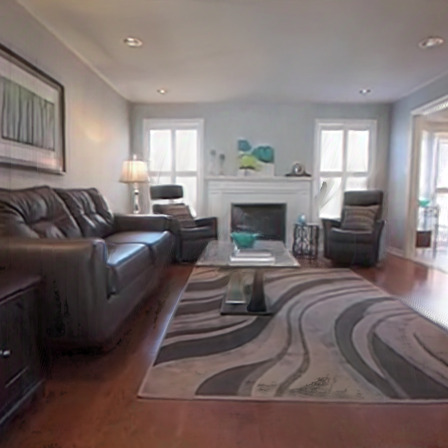}
\qt{540}{163}{\textbf{27.12}\,/\,\textbf{.879}\,/\,\textbf{.209}}
\qi{491}{34}{97}{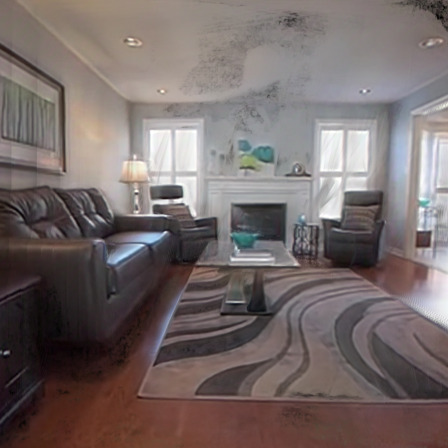}
\qt{540}{32}{24.81\,/\,.853\,/\,.262\\\textit{GaussianPOP}}
\qi{594}{165}{97}{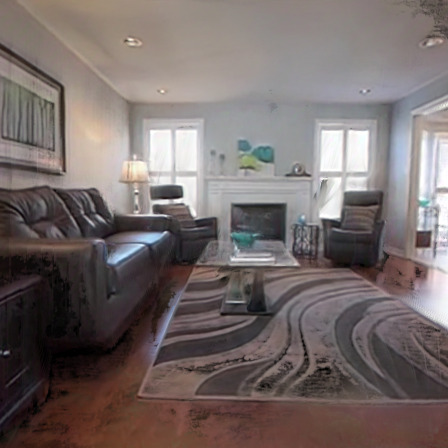}
\qt{642}{163}{\textbf{24.58}\,/\,\textbf{.823}\,/\,\textbf{.271}}
\qi{594}{34}{97}{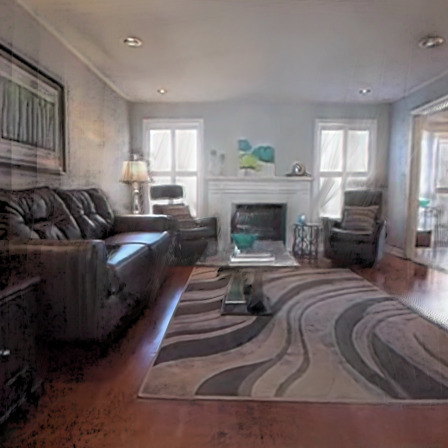}
\qt{642}{32}{22.22\,/\,.801\,/\,.298\\\textit{RAP}}
\qi{697}{165}{97}{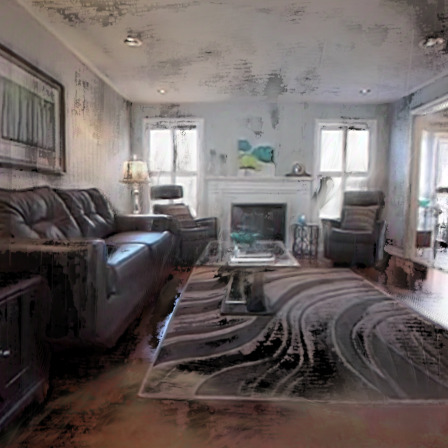}
\qt{746}{163}{\textbf{19.58}\,/\,\textbf{.699}\,/\,\textbf{.403}}
\qi{697}{34}{97}{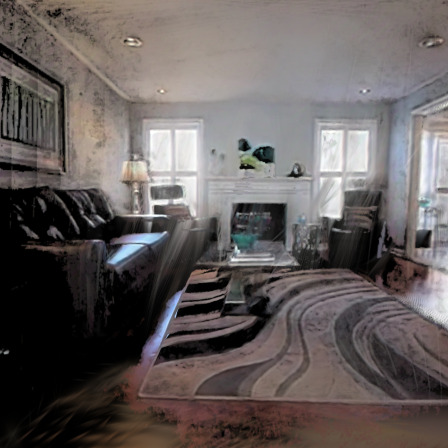}
\qt{746}{32}{18.24\,/\,.626\,/\,.421\\\textit{RAP}}
\qi{800}{165}{97}{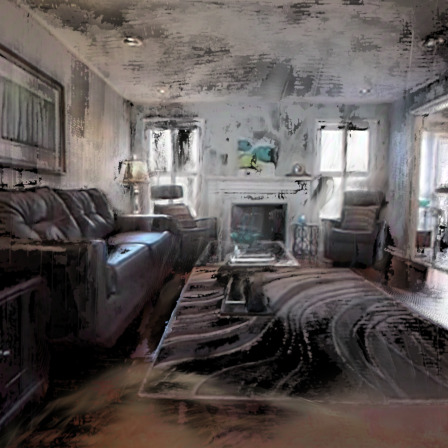}
\qt{848}{163}{\textbf{16.16}\,/\,\textbf{.555}\,/\,\textbf{.494}}
\qi{800}{34}{97}{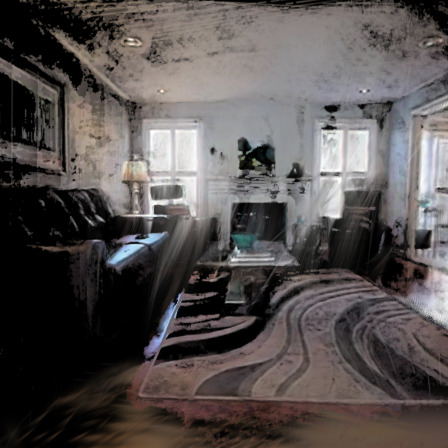}
\qt{848}{32}{14.79\,/\,.483\,/\,.514\\\textit{RAP}}
\qi{903}{165}{97}{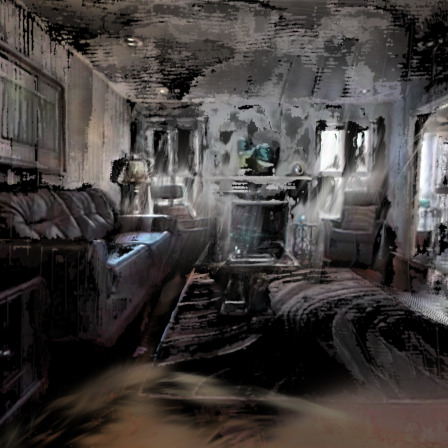}
\qt{952}{163}{\textbf{11.74}\,/\,\textbf{.369}\,/\,\textbf{.584}}
\qi{903}{34}{97}{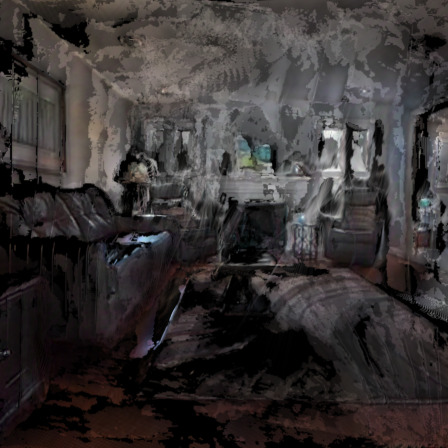}
\qt{952}{32}{10.75\,/\,.358\,/\,.601\\\textit{Mini-Splatting}}
\end{picture}%
\caption{AnyPrune against the optimization-free pruning method with the highest PSNR on the shown view at each kept share. At 100\%, AnyPrune only adjusts the Gaussians. Under each render: PSNR\,/\,SSIM\,/\,LPIPS of the shown view, in bold where better. In parentheses: the share kept by the pipeline's internal pruning dial, or off.}
\label{fig:qual-best-2}
\end{figure*}

\Cref{fig:qual-full-1,fig:qual-full-2} show AnyPrune and every optimization-free pruning method of \cref{sec:exp-pruning} on AdaptiveSplat, with its internal pruning at 60\%, and on Off The Grid, from 10\% to 100\% of the Gaussians kept.
\Cref{fig:qual-best-1,fig:qual-best-2} show the other pipelines, each against the best other method at every kept share.
In the figures, the setting of a pipeline's internal pruning dial is in parentheses.
When few Gaussians are kept, the other methods leave holes and floaters, while AnyPrune keeps the surfaces covered.
Its adjustments also remove artifacts of the unpruned field, such as the grid pattern of EcoSplat at 4 views.

\end{document}